# Sanskrit Knowledge-based Systems: Annotation and Computational Tools

*A thesis submitted*

in Partial Fulfillment of the Requirements

for the Degree of

Doctor of Philosophy

by

**Hrishikesh Rajesh Terdalkar**

14111265

*to the*

DEPARTMENT OF COMPUTER SCIENCE AND ENGINEERING

INDIAN INSTITUTE OF TECHNOLOGY KANPUR

June, 2023

# CERTIFICATE

It is certified that the work contained in the thesis titled **Sanskrit Knowledge-based Systems: Annotation and Computational Tools**, by **Hrishikesh Rajesh Terdalkar**, has been carried out under my supervision and that this work has not been submitted elsewhere for a degree.

Prof. Arnab Bhattacharya
Department of Computer Science and Engineering
IIT Kanpur

June, 2023

# DECLARATION

This is to certify that the thesis titled **Sanskrit Knowledge-based Systems: Annotation and Computational Tools** has been authored by me. It presents the research conducted by me under the supervision of **Prof. Arnab Bhattacharya**.
To the best of my knowledge, it is an original work, both in terms of research content and narrative, and has not been submitted elsewhere, in part or in full, for a degree. Further, due credit has been attributed to the relevant state-of-the-art and collaborations (if any) with appropriate citations and acknowledgments, in line with established norms and practices.

Signature

Name: Hrishikesh Rajesh Terdalkar
Programme: PhD
Department: Computer Science and Engineering
Indian Institute of Technology Kanpur
Kanpur 208016

# ABSTRACT

Name of Student: **Hrishikesh Rajesh Terdalkar** Roll No.: **14111265**

Degree for which submitted: **Doctor of Philosophy**

Department: **Computer Science and Engineering**

Thesis Title: **Sanskrit Knowledge-based Systems: Annotation and Computational Tools**

Name of Thesis Supervisor: **Prof. Arnab Bhattacharya**

A Knowledge Base (KB) is a representation of real-world knowledge in a particular domain in a computer system. Knowledge Graphs (KG) are KBs that use graph as the underlying data structure. Knowledge-based systems are software that utilize knowledge bases to solve problems. In the field of Natural Language Processing (NLP), Question Answering (QA) is a problem of finding answers to natural language questions posed by humans. KGs are an integral part of addressing the problem of question answering. Difficulty of constructing knowledge graphs depends greatly on the language and the resources available, such as datasets, tools and technologies.

Sanskrit is a classical language with a vast amount of written literature on a wide variety of topics. However, most of this literature is not available in a format that is readily usable by computer systems. As a result, from a computational perspective, Sanskrit is still considered a low-resource language.

In this thesis, we make contributions towards the ultimate goal of question answering in Sanskrit through construction of various knowledge-based systems in Sanskrit. We first present a framework that attempts to answer factual questions through an automated construction of KGs. We highlight the shortcomings and lim-itations of the state-of-the-art of Sanskrit NLP. The scarcity of appropriate datasets poses a significant challenge in the development and evaluation of automated systems for knowledge graph construction. Human annotation plays an important role for the creation of such datasets. There is also a need for annotation tools with task-specific and intuitive interfaces to simplify the tedious task of manual annotation.

We present *Sangrahaka*, an annotator-friendly, web-based tool for ontology-driven annotation of entities and relationships towards the construction of knowledge graphs. It also supports querying. The tool is language and corpus-agnostic but customizable for specific needs. We demonstrate the usefulness of the tool through a real-world annotation task on Bhāvaprakāśanighaṇṭu, an Āyurveda text. We showcase a carefully constructed extensive ontology suitable for this task, resulting in annotations that contribute to the development of a knowledge graph and querying framework. These contributions are based on three chapters from Bhāvaprakāśanighaṇṭu.

Then, we present *Antarlekhaka*, a general purpose multi-task annotation system for manual annotation of a comprehensive set of NLP tasks. The system supports annotation towards multiple categories of NLP tasks: sentence boundary detection, canonical word ordering, token annotation, token classification, token graph, sentence classification and sentence graph. The annotation is performed in a sequential manner for small logical units of text (e.g., a verse). We highlight the utility of the tool through the application of the tool for the annotation of Vālmīki Rāmāyaṇa, resulting in datasets for NLP tasks of sentence boundary detection, canonical word ordering, named entity recognition, action graph construction and co-reference resolution.

Both *Sangrahaka* and *Antarlekhaka* are presented as full-stack web-based software to support distributed annotation. They are designed to be easily configurable, web-deployable, customizable and with a multi-tier permission system. They are actively being used in real-world annotation tasks. The annotator-friendly and intuitive annotation interfaces of these tools have received positive feedback from the users, and they outperform other annotation tools in objective evaluation.

Sanskrit text corpora have undergone large-scale digitization efforts using OCR

technology, inadvertently leading to the introduction of various errors. In this thesis, we present Chandojñānam, a system for identifying and utilizing Sanskrit meters. Apart from its core functionality of meter identification, the system also enables finding fuzzy matches based on sequence matching, thereby facilitating the correction of inaccuracies in digital corpora. The user-friendly interface of Chandojñānam displays the scansion, a graphical representation of the metrical pattern. Additionally, the system supports meter identification from uploaded images through the utilization of optical character recognition (OCR) engines. The text can be processed in either line-by-line mode or verse-by-verse mode.

Finally, as part of our research contribution, we offer an extensive range of web-interfaces, tools, and software libraries specifically designed to highlight and utilize the computational aspects of Sanskrit. This diverse compilation includes Jñānasaṅgrahaḥ, a comprehensive web-based collection of various computational applications dedicated to the Sanskrit language. The overarching aim of Jñānasaṅgrahaḥ is to present the features of the Sanskrit language in an accessible manner, even for enthusiastic users with limited Sanskrit backgrounds. Within this collection, you will find Saṅkhyāpaddhatiḥ, a web-interface that encompasses three ancient numeral systems, enabling the representation of numbers as text. Additionally, we offer Chandojñānam, a system for Sanskrit meter identification and utilization, as well as Varṇajñānam, a utility pertaining to varṇa, a phonetic unit of the Sanskrit language. Furthermore, our contributions extend to a Telegram bot designed to assist learners in comprehending Sanskrit grammar. Lastly, we have developed a set of Python libraries to aid programmers in working with Sanskrit corpora. These include *PyCDSL*, a Python library and a Command Line Interface (CLI) to simplify the processes of downloading, managing, and accessing Sanskrit dictionaries, *Heritage.py*, a Python interface to The Sanskrit Heritage site and *sanskrit-text*, a library for the manipulation of Sanskrit alphabet. Collectively, these resources serve as catalysts for encouraging and enabling a wider audience to delve into the richness of Sanskrit and its profound cultural heritage.

In conclusion, this thesis addresses the challenges and opportunities in the development of knowledge systems for Sanskrit, with a focus on question answering. By proposing a framework for the automated construction of knowledge graphs, introducing annotation tools for ontology-driven and general-purpose tasks, and offering a diverse collection of web-interfaces, tools, and software libraries, we have made significant contributions to the field of computational Sanskrit. These contributions not only enhance the accessibility and accuracy of Sanskrit text analysis but also pave the way for further advancements in knowledge representation and language processing. Ultimately, this research contributes to the preservation, understanding, and utilization of the rich linguistic information embodied in Sanskrit texts.

# Acknowledgements

I am humbled and immensely grateful to all the individuals and organizations who have played pivotal roles in the completion of this thesis. Their support, guidance, and contributions have been instrumental in shaping this work and have enriched my research journey. I would like to express my heartfelt gratitude to each of them.

First and foremost, I owe a great debt of gratitude to my thesis advisor and guide, Prof. Arnab Bhattacharya, for accepting me as his student, for his invaluable insights, esteemed guidance and constant support throughout my research journey. I am truly grateful for his patience and support during the challenging times. His expertise and encouragement have not only fueled my passion but have also shaped the direction and outcomes of this thesis.

I extend my thanks to Dr. Kripabandhu Ghosh, Dr. Sai Susarla, Prof. Anil Kumar Gourishetty, Dr. Amrith Krishna, Prof. Amba Kulkarni, Prof. Pawan Goyal and Jivnesh Sandhan for their discussions, insightful feedback and collaborations during various phases of my research.

I am grateful to the annotators who actively participated and made significant contributions to the annotation tasks. I would like to extend my thanks to Vishakha Deulgaokar for sharing her valuable insights and dedicated involvement as an annotator.

My sincere appreciation goes to all the professors at the Department of CSE at IIT Kanpur for their help, guidance, insightful interactions, and encouragement on various occasions. Their expertise and support have been invaluable. I would like to mention Prof. R K Ghosh, Prof. Satyadev Nandakumar, Prof. Amey Karkare,

Prof. Subhajit Roy, Prof. Sumit Ganguly, Prof. Rajat Mittal, Prof. Indranil Saha, Prof. Piyush Rai, Prof. Nisheeth Srivastava, Prof. T V Prabhakar, Prof. Vinay P Namboodiri, Prof. Sandeep Shukla, and Prof. Sruti Srinivasa Ragavan.

I would like to express my gratitude to all the supporting staff in the institute including but not limited to the CSE Department, Hall 4, and Hall 8 for their assistance and contributions, which have made my life at the institute easier.

I am thankful to the anonymous reviewers for their valuable comments and suggestions, which have significantly enhanced the quality of this work.

I would like to express my deep appreciation to Central Sanskrit University and our Sanskrit teacher, Pralay Manna, for their invaluable contribution in enhancing our understanding of the language. I am truly grateful to A V S D S Mahesh and Dr. Chaitali Dangarikar for sharing their profound insights on various Sanskrit-related topics. Furthermore, I want to sincerely appreciate the Vyoma Linguistic Labs Foundation for their resolute dedication to reviving and preserving Sanskrit through traditional teaching methods. In particular, I am thankful to teacher Sowmya Krishnapur for introducing me to the fascinating world of Vyakarana Shastra and providing me with invaluable guidance along the way.

I am grateful to my friends, relatives and teachers who have been a part of my educational journey and have provided guidance, motivation, and encouragement along the way.

I am immensely grateful to Garima Gaur, Keerti Chaudhury, Rujuta Pimprikar, and Umair Ahmed for their incredible enthusiasm, curiosity, guidance, and support. Whether we were attending conferences, engaging in random discussions in the canteen, playing cards, or tackling TA duties, those moments have truly enriched my experience at IITK. I would also like to give a special shout-out to Shubham Sahai and Adarsh Jagannatha for their camaraderie, helpfulness, and active participation in various group activities. The memories of our fun-filled ‘chai pe charcha’ sessions with Tejas Gandhi and Saurabh Srivastava always bring a smile to my face. Those lively conversations, filled with gossip and valuable insights, added an excit-ing touch to our days. Additionally, I would like to mention Arvapalli Sai Susmitha, Aakankshka Verma, Sumit Lahiri, and Pankaj Kumar Kalita for the numerous meaningful interactions.

I am filled with deep gratitude for the continual support and friendship of Awanish Pandey. Throughout my journey of PhD, his presence has been a beacon of light, guiding me through both curricular and extracurricular activities. The moments we shared in KD106, delving into discussions on politics, science, stocks, exams, movies, songs, and every conceivable topic, have undoubtedly been some of the most cherished and treasured moments of my life at IITK. The laughter we've shared, the obstacles we've conquered, and the countless moments of camaraderie have left an indelible mark on my heart.

My companion since the undergraduate years, Abhishek Dang, has had a profound impact on my life. Spanning over more than 15 years, our friendship has been a constant that not many could understand. Together, we have embarked on countless adventures, be it our sessions of playing or discussing DotA, the joyous card and board game nights, the exhilarating cycling trips, or the thought-provoking discussions. My association with Abhishek holds a special place in my heart. I extend my sincere thanks for being a constant source of genuine advice that has guided me through various ups and downs.

No words of thanks can be enough to express my heartfelt gratitude to my wife, Shubhangi Agarwal, for her unwavering support, understanding, and patience throughout this endeavor. She has been my rock in this PhD journey. Together, we have experienced the highs and lows, the challenges and triumphs, and she has been an active part in all aspects of my academic and non-academic pursuits. In times of uncertainty and fatigue, she has provided kind encouragement and a listening ear, helping me navigate through the hurdles with steadfast determination. Her presence has brought stability to my life, and I am forever grateful to her.

Last, but not the least, I am truly indebted to my parents, Vinda and Rajesh Terdalkar, for their boundless love, unconditional support, and steadfast belief in me. They have always been there for me, through thick and thin. I attribute all my achievements to their selfless sacrifices. I know that I can always count on their love and encouragement, and I am forever grateful for everything they have done for me. I hope that I can make them proud. I would also like to remember the treasured memories of my late grandparents, Sarita and Ratnakar Terdalkar. Their profound influence has shaped not only my character but also the trajectory of my academic pursuits. Their love and legacy will continue to live on in me.

*To Aai and Baba*

*(Vinda and Rajesh Terdalkar)*

# Contents

# List of Tables

# List of Figures

# Chapter 1

# Introduction

ॐ न हि ज्ञानेन सदृशं पवित्रमिह विद्यते।

*In this world, there is nothing as purifying as Knowledge.*

---

Śrīmadbhagavadgītā 4.38

A Knowledge Base (KB) serves as a repository of real-world knowledge within a specific domain, encompassing domain-specific information, factual data, rules, and procedures that are employed to represent and address problems in a particular field or domain. Knowledge Graphs (KG), on the other hand, utilize a graph structure as the foundational data representation for KBs. Knowledge-based systems are software applications designed to utilize knowledge bases for solving problems.

Natural language refers to the language used by humans to communicate with each other, such as Spanish, English, Hindi, etc. It is characterized by its complexity, flexibility, and ambiguity, which makes it challenging for computers to understand and generate. Natural Language Processing (NLP) is a subfield of Artificial Intelligence (AI) that focuses on enabling computers to process, understand, and generate natural language [Jurafsky and Martin, 2008, Winston, 1984]. NLP techniques are used to build a wide range of applications, such as machine translation, sentiment analysis, speech recognition, text summarization, and question answering. Question Answering (QA) is a field of NLP that involves building computer algorithms that can automatically answer questions posed in a natural language by humans.

The goal of QA is to develop systems that can understand the meaning of a question and provide a relevant and accurate answer based on a large collection of knowledge sources, such as documents, databases, or other structured and unstructured data. QA can be used in a variety of applications, such as virtual assistants, search engines, customer service, and information retrieval systems. The development of QA systems involves several techniques, including information retrieval, natural language understanding, machine learning, and knowledge representation.

Since the introduction of QA task by [Voorhees, 1999], one of the prominent approaches for QA has been through use of KBs [Hirschman and Gaizauskas, 2001, Kiyota et al., 2002, Yih et al., 2015]. KGs play a vital role by providing a structured framework for effective QA [Diefenbach et al., 2018, Gutiérrez and Sequeda, 2021]. The complexity involved in constructing knowledge graphs is heavily influenced by factors such as the specific language being studied and the availability of diverse resources, including datasets, tools, and technologies.

## 1.1 Motivation

Sanskrit (Devanagari: संस्कृत, IAST[1]: Saṃskṛta) is a classical language with a vast amount of written literature on a wide variety of topics including but not limited to philosophy, history, mathematics, science and religion. It is considered one of the oldest and most well-preserved languages in the world, with a rich literary tradition spanning several thousand years [Burrow, 2001, Macdonell, 1915]. However, the large volume of such works and the relative lack of proficiency in the language have kept treasures in those text hidden from the common man.

Many of the Sanskrit texts are technical in nature; prime examples of which include Āyurveda (आयुर्वेद) texts such as Bhāvaprakāśa (भावप्रकाश). The nighaṇṭu (निघण्टु) portion of Bhāvaprakāśa is compiled as a glossary of the various substances and their properties (guṇa, गुण). Although the information is generally provided in a format that enables scholars to study and analyse it systematically, the large volume of such texts makes it harder for any individual to extract all the information. An automated system can, therefore, greatly aid this processing of information. However, to the best of our knowledge, there does not exist any system that can query this knowledge trove directly and automatically.

[1]In this thesis, the International Alphabet of Sanskrit Transliteration (IAST) encoding scheme is utilized for the romanized format of Sanskrit words. `https://en.wikipedia.org/wiki/International_Alphabet_of_Sanskrit_Transliteration`

While it can be argued that English translations of Bhāvaprakāśanighaṇṭu are available, and building information retrieval (IR) systems for it is a routine for today's IR/NLP tools, there are two main shortcomings of it. First, there are many such nighaṇṭu texts and translations in English are available for only a minuscule number of them. Second, and more importantly, many of the translations of Sanskrit texts had been done without a proper understanding of the context and culture in which they were composed in the first place. They may had been forced to use English words and phrases that are not a true reflection of the spirit of the original meaning [Malhotra and Babaji, 2020]. A notable case in point, as mentioned by Swami Vivekananda [Vivekananda, 2019], is the word śraddhā (श्रद्धा), for which the English translation "regards" is not enough. Thus, it is always best to rely on the original language. The need of the hour, hence, is to use natural language processing (NLP) of Sanskrit itself to understand the texts in Sanskrit.

This research aims to contribute to the development of knowledge systems in Sanskrit towards the ultimate goal of question answering. Overall, the thesis presents a comprehensive approach to addressing the challenges of developing knowledge systems in Sanskrit and provides a set of tools and resources that can be used by researchers and practitioners in the field of Sanskrit NLP.

## 1.2 Challenges

Unraveling information from Sanskrit texts in a targeted and systematic manner can not only help in enhancing the knowledge systems but can also revive an interest in the language. Unfortunately, most of this literature is not available in plain text format that is readily usable by computer systems. As a result, from a computational perspective, Sanskrit is still considered a low-resource language. We bring to the fore multiple challenges in processing of Sanskrit texts arising from the complexity and intricacy of Sanskrit as well as limitations of state-of-the-art techniques.

First, the state of the art of natural language processing in Indian languages, unfortunately, is not as advanced as that in English or some other European languages [Kurian, 2014, Harish and Rangan, 2020]. Indian languages, and in particular Sanskrit, are morphologically richer. Therefore, tasks such as lemmatization and parts-of-speech tagging are harder and more error-prone in these languages.

Second, some technical texts use their own jargon where certain words may be used in a specific meaning. For example, Aṣṭādhyāyī, a work on Sanskrit grammar by Pāṇini uses specific combinations[2] of grammatical cases (vibhakti) to denote which action is to be performed.

Third, names in Sanskrit are meaningful words and, therefore, identifying named entities is particularly hard. An interesting example in Rāmāyaṇa is janaka (जनक), which means "father" in general, but is also the name of a prominent character.

Fourth, synonyms are often used to refer to the same person. In many cases, higher-order grammar rules may be required to parse the meaning of a word and understand that it is a synonym. For example, it is not mentioned anywhere in the Rāmāyaṇa text that Dāśarathī refers to the son of Daśaratha and, hence, mostly used as a synonym to Rāma. However, Sanskrit grammar rules make it clear to someone who understands the language. Unfortunately, automatic language processing tools are incapable of using such higher-order rules at present. Moreover, even if we somehow infer that Dāśarathī means 'the son of Daśaratha', it is not evident which son of Daśaratha it specifically refers to. Another example is that of the word Rāghava, signifying a descendant of Raghu, which is used in different contexts in Rāmāyaṇa to denote both Rāma and Daśaratha. Similarly, the term Bhārata is context-dependent in the Mahābhārata, as it has been used to refer to both Yudhiṣṭhira and Arjuna in different contexts. The precise referent of these terms greatly relies on the specific context in which they are employed.

[2]The presence of nominative (prathamā), genitive (ṣaṣṭhī) and locative (saptamī) cases in the same sentence might not convey any special meaning in a normal text, but, in Aṣṭādhyāyī, it specifies a process to be followed to transform words, e.g., rule 6.1.77 from Aṣṭādhyāyī (iko yaṇaci, इको यणचि) contains words ikaḥ (ṣaṣṭhī), yaṇ (prathamā), aci (saptamī), which is to be interpreted as "an इक् letter which is followed by an अच् letter is converted to a corresponding यण् letter".

Fifth, the rules of Sanskrit grammar, as we understand them today, are not always strictly adhered to in ancient texts like the Rāmāyaṇa. This non-conformity adds an intriguing dimension to the language and showcases the fluidity and evolution of Sanskrit over time. For instance, in contemporary Sanskrit, following the rules of Pāṇini, it is customary to use the dative case (caturthī) in combination with the verb ruc (simple present tense rocate). However, in the Rāmāyaṇa, we come across instances such as 'na rocate mamāpyetadārye'[3] where the genitive case (ṣaṣṭhī) is employed instead of the dative. There are numerous other examples where the Pāṇini's expectations of grammatical cases are violated. These deviations from grammatical norms pose formidable obstacles for NLP systems, which rely on consistent patterns and structures to analyse and process text.

## 1.3 Manual Annotation

*Annotation* is a process of marking, highlighting or extracting relevant information from a corpus. It is important in various fields of computer science. A generic application of an annotation procedure is in creating a dataset that can be used as a training or testing set for various machine learning tasks. The exact nature of the annotation process can vary widely based on the targeted task, though.

Manual annotation plays an important role in NLP. It is particularly important in the context of low-resource languages for the creation of datasets. In the context of NLP, annotation often refers to identifying and highlighting various parts of the sentence (e.g., characters, words or phrases) along with syntactic or semantic information. There are a number of syntactic and semantic tasks in NLP which can utilize annotation by domain experts. Lemmatization, morphological analysis, parts-of-speech tagging, named entity recognition, dependency parsing, constituency parsing, co-reference resolution, sentiment detection, discourse analysis etc. are some examples of such common NLP tasks. Semantic tasks are high-level tasks dealing with the meaning of linguistic units and are considered among the most difficult tasks in NLP for any language. The task of question answering is a prominent example of semantic tasks. It requires a machine to 'understand' the language, i.e., identify the intent of the question, and then search for relevant information in the available text. This often encompasses other NLP tasks such as parts-of-speech tagging, named entity recognition, co-reference resolution, and dependency parsing [Jurafsky, 2000].

[3]Rearranging as per the anvaya order, we obtain, 'ārye, etat mama api na rocate', where, a genitive case form of 'asmad', 'mama', is used instead of a dative case form 'mahyam'.

Utilizing knowledge bases is a common approach for the QA task [Voorhees, 1999, Hirschman and Gaizauskas, 2001, Kiyota et al., 2002, Yih et al., 2015]. Construction of knowledge graphs (KGs) from free-form text, however, can be very challenging, even for English. The situation for other languages, whose state-of-the-art in NLP is not as advanced in English, is worse. As an example, consider the epic *Mahabharata*[4] in Sanskrit. The state-of-the-art in Sanskrit NLP is, unfortunately, not advanced enough to identify the entities in the text and their inter-relationships. Thus, human annotation is currently the only way of constructing a KG from it.

Even a literal sentence-to-sentence translation of Mahabharata in English, which probably boasts of the best state-of-the-art in NLP, is not good enough. Consider, for example, the following sentence from (an English translation of) "The Mahabharata" [Ganguli et al., 1884]:

> *Ugrasrava*, the son of *Lomaharshana*, surnamed *Sauti*, well-versed in the *Puranas*, bending with humility, one day approached the great sages of rigid vows, sitting at their ease, who had attended the twelve years' sacrifice of *Saunaka*, surnamed *Kulapati*, in the forest of *Naimisha*.

The above sentence contains numerous entities, e.g., *Ugrasrava*, *Lomaharshana*,

[4] Mahabharata is one of the two epics in India (the other being Ramayana) and is probably the largest book in any literature, containing nearly 1,00,000 sentences. It was originally composed in Sanskrit.

| Original text | | Canonical word order |
|---|---|---|
| [ na rocate mama-api-etad-ārye $]_1$ [ yad-rāghavo vanam /<br>tyaktvā rājyaśriyaṃ gacchet $]_2$ [ striyā vākyavaśaṃ gataḥ // 2<br>viparītaś ca vṛddhaś ca viṣayaiś ca pradharṣitaḥ /<br>nṛpaḥ kim iva na brūyāc codyamānaḥ samanmathaḥ // 3 $]_3$<br>[ ... ] | ➡ | [ ārye etad mama api na rocate $]_1$<br>[ yad rāghavo rājyaśriyaṃ tyaktvā vanam gacchet $]_2$<br>[ viparītaḥ vṛddhaḥ ca viṣayaiḥ pradharṣitaḥ ca codyamānaḥ<br>samanmathaḥ ca striyā vākyavaśaṃ gataḥ nṛpaḥ kim iva na brūyāt $]_3$<br>[ ... ] |

**Figure 1.1:** Sanskrit verses from Vālmīki Rāmāyaṇa using IAST transliteration scheme. Original text appears on the left with sentence boundary markers added. The canonical word order is shown on the right.

as well as multiple relationships, e.g., *Ugrasrava is-son-of Lomaharshana*. One of the required tasks in building a KG for Mahabharata is to extract these entities and relationships.

Even a state-of-the-art tool such as *spaCy* [Honnibal et al., 2020] makes numerous mistakes in identifying the entities; it misses out on *Ugrasrava* and identifies types wrongly of several entities, e.g., *Lomaharshana* is identified as an *Organization* instead of a *Person*, and *Saunaka* as a *Location* instead of a *Person*.[5] Consequently, relationships identified are also erroneous. This highlights the difficulty of the task for machines and substantiates the need for human annotation.

Manual annotation of text is a prime necessity for a low-resource language such as Sanskrit. Further, most of classical Sanskrit literature is in poetry form following mostly free word order [Kulkarni et al., 2015], without any punctuation marks. Therefore, there are certain specialized tasks needed for Sanskrit text processing, such as sentence boundary detection and canonical word ordering[6].

Consider an example from Vālmīki Rāmāyaṇa [Dutt et al., 1891] shown in Figure 1.1. The sentence boundaries are denoted using square brackets ([ and ]), and the verse boundaries are marked by two forward slashes (//). It can be observed that the sentence boundaries do not coincide with the verse boundaries. In particular, there may be multiple sentences present in a single verse, or a sentence may extend across multiple verses. Further, it can be seen on the right side of the arrow that the canonical word order is different from the order in which words appear in the original text.

For such languages that either do not use punctuations or use them in a limited amount, sentence boundary detection is an important task. Additionally, in languages with relatively free word order, decision of a canonical word order is also relevant. These two tasks also play a vital role when dealing with the corpora in the form of poetry, making them potentially relevant for all languages.

[5]This is not a criticism of spaCy; rather, it highlights the hardness of semantic tasks such as NER.
[6]The order which most effectively conveys the intended meaning of a sentence to the reader.

It is often required that multiple annotation tasks be performed on the same corpus. The order in which these tasks are performed can also be relevant due to interdependence of the tasks. Specifically, whenever[7] the task of sentence boundary detection is relevant, it needs to be performed first before any other annotation task. For example, one cannot decide the word order of a sentence before first finalizing the constituent words of a sentence. Same is the case for tasks such as dependency parsing, sentence classification, discourse analysis, and so on.

An annotation tool is crucial for the successful completion of any annotation task, and its success relies heavily on its user-friendliness for the annotators. Apart from this, the tool should be easy to install and should support web deployment for distributed annotation, allowing multiple annotators to work on the same task from different locations. The administration interface of the tool should also be intuitive and should provide easy access to common administrative tasks such as corpus upload, ontology creation, and user access management.

The features required from an annotation tool are also dependent on the type of task it is being used for. For example, for the purpose of knowledge graph focused annotation, it is important to have capabilities for multi-label annotations and support for annotating relationships. For linguistic tasks, often there is a need for multiple annotation tasks to be completed on the same corpus. Further, the tool should support sequential annotation, which is necessary for a set of annotation tasks that involve sentence boundary detection. A well-designed annotation tool should possess all these features to ensure that the annotation process is smooth, efficient, and accurate.

[7]corpora without clear sentence boundaries, such as languages with limited punctuation or corpora containing poetry

## 1.4 Related Work

The thesis explores various aspects within the broader theme of computational methods and tools for Sanskrit knowledge-based systems. These areas encompass construction of knowledge-graphs, question answering, manual annotation, meter identification for corpus correction, and software development. As such, each chapter offers a detailed survey of the related work specific to its subject. We now introduce some literature that holds a central place in the area of Sanskrit computational linguistics.

Within the realm of Sanskrit corpora, various tools and resources have emerged, including those related to corpus management [Huet, 2020, Goyal and Huet, 2016, Huet and Lankri, 2020], dictionaries like Amarakośa [Nair and Kulkarni, 2010], WordNet [Kulkarni et al., 2010], and CDSL [cds, 2022], and corpora such as the Digital Corpus of Sanskrit (DCS) [Hellwig, 2021] and GRETIL [gre, 2023].

Given the intricate morphological complexity of Sanskrit, the morphological analysis of words is a prominent task. Some noteworthy toolkits have been developed to address this challenge, including *Samsaadhanii* [Kulkarni, 2016] and *The Sanskrit Heritage Platform* [Goyal et al., 2012, Huet, 2005, Huet, 2020, Huet and Lankri, 2020]. Among the key tasks essential for Sanskrit text processing, the resolution of sandhi and samāsa holds paramount significance. In the literature, these tasks are often collectively handled as a unified 'word segmentation' task [Hellwig and Nehrdich, 2018, Krishna et al., 2016]. There also have been efforts to transform a verse into prose (anvaya) ([Vikram and Kulkarni, 2020, Krishna et al., 2019]).

Explorations in the field of dependency parsing have encompassed the development of tagsets [Kulkarni, 2020, Kulkarni et al., 2020] and parsers [Kulkarni, 2021, Kulkarni et al., 2019, Krishna et al., 2020a, Sandhan et al., 2021].

In recent times, there has been a surge of research endeavors focusing on neural approaches to address a range of linguistic tasks, including compound identification [Sandhan et al., 2019, Sandhan et al., 2022], and the development of architec-tures for diverse joint prediction tasks [Krishna et al., 2020b, Krishna et al., 2021].

## 1.5 Objectives

Knowledge-based systems in Sanskrit can aid in the preservation and dissemination of the vast amount of knowledge and literature that exists in the language. Many of these texts contain valuable information on subjects such as philosophy, science, medicine, and linguistics, among others. Accessing and utilizing the vast knowledge and literature present in Sanskrit texts poses significant challenges, primarily due to the language's complexity and the scarcity of suitable datasets, tools and technologies. This thesis strives to contribute to the advancement of knowledge-based systems in Sanskrit, promoting the dissemination of knowledge for the benefit of researchers, scholars, and the wider community.

### 1.5.1 Question Answering System for Sanskrit

The first objective of this thesis is to develop a Question Answering (QA) system specifically designed for the Sanskrit language. This system will utilize the knowledge graph constructed from Sanskrit texts to provide relevant answers to natural language questions posed by users. By leveraging automated methods for knowledge graph construction, the QA system will enable users to interactively engage with the wealth of Sanskrit knowledge.

### 1.5.2 Intuitive and Accessible Annotation Tools

The second objective is to design and implement task-specific annotation tools that facilitate the creation and enrichment of knowledge graphs from Sanskrit texts. These annotation tools will enable researchers and linguists to annotate entities, relationships, and other task-specific linguistic features within the Sanskrit texts. By providing intuitive and customizable interfaces, the annotation tools will simplify the otherwise tedious process of manual annotation, thus accelerating the progress of natural language processing in Sanskrit. The tools will be designed to be accessible to domain experts without requiring any programming knowledge, allowing them to use the tools effectively regardless of their background in Computer Science.

### 1.5.3 User-Centric Solutions for Sanskrit Awareness and Research

The third objective is to develop user-friendly interfaces that promote Sanskrit awareness and facilitate research activities. These interfaces will serve as gateways to Sanskrit knowledge, allowing users to explore and access various resources, including the knowledge graph, Sanskrit dictionaries, ancient numeral systems, Sanskrit grammar and Sanskrit prosody. By providing accessible and interactive interfaces, we aim to enhance the engagement and participation of users in Sanskrit research and learning endeavors.

## 1.6 Contributions

In this section, we will delve into the primary contributions of this thesis, highlighting the significant advancements and insights it brings forth in the field. The contributions have been consolidated and made accessible through the online platform at `https://sanskrit.iitk.ac.in/`.

- **Sanskrit Question Answering Framework**
  We present a framework for automated construction of knowledge graphs that can answer domain specific factual questions in Sanskrit. By leveraging the vast and varied literature in Sanskrit, including texts such as Mahābhārata and Rāmāyaṇa, we construct a knowledge graph specific to kinship relationships found in these texts. We also explore automatic extraction of certain relationships from Bhāvaprakāśanighaṇṭu, an Āyurveda text. Our natural language question answering system utilizes this knowledge graph to answer factual questions, achieving a success rate of approximately 50%. We also conduct a detailed analysis of the system's limitations at each step and explore poten-tial avenues for improvement. We identify that the state-of-the-art in Sanskrit natural language processing is limited by the lack of suitable datasets for the development and evaluation of such systems.

- **Annotation Tools**

  We emphasize the importance of human annotation in the creation of these datasets and the need for task-specific and intuitive annotation tools. We introduce two software tools to fulfill this need.

  – ***Sangrahaka***

    *Sangrahaka* is an annotator-friendly, web-based tool for ontology-driven annotation of entities and relationships that can be used for knowledge graph construction. It also supports querying using natural language query templates. Additionally, it has an interfaces for queries in the form of graphs and a graph browser interface for freely exploring the knowledge graph.

  – ***Antarlekhaka***

    *Antarlekhaka* is a versatile general-purpose multi-task annotation system that supports the manual annotation of a comprehensive set of NLP tasks. This system allows users to annotate small units of text with multiple categories of NLP tasks in a sequential manner. The system not only addresses the standard NLP tasks but also provides support for specific tasks such as identifying sentence boundaries and establishing canonical word order, making it especially useful for Sanskrit and other poetic corpora. It supports a total of eight generic categories of annotation tasks, amounting to the support for a much larger set of NLP tasks. Each category of tasks has a unique user-friendly and intuitive interface for annotation, making the tedious task of annotation much more accessible.

Both *Sangrahaka* and *Antarlekhaka* are full-stack web-based software with annotator-friendly interfaces that are easily configurable, web-deployable, and have a multi-tier permission system. Both tools have received mostly positive reviews in subjective evaluation and outperform other annotation tools in objective evaluation.

- **Knowledge Graphs and Datasets**
  - Knowledge Graph on Bhāvaprakāśanighaṇṭu

    We, through collaboration with Āyurveda experts, have created a rich and extensive ontology suitable for the annotation of Bhāvaprakāśanighaṇṭu, an Āyurveda text detailing medicinal substances, their properties and medical applications. It consists of 300 node labels and 320 relationship labels. We use this ontology and a custom deployment of *Sangrahaka* to perform manual annotation on Bhāvaprakāśanighaṇṭu and subsequently construct a knowledge graph (KG), specifically focusing on the three chapters from Bhāvaprakāśanighaṇṭu, namely, Dhānyavarga, Śākavarga and Māṃsavarga. The constructed knowledge graph contains 1606 entities and 1707 relationships, capturing the semantics of entity and relationship types present in the text. To facilitate querying the knowledge graph, we design 31 query templates that cover common question patterns.
  - Task-specific NLP Datasets on Vālmīki Rāmāyaṇa

    We have undertaken a large-scale annotation project using *Antarlekhaka* to annotate the Sanskrit corpus of Vālmīki Rāmāyaṇa, targeting various NLP tasks. In this project, we have focused on five tasks relevant to Sanskrit NLP: sentence boundary detection, canonical word ordering (anvaya), named entity recognition (NER), and co-reference resolution. With the support of 26 annotators, we have made significant progress in annotating the corpus. So far, we have annotated 883 verses out of the total 18754 verses, completing a total of 3532 annotation tasks. As a result of this endeavor, we have generated valuable datasets that encompass dif-

ferent aspects of the text. This includes 1928 sentence boundary annotations and 1847 canonical word ordering annotations. We have also developed a rich ontology consisting of 89 categories for NER, and collected annotations adhering to this ontology, resulting in the identification and classification of 1644 named entities. Additionally, we have established 2226 co-reference connections across 927 verses. We have shortlisted 44 types of relations relevant for action graphs, which are sentence level graphs capturing various actions. Based on the annotations adhering to these relation types, we have also collected 29 action graphs consisting of 250 relations.

- **Chandojñānam: Sanskrit Meter Identification and Utilization**

  Sanskrit prosody, the study of the structure and rules of Sanskrit poetry, can be used for digitization by helping to identify and correct errors in scanned or digitized texts. We present Chandojñānam, a Sanskrit meter identification and utilization system capable of identify meters from text as well as images. It can also provide approximate and close matches in the case of erroneous texts opening up the scope for a correction of erroneous digital corpora.

- **Miscellaneous Computational Tools and Interfaces**

  In addition to these software tools, the thesis presents a collection of web-interfaces, tools, and software libraries related to computational aspects of Sanskrit.

  – **Jñānasaṅgrahaḥ**

    Jñānasaṅgrahaḥ is a web-based collection of several computational applications related to the Sanskrit language. The aim is to highlight the features of Sanskrit language in a way that is approachable for an enthusiastic user, even if she has a limited Sanskrit background. The applications part of Jñānasaṅgrahaḥ are as follows.

    * **Saṅkhyāpaddhatiḥ**

In the ancient India, it was a common practice to represent numeric values using letters, syllables or words from a natural language. The primary reason to use such systems is, ease of remembrance of numbers. We present a user-friendly web-based interface, Saṅkhyāpaddhatiḥ, which implements three such ancient Indian numeral systems, Kaṭapayādi Saṅkhyā, Āryabhaṭīya Saṅkhyā and Bhūtasaṅkhyā. The core interface for each of the system consists of an encoding interface to encode numeric values into a valid text representation a decoding interface to decode any valid text representation into the corresponding numeric value.

* **Varṇajñānam**

We have developed an interface and a library of utility functions related to varṇa (phonetic unit of Sanskrit language) information and manipulation.

– **Vaiyyākaraṇaḥ**

Vaiyyākaraṇaḥ is a Telegram[8] bot that offers various tools for a Sanskrit learner including stem finder, root finder, declension generator, conjugation generator, and compound word splitter by making use of extant Sanskrit linguistic tools. This tool is being used by many Sanskrit learners and enthusiasts.

– **Python Libraries**

To facilitate the processing of Sanskrit text and corpora for programmers, we have developed a set of Python packages that are available on the Python Package Index (PyPI[9]). These packages can be installed using the '`pip install`' command.

* **PyCDSL**

*PyCDSL* is a Python library that provides programmer friendly inter-face to Cologne Digital Sanskrit Dictionaries (CDSD) [cds, 2022]. The library serves as a corpus management tool to download, update and access dictionaries from CDSD. The tool provides a command line interface (CLI) for ease of search and a programmable interface for using CDSD in computational linguistic projects written in Python 3.

[8] `https://telegram.org/`
[9] `https://pypi.org/`

* **Heritage.py**

  *Heritage.py* is a Python package that serves as an interface to The Sanskrit Heritage Site[10] [Goyal et al., 2012]. It provides a number of features for working with Sanskrit, including morphological analysis, sandhi formation, declensions and conjugations.

* **sanskrit-text**

  *sanskrit-text* is a Python package that provides a variety of utility functions for working with Sanskrit text in Devanagari script. It includes functions for syllabification, varṇa viccheda (breaking down words into their constituent sounds), pratyāhāra encoding-decoding, and uccāraṇa sthāna yatna (detailed information about the pronunciation of a word).

## 1.7 Significance

The significance of this research lies in its contributions towards the development of computational tools and resources for Sanskrit language processing. While Sanskrit is a central theme of the thesis, it's important to note that the annotation tools created are language-agnostic in nature, making them valuable for a wide array of languages.

The thesis addresses a critical challenge in the field of natural language processing, which is the lack of suitable datasets for developing and evaluating systems for low-resource languages such as Sanskrit.

The annotated datasets created in this research can be used for developing method-ologies to perform various computational linguistic tasks in Sanskrit. Additionally, this research has the potential to increase enthusiasm in Sanskrit as it presents a user-friendly and accessible approach to learning and utilizing the language through various software tools and interfaces.

[10] https://sanskrit.inria.fr/DICO/index.en.html

Overall, the research presented in this thesis has the potential to contribute to the development of language technologies for low-resource languages, promote the study and preservation of Sanskrit language, and advance our understanding of linguistics and literature.

## 1.8 Limitations

As with any research, there are limitations to this work that should be acknowledged. Some of the limitations of this research include:

- One limitation of the research is the absence of neural network-based approaches. Although the proposed framework and tools are effective in their respective tasks, the use of neural networks could potentially improve the efficiency of the system. However, the performance of the neural network is unpredictable and building explainable models is a research area in itself. Further, due to the lack of suitable training data, the implementation of such approaches in Sanskrit NLP is currently limited for high-level tasks such as Question Answering.

- There is a significant amount of research focused on low-level NLP tasks such as word segmentation, part-of-speech tagging, and dependency parsing. These tasks are considered foundational in natural language processing and are critical for building more advanced systems that can perform tasks like machine translation, text classification, and question answering. While this particular research may not directly address these low-level tasks, it does contribute to the field of Sanskrit natural language processing by developing new tools and techniques for creating datasets and knowledge graphs through manual anno-tation. These contributions can potentially serve as building blocks for future work on more advanced NLP tasks in Sanskrit and other languages.

- The integration of state-of-the-art NLP tools with annotation tools is a complex task, primarily due to the diverse specifications and the ever-evolving nature of such tools. While our current annotation tools do not include this capability, our plan is to introduce this support. Furthermore, we aim to establish specifications that will enable future tools to seamlessly conform to our framework, allowing them to be easily integrated as pluggable components.

## 1.9 Outline

In this section, we will give a brief overview of the organization of this thesis and provide a concise summary of the contents of each chapter.

- In **Chapter 1**, we introduce and motivate the primary problem addressed in this thesis. We clearly state the research objectives and the contributions, also touching upon the significance as well as limitations of this research.

- In **Chapter 2**, we propose a knowledge graph-based framework for automatically building a question answering system in Sanskrit. We also explore the limitations of such a system and provide a detailed error analysis. We recognize the limitation of the state-of-the-art in Sanskrit natural language processing due to the absence of appropriate datasets for system development and evaluation. As a result, we emphasize the necessity of manual annotation.

- In **Chapter 3**, we introduce *Sangrahaka*, a web-based tool designed for the annotation of entities and relationships in text corpora. *Sangrahaka* not only facilitates the construction of knowledge graphs through annotation but also supports querying using templatized natural language questions. The tool is language and corpus agnostic, allowing customization to meet specific re-quirements. We also describe our efforts in the ontology-driven manual annotation of Bhāvaprakāśanighaṇṭu, an Āyurveda text.

- In **Chapter 4**, we introduce *Antarlekhaka*, a comprehensive tool for manual annotation of various NLP tasks. The tool is Unicode compatible, language-agnostic, and supports distributed annotation. It includes user-friendly interfaces for a wide range of annotation tasks, including two novel tasks: sentence boundary detection and deciding canonical word order, which are particularly important for analyzing poetic texts. We also present the task-specific datasets created through manual annotation of several chapters from Vālmīki Rāmāyaṇa using *Antarlekhaka*, which provide valuable resources for training and evaluating machine learning models in low-resource languages.

- In **Chapter 5**, our focus is on Chandojñānam, a Sanskrit meter identification and utilization system. The tool incorporates computational techniques and Sanskrit prosody to identify the meter of a given Sanskrit verse. We highlight the error-tolerance of Chandojñānam owing to its capability to identify close and approximate matches based on sequence matching, thereby showcasing the potential for use in error correction of digital corpora.

- In **Chapter 6**, we describe a collection of innovative and user-friendly computational applications related to the Sanskrit language. These tools contribute to the field by offering practical solutions and insights for enthusiasts and learners with varying levels of Sanskrit proficiency. These tools include Jñānasaṅgrahaḥ, a collection of interesting Sanskrit related web-interfaces, Vaiyyākaraṇaḥ, a Telegram bot for learners of Sanskrit grammar and Python packages such as *PyCDSL*, *Heritage.py* and *sanskrit-text*.

- In **Chapter 7**, we make concluding remarks including summary of the primary contributions, future work and the research directions enabled by this work.

# Chapter 2

# Sanskrit Question Answering Framework

Extracting the knowledge from Sanskrit texts is a challenging task due to multiple reasons including complexity of the language and paucity of standard natural language processing tools. In this chapter, we target the problem of building domain-specific knowledge graphs from Sanskrit texts. We build a natural language question answering system in Sanskrit that uses the knowledge graph to answer factual questions. We design a framework for the overall system and implement two separate instances of the system on human relationships from Mahābhārata and Rāmāyaṇa, and one instance on synonymous relationships from Bhāvaprakāśanighaṇṭu, a technical text from Āyurveda. We show that about 50% of the factual questions can be answered correctly by the system. More importantly, we analyse the shortcomings of the system in detail for each step, and discuss the possible ways forward.

## 2.1 Introduction

We aim to take the first step towards a concrete NLP task, namely, natural language question answering in Sanskrit. In particular, we aim to design a *framework* that processes Sanskrit texts, extracts the information in it, and stores it in a format that can be queried using questions posed in Sanskrit.

We propose to store the knowledge base (KB) in a knowledge graph (KG) format. KGs have a rich structure and store the information in the form of entities (as nodes) and relationships (as edges between the nodes). The edges are directed, and both the nodes and edges can store labels describing their attributes. There are multiple off-the-shelf tools available for storing and querying KGs, including graph databases[1], Property Graphs[2], Resource Description Framework (RDF) [Lassila et al., 1998], Gremlin queries[3], SPARQL queries[4], etc. The popularity of knowledge bases such as YAGO [Suchanek et al., 2007], DBpedia [Auer et al., 2007] and Freebase [Bollacker et al., 2008] is a testament to their success.

We also propose question answering as a concrete example of the use of such KGs and a way of measuring the effectiveness of the system. Various online question answering fora such as Quora[5] and quizzes serve as a motivation. We particularly choose the two epics of India, namely, Mahābhārata and Rāmāyaṇa, categorized as Itihāsa in Sanskrit literature, and questions on human relationships within them, as examples for our framework due to their popularity and ease of establishment of the ground truth. We also work with Bhāvaprakāśanighaṇṭu to highlight the usage for technical texts.

A model for extracting implicit knowledge from Amarakoṣa and storing it in a structured manner, and a tool for answering queries using this knowledge. was proposed by [Nair and Kulkarni, 2010]. Sanskrit WordNet[6] [Kulkarni et al., 2010] was built by expanding the Hindi WordNet. A production grammar for kinship terminology in Sanskrit was proposed by [Bhargava and Lambek, 1992], which explores the suffixes and morphological and clues in the formation of relationship words. While it provides an insight in the nomenclature of such terms, in order to be applicable towards the identification of the relationships between entities in text, a performant suffix analyser is required. Automatic translation tools, if available, can also be used

[1] https://en.wikipedia.org/wiki/Graph_database
[2] https://en.wikipedia.org/wiki/Graph_database#Labeled-property_graph
[3] https://docs.janusgraph.org/latest/gremlin.html
[4] https://www.w3.org/TR/rdf-sparql-query/
[5] https://www.quora.com
[6] http://www.cfilt.iitb.ac.in/wordnet/webswn/english_version.php

where the entire text is translated to English and the KG is built from the translated text. However, we could not find any such tools. Although Sanskrit-English dictionaries[7] provide a word-level translation of selected words from Sanskrit to English, word-level translation often does not produce meaningful or grammatically correct text. We, thus, decided to use only the text as available in Sanskrit. In future, we will explore the use of such tools and methods.

The rest of the chapter is organized as follows. In Section 2.2, we explain the generic framework of the question answering system. There exist some excellent tools for Sanskrit that aid us in the analysis. For other cases, we build our own heuristic rule-based systems. In Section 2.3, we describe the automatic construction of the knowledge graph while the details of the various modules of the system are described in Section 2.4. Since Bhāvaprakāśanighaṇṭu is a technical text, we highlight its specialized processing in Section 2.5. In Section 2.6, we analyse the results of our experiments. Finally, in Section 2.7, we discuss the lessons learnt and future directions.

## 2.2 Proposed Framework

### 2.2.1 Knowledge Graphs (KG)

Knowledge graphs (KG) model real-world entities as nodes. Relationships among the entities are modelled as (directed) edges. For example, in a KG about human relationships in Mahābhārata, arjuna and abhimanyu are nodes. They are connected by a directed edge from arjuna to abhimanyu labelled by the relationship “has-son” (putra).

In English, there have been several efforts in automated KG construction, notable among them being YAGO, DBpedia, Freebase, etc. YAGO ontology [Suchanek et al., 2007] was built by crawling the Wikipedia and uniting it with WordNet using a combination of both rule-based as well as heuristic methods. DBpedia [Auer et al., 2007] extracts knowledge present in a structured form on Wikipedia by template detection using pattern matching coupled with post-processing for quality improvement. Freebase [Bollacker et al., 2008] is a database of tuples that is created, edited and maintained in a collaborative manner. Unfortunately, however, none of the above techniques are applicable for automatically building knowledge graphs in Sanskrit.

[7] `https://www.sanskrit-lexicon.uni-koeln.de/`

Processing of text for YAGO depends on many Information Retrieval (IR) and Natural Language Processing (NLP) tools that are available only in English and a handful of other languages, mostly European. The state of the art of these tools in Sanskrit is still not standardized and may not be directly useful. Sanskrit Wikipedia[8] also is not as resourceful as its counterpart in English. Hence, the amount of structured information available there is minuscule compared to the vast Sanskrit literature that is developed over several millennia. Thus, a system such as DBpedia is not possible. A collaborative effort such as Freebase is also ruled out due to a paucity of active Sanskrit users adept in digital technologies. To the best of our knowledge, there is no work that directly builds a knowledge graph from Sanskrit texts.

### 2.2.2 Triplets

A common way of encoding the relationship information is in the form of *semantic triplets*. A triplet has the structure `[subject, predicate, object]` which indicates that the entity `subject` has the relationship `predicate` with the entity `object`. Hence, the fact that arjuna has a son abhimanyu is encoded as the triplet [arjuna, has-son (putra), abhimanyu] ([अर्जुन, पुत्र, अभिमन्यु]).

The KG is built automatically by extracting such triplets from the text. We target KGs on specific types of relationships, namely, human relationships for epics, and synonymous relationships in nighaṇṭu. One of the foremost jobs, therefore, is to identify the relationship words. This is a corpus-independent set and depends only on the language. However, since the text is free-flowing (except in technical texts where there is a structure) and almost always written in poetry in the form of śloka, even when a relationship word is identified, the subject and object words may be anywhere around it (both before and after). Śloka (श्लोक) is a semantic unit in Sanskrit and is equivalent to a *verse*. Sometimes, one or both of these entities may not be even in the same śloka. Hence, a *context* window around the relationship word must be defined and searched for the relevant entities. Specifying the length of such a context window is not easy; if it is too short, relationships may be missed, while if it is too long, too many spurious relationships may be inferred. Even identifying the śloka boundaries may not always be trivial. Fortunately, however, these boundaries are clearly marked in the texts that we have worked on.

[8] `https://sa.wikipedia.org/wiki`

The details of how such triplets are extracted are explained in Section 2.3. The knowledge graph is maintained in an RDF format as a set of all such extracted triplets.

### 2.2.3 Questions

The next important task in the pipeline is to parse the natural language question. Since the question is also in Sanskrit, we adopt similar processing as the text to extract triplets. In this work, we assume only factual based questions such as "Who is the son of arjuna?" (अर्जुनस्य पुत्रः कः?) The triplet extracted from the above question will be [arjuna, has-son, X] ([अर्जुन, पुत्र, किम्]).

Since Sanskrit is quite free with word ordering, the above question may be asked in different manners, such as अर्जुनस्य पुत्रः कः? or कः अर्जुनस्य पुत्रः? or अर्जुनस्य कः पुत्रः? All of these should yield the same triplet [अर्जुन, पुत्र, किम्].

The *inverse* question may also be asked: "Who is the father of abhimanyu?" (कः अभिमन्योः पिता?) The above can be answered only if it is known that the inverse of "has-father" is the relationship "has-son". This, again, is a property of the language and must be explicitly mentioned.

Hence, we maintain a map of such inverse relationship rules. Note that it is not always one-to-one. For example, "has-mother" is also the inverse of "has-son", and "has-father" is the inverse of "has-daughter" as well. Gender information, therefore, becomes important.

We augment the initially built knowledge graph by adding appropriate inverse relationship edges. It is ensured that an inferred inverse relationship does not contradict a directly inferred relationship from the text. The details are in Section 2.3.4.

Even though the questions are simple and short, they may contain *multiple* triplets. For example, a question पाण्डोः पत्न्याः भ्राता कः? may be asked by someone who does not know what the relation brother-of-wife is called in Sanskrit. This question contains two relationships, पत्नी and भ्राता. The triplet form of these relationships would be [पाण्डु, पत्नी, किम्] corresponding to the subquestion 'Who was the wife of pāṇḍu?' and [पत्नी, भ्राता, किम्] corresponding to the subquestion 'Who was the brother of wife (of pāṇḍu)?'. All of these must be extracted correctly.

Further, they must be linked properly. In the example above, we must ensure that the object of the first triplet is the subject of the second triplet, that is, the correct triplets are [पाण्डु, पत्नी, X] and [X, भ्राता, किम्]. Here, a variable is used to denote the person that satisfies both the triplets.

Once these are correctly linked, a SPARQL query pattern is formed. The SPARQL query equivalent for the above question is

```
SELECT ?A
WHERE {
        :पाण्डु :पत्नी ?X .
        ?X :भ्रातृ ?A .
}
```

This is finally directly queried against the KG, and the answer is returned. Section 2.4 describes in detail the intricacies of the different steps of the question answering system.

Figure 2.1 describes the overall framework. The final accuracy of the system is dependent on each of the modules of the architecture. For example, if the extracting triplets component is very erroneous, then neither the KG information is captured correctly, nor is the intention of the question understood. The overall error is a cascading effect of the errors in each of the individual components. Thus, for a successful system, each component must be reasonably accurate.

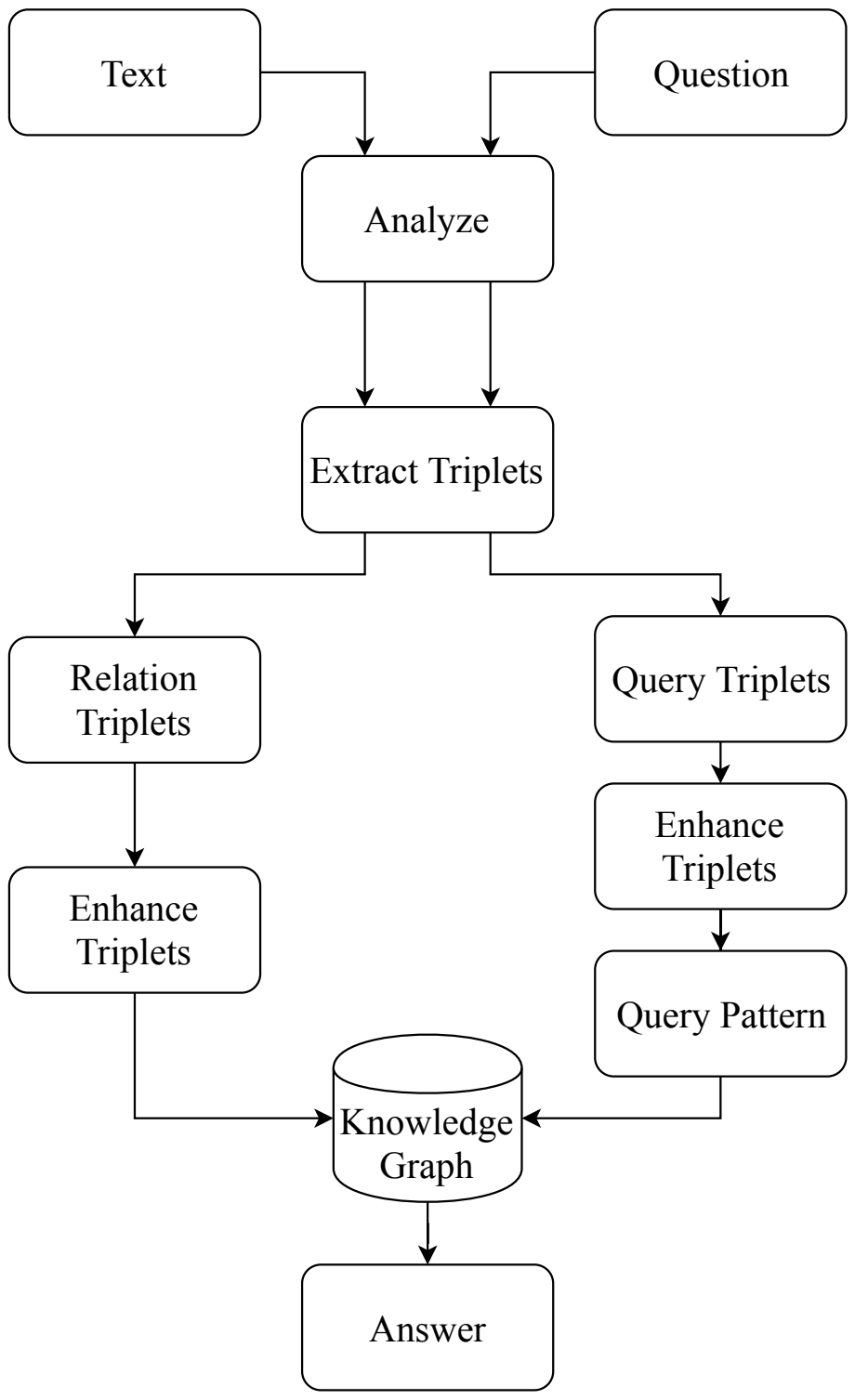

**Figure 2.1:** Overall framework of the system.

## 2.3 Construction of Knowledge Graph

In this section, we describe in detail the automated construction of knowledge graph (KG). The input consists of Sanskrit text (in digital Unicode format) of an entire work (such as Mahābhārata, Bhāvaprakāśanighaṇṭu, etc.) and the *type* of relationships intended (e.g., human relationships, synonymous words, etc.). The output is a set of triplets in the form [subject, predicate, object] where the predicate is of the relationship type intended and subject and object are entities. If `[a, R, b]` is an output triplet, then it implies that *object* `b` is *relation* `R` of *subject* `a`.

### 2.3.1 Pre-Processing of Text

Sanskrit is a morphologically rich language. A single noun root, called prātipadika (प्रातिपदिक), can yield many forms depending on the case, gender and number. Similarly, a single verb root, called dhātu (धातु), can lead to many forms as well depending on the tense, person and number. In addition, various prefixes (upasarga, उपसर्ग) and suffixes (pratyaya, प्रत्यय) get affixed to these forms to generate thousands of other forms.

Further, words are very often joined together to form compound words using either pronunciation rules through a process called sandhi (सन्धि) or semantic rules through a process called samāsa (समास). Often, both are invoked together, and a series of words are joined together to form one big compound word.

Splitting these compound words into their base words is a highly complicated procedure and may not always be unambiguous. For this step, we make use of the *Sanskrit Sandhi and Compound Splitter*, a tool[9] by [Hellwig and Nehrdich, 2018]. For example, if the input text is कर्णार्जुनयोः कश्श्रेष्ठः the output is कर्ण-अर्जुनयोः कः श्रेष्ठः.

The next task is to semantically analyse the *form* of the word. Again, we use a third-party *analyser tool*, *The Sanskrit Reader Companion*[10] from *The Sanskrit Heritage Platform* by [Goyal et al., 2012]. This tool outputs the case (vibhakti, विभक्ति), number (vacana, वचन) and gender (liṅga, लिङ्ग) for each word. The tool uses various abbreviations[11] to convey the linguistic information. The tool provides multiple potential analyses for each input. For the sake of automated processing, we opt for the first result.

For the running example, the analysis yields

कर्ण ['voc.', 'sg.', 'm.']

अर्जुन ['loc.', 'du.', 'm.']

किम् ['nom.', 'sg.', 'm.']

श्रेष्ठ ['nom.', 'sg.', 'm.']

Here, 'nom.', 'loc.' and 'voc.' are abbreviations used to denote nominative case (प्रथमा), locative case (सप्तमी) and vocative case (सम्बोधन) respectively. Similarly, 'sg.' and 'du.' indicate singular and dual number (एकवचन and द्विवचन). While 'm.' denotes the masculine gender (पुंलिङ्ग).

---

[9]`https://github.com/OliverHellwig/sanskrit/tree/master/papers/2018emnlp`

[10]`https://sanskrit.inria.fr/DICO/reader.fr.html`

[11]All the abbreviations used by the tool are listed at `https://sanskrit.inria.fr/abrevs.pdf`.

The word श्रेष्ठ gets correctly analysed: it is in the nominative case, is in singular number, and masculine gender. However, the other words require some more adjustments. For example, the word अर्जुन is shown to be in dual number. This is output since the original compound word consisted of two persons. However, now that they are separated, it should no longer be in dual number, but adjusted to be in singular number. Similarly, the case analysis for कर्ण is wrongly output to be vocative. The reason for this again is the fact that the original structure of the compound word was lost. We adjust the case of previous words in a compound word by adopting the case of the last word in the compound word. Thus, the case for कर्ण is changed to locative, since that is the case for अर्जुन.

### 2.3.2 Identifying Relationship Words

Given a particular relationship type, the set of words pertaining to it is a property of the language and is corpus-independent. For example, if human relationships are targeted, in Sanskrit, the (roots of the) relevant words are pitṛ (father, पितृ), mātṛ (mother, मातृ), putra (son, पुत्र), putrī (daughter, पुत्री), pati (husband, पति), patnī (wife, पत्नी), etc. Of course, these words can appear in various forms. More importantly, their synonyms can also appear. For example, the words दुहितृ (duhitṛ), तनया (tanayā), आत्मजा ātmajā are synonymous with पुत्री (putrī).

While these can be learned, since the set is mostly fixed, we have employed a key-value based approach where we have listed many of such relationship words along with their synonyms. For each such group of synonyms, there is a canonical word (e.g., पुत्री for the group of words indicating daughter) that is used in the KG.

The identification of a relationship word is simply a match from this entire set of words.

### 2.3.3 Identification of Triplets

Once a relationship word is identified, it forms the predicate of a triplet. The next task, therefore, is to identify the subject and object corresponding to it.

It is expected that the subject and object entities will not be too far off from the predicate word. To bound the sphere of influence or context, we use śloka (श्लोक) boundaries. Each śloka considered as a semantic unit and is akin to a verse. Fortunately, for the texts we have used, the śloka *boundaries* are clearly marked. In this work, we restrict the context to be *one* śloka before and after the one where the predicate is found, i.e., a total of 3 śloka.

Since subjects and objects are entities, they generally occur as nouns in a language. The analyser tool (*The Sanskrit Reader Companion*) described earlier marks the parts-of-speech tags of words. It, however, does not distinguish between nouns, pronouns and adjectives. Since there is a fixed set of pronouns for Sanskrit, we use that set to correct some of the nouns. We, however, fail to distinguish the adjectives from the nouns in a satisfactory and consistent manner. This is a major future work.

Within the nouns (and adjectives), we look for those that are in the *genitive* case (षष्ठी विभक्ति). The genitive case pertains to the ṣaṣṭhī vibhakti (genitive case) and denotes sambandha (सम्बन्ध). The word sambandha in Sanskrit literally means relationship and, therefore, a noun exhibiting genitive case is the most likely candidate for a subject. For example, the अर्जुनस्य पुत्रः अभिमन्युः आसीत् means abhimanyu was son *of* arjuna. Here, 'of arjuna' is expressed by the genitive case of the word (अर्जुन), i.e., अर्जुनस्य. Hence, all such nouns in the genitive case are marked as subjects.

The relationship word or the predicate can be in different cases, numbers and gender, though. Since the object follows the predicate, according to Sanskrit grammar, it must be in the same case, number and gender as the predicate. We use this rule to extract objects. To be precise, an object is a noun that exhibits the same case, number and gender as the predicate word. In the sentence अर्जुनस्य पुत्रः अभिमन्युः आसीत्, word पुत्रः is the predicate word and the word अभिमन्युः is the object and both of these words are in the nominative case (प्रथमा विभक्ति).

We insert all such extracted triplets in the KG. We assume that if an entity appears multiple times, it refers to the *same* person. The above assumption is almost always correct barring some exceptional cases.[12]

### 2.3.4 Enhancement of Relationships

As explained earlier (in Section 2.2), just the base relationships may not always be enough to answer a question. If the triplet [arjuna, has-son, abhimanyu] ([अर्जुन, पुत्र, अभिमन्यु]) is stored, the question “Who is the father of abhimanyu?” (कः अभिमन्योः पिता?) cannot be answered, even though the information is present.

To be able to answer such queries, we have enhanced the KG with inverse relationships. For example, the inverse of “has-father” is “has-son”. This, again, is a property of the language and are explicitly stored.

As discussed earlier, the inverse relationships are not always one-to-one. For example, “has-mother” is also the inverse of “has-son”, and “has-father” is the inverse of “has-daughter” as well. Hence, we use the gender information of the subject and the object to disambiguate.

The complication does not end here. Imagine a question “Who is maternal uncle of Nakula?” (नकुलस्य मातुलः कः). This information may not be directly stored in the KG. The relationship मातुल is a composition of मातृ and भ्रातृ. These components [नकुल, मातृ, माद्री] and [माद्री, भ्रातृ, शल्य] may be present in the KG. Again, the situation is that the KG contains the information but cannot answer the question.

To solve this, derived relations could be broken into their component base parts. Thus, “has-maternal uncle” is stored as “has-mother” and “has-brother” with an additional (possibly unnamed) node in between. In particular, from the triplet [नकुल, मातुल, शल्य], two more triplets [नकुल, मातृ, X] and [X, भ्रातृ, शल्य] could be generated. If there is already such a node X, it could be used; otherwise, a new node could be created. However, addition of such *dummy* nodes has not been explored in this work.

We achieve the same result by handling this issue at the time of querying. This is discussed in Section 2.4.2. We maintain a list of relationships and their possible derivations from base relationships. Once more this mapping is rarely one-to-one.

[12] karṇa was the son of kuntī, and one of the kaurava was also named karṇa.

For example, “brother-of” can be composed of “son-of-father” and “son-of-mother”. Also, the gender must be taken care of.

A particularly interesting case is “has-ancestor” and “has-descendant”. These are recursive relationships, and the depth of recursion can be anything, i.e., a ‘father’ is an ancestor, so is an ‘ancestor-of-father’, and so on. We do not handle these cases in the current work.

## 2.4 Question Answering

We now describe one application, that of question answering. We assume that the questions are asked directly in Sanskrit and are about facts, i.e., about a single piece of information. We also assume that the questions are only about the relationships that the knowledge graph encodes. If not, the question is ignored, since clearly the KG is incapable of answering it. Further, the questions are assumed to be short and consist of a single sentence only.

The question is first pre-processed in the same manner as the text (Section 2.3.1). To be more precise, compound words are split using *Sanskrit Sandhi and Compound Splitter* a tool by [Hellwig and Nehrdich, 2018], the component words are analysed using *The Sanskrit Reader Companion* from *The Sanskrit Heritage Site*, and relationship words and nouns are identified. Next, triplets are extracted.

### 2.4.1 Identifying Triplets

A blank triplet is initialized. The question words are scanned one by one. For each word, it is determined if it can be a subject word, a predicate word or an object word. If the word is a noun in genitive case but is not a relationship word, then it is likely to be a subject word. The relationship words directly give the predicates. The object word is generally in the nominative case. For example, consider the question अर्जुनस्य पुत्रः कः? (“Who is the son of arjuna?”). Since अर्जुन is in genitive case, it is the subject. The word पुत्र is the predicate. The object is किम्. The triplet formed, therefore, is

[अर्जुन, पुत्र, किम्].

Once a triplet is filled up, another new triplet is initialized. This is necessary since there may be chain questions of the form अर्जुनस्य पुत्रस्य पुत्रः कः? The triplets generated from this are [अर्जुन, पुत्र, X] and [X, पुत्र, किम्].

The process goes on till all the words in the question are processed.

At the end of this phase, the triplets thus formed are called *query triplets*.

### 2.4.2 Enhancing Triplets

Each query triplet is next enhanced to a set of triplets, called the *enhanced triplet set*. The rules for enhancing the relationship of a query triplet is the same as that used in processing the KG triplets. In particular, each complex relation is broken into its constituent parts and new triplets are created using the aforementioned mapping of relationships to its constituents.

Suppose, a predicate (i.e., relation) `R` can be decomposed to two base predicates `R1` and `R2`. Then, if a query triplet is of the form `[A, R, B]`, then two triplets of the form `[A, R1, X]` and `[X, R2, B]` are generated. Note that `{[A, R, B]}` and `{[A, R1, X], [X, R2, B]}` are *equivalent expressions* and either of them can return the correct answer from the KG. However, since it is not known which information is stored in the KG, *both* are used.

Thus, each query triplet $QT_i$ is replaced by its enhanced triplet set $ET_i = \{QT_i\} \cup IT_i^j$ where $IT_i^j$ is a set of triplets inferred from $QT_i$, as shown in the example below.

For the question अर्जुनस्य मातुलस्य पिता कः, we first obtain the triplets {[अर्जुन, मातुल, X], [X, पितृ, किम्]}. These triplets are then enhanced by appropriately splitting the relationship मातुल using the rule मातुल = मातृ + भ्रातृ. Here, $QT =$ [अर्जुन, मातुल, X] and $IT =$ {[अर्जुन, मातृ, Y], [Y, भ्रातृ, X]}. As a result, we get two triplet sequences for this question, {[अर्जुन, मातृ, Y], [Y, भ्रातृ, X], [X, पितृ, किम्]} and {[अर्जुन, मातुल, X], [X, पितृ, किम्]}.

### 2.4.3 Query Pattern

If the question contains only one query triplet, then members of its enhanced triplet set form the alternate query patterns. Suppose, however, the question contains $n$ query triplets with their corresponding $n$ enhanced triplet sets $ET_1, ET_2, \cdots, ET_n$. The Cartesian product of the elements of these sets form the *alternate query patterns*. Thus, if there are $2$ enhanced sets with $2$ and $3$ elements in them, the total number of alternate query patterns is $2 \times 3 = 6$.

Each of these alternate query patterns are posed to the KG and answer triplets are returned. The correct field of the answer triplet is returned as the factual answer.

We have not encountered a case where alternate query patterns return different answers. If, however, such a situation arises, a further disambiguation step (possibly using majority voting, etc.) is required.

## 2.5 Technical Texts

We have chosen a technical text Bhāvaprakāśa which is one of the important texts from Āyurveda. Bhāvaprakāśanighaṇṭu is a glossary chapter from this text, which contains detailed information about the medicinal properties of various plants, animals and minerals written in a śloka format. There are $23$ adhyāya in this chapter. Being a technical text, Bhāvaprakāśanighaṇṭu has more structure than Rāmāyaṇa or Mahābhārata.

### 2.5.1 Structure

The text Bhāvaprakāśanighaṇṭu loosely adheres to the following structure.

- Substances (dravya, द्रव्य) with similar properties or from the same class occur in the same chapter. For example, all the flowers are in one chapter, all the metals are in another chapter.

- Each chapter consists of various *blocks* (sets of consecutive śloka), where each block speaks about one substance.
- Each block generally has the following internal components:
  - Synonyms of the concerned substance
  - Where that substance can be found
  - Properties of the substance. e.g., colour, smell, texture, composition and other medicinal properties
  - Differences between the different varieties of the substance

While the blocks are structured to some extent, the following deviations exist.

- The length of each block is not fixed.
- The number of synonyms of each substance are not fixed.
- The order of the components of the block varies from substance to substance to a certain extent.
- Some of the internal components may, at times, be absent such as the varieties of a substance.

Importantly, the *separation* between two consecutive blocks is not marked in the text.

These points of deviation from the pattern act as hurdles in the process of understanding and exploiting the structure of a text to extract information. Understanding the structure of a text can be a challenging task. We have taken the help of domain experts[13] to form our understanding of the structure described above.

Properties (guṇa, गुण) are of the form `(name, value)`. A property value can be directly attached to a substance, or it can be attached through a `property-name`. For example, a substance is "red", or, a substance has *colour* "red".

[13]We acknowledge Dr. Sai Susarla, Dean at Maharshi Veda Vyas MIT School of Vedic Sciences, Pune, India, and his team for sharing their expertise with us.

There are various types of relationships that can be of interest. Examples of these relationships include (`substance-1`, `is-synonym-of`, `substance-2`), (`substance`, `property-name`, `property-value`), (`substance`, `has-property`, `property-value`), and (`substance`, `found-at`, `location`). There are cases where a `property-value` has different meanings in different contexts, i.e., for different `property-name` instances. For example, guru, as a size property refers to the size of a substance. Without such specification, in the context of Āyurveda, guru holds the meaning *hard to digest*. We use the relation `has-property` when the `property-value` is not associated with a specific `property-name` type.

We focused our efforts on a single relationship in the Bhāvaprakāśanighaṇṭu, namely, `is-synonym-of`. In other words, the triplets that we are interested in are of the form (`substance-1`, `is-synonym-of`, `substance-2`). Since the predicate is same for all triplets, we choose to get rid of it and think of the problem as simply *finding pairs of synonyms*.

This task is subdivided into two tasks, (1) finding śloka that contain the synonyms, and (2) given such a śloka, finding pairs of synonyms from it.

### 2.5.2 Property Words

The corpus is initially pre-processed in a similar manner as described in Section 2.3.1. However, a next layer of processing is done to extract more information.

The set of properties is a relatively small set of words. The names and values of these properties together are called *property words*. Since the *property words* recur heavily in every block that describes a substance, they are expected to have much higher frequencies than the names of substances. We test this hypothesis by performing a frequency analysis of the top words and nouns in the entire text.

Table 2.1 lists the top-10 most frequent words and nouns along with their frequencies. Notice that most frequent words also contain stopwords like च, तद् etc., while the list of nouns indicates that the standard property words such as वात, पित्त, कफ have a high frequency. Following this empirical evidence, we choose the top-50

**Table 2.1:** Top-10 most frequent words, nouns and their frequencies from Bhāvaprakāśanighaṇṭu.

**Words**

| Adhyāya 1 | Adhyāya 2 | All Adhyāya |
|---|---|---|
| (च, 127) | (च, 56) | (च, 946) |
| (तद्, 85) | (तिक्त, 39) | (तद्, 786) |
| (किम्, 55) | (लघु, 37) | (पित्त, 461) |
| (कफ, 53) | (कफ, 31) | (कफ, 438) |
| (उष्ण, 47) | (तु, 24) | (तु, 394) |
| (पित्त, 45) | (किम्, 24) | (लघु, 321) |
| (तु, 39) | (तद्, 22) | (वा, 278) |
| (तथा, 35) | (विष, 22) | (अपि, 268) |
| (अपि, 34) | (उष्ण, 21) | (किम्, 266) |
| (तिक्त, 34) | (हृत्, 20) | (गुरु, 254) |

**Nouns**

| Adhyāya 1 | Adhyāya 2 | All Adhyāya |
|---|---|---|
| (कफ, 53) | (तिक्त, 39) | (पित्त, 461) |
| (उष्ण, 47) | (कफ, 31) | (कफ, 438) |
| (पित्त, 45) | (विष, 22) | (गुरु, 254) |
| (तिक्त, 34) | (उष्ण, 21) | (उष्ण, 240) |
| (वात, 32) | (पित्त, 19) | (तिक्त, 237) |
| (शूल, 29) | (कुष्ठ, 18) | (वात, 204) |
| (कुष्ठ, 28) | (अस्र, 18) | (स्मृत, 194) |
| (कास, 25) | (स्मृत, 17) | (कुष्ठ, 177) |
| (कटु, 25) | (कण्डु, 16) | (गुण, 160) |
| (श्वास, 24) | (कटु, 16) | (लघु, 160) |

most frequent nouns as “properties”. The substances are chosen from the rest of the nouns.

### 2.5.3 Synonym Śloka Identification

Generally, the different synonyms of a substance are listed in a single śloka at the beginning of a block. A set $\{n_1, n_2, \ldots n_k\}$ of nouns is called a *synonym-group* if every $n_i$ is a synonym of every other $n_j$. Any such $(n_i, n_j)$ pair is called a *synonym-pair*. A śloka that gives information about a synonym-group or synonym-pairs is referred to as a *synonym* śloka. The first task is to identify instances of such synonym śloka.

To identify a synonym śloka automatically, we use various linguistic features of a śloka and then use them in a classifier. We create a 42-dimensional feature vector per śloka. Table 2.2 enlists all the features used. The features are based on counts and their ratios. Some of the notable features include number of nouns, pronouns and verbs, number of property words present in a śloka, ratios of property words to total number of words, number of words in each case (विभक्ति), and so on. The category "specials" contains adverbs, conjunctions and prepositions.

**Table 2.2:** Features of a śloka.

| | |
|---|---|
| **Counts** | Words, Nouns, Properties, Non-Properties, Special Words, Pronouns, Verbs, Case-$i$ Nouns ($i = 1, \ldots, 8$), Number-$j$ Nouns ($j =$ singular, dual, plural) |
| **Ratio to Words** | Nouns, Properties, Non-Properties, Special Words |
| **Ratio to Nouns** | Properties, Non-Properties, Special Words, Case-$i$ Nouns ($i = 1, \ldots, 8$), Number-$j$ Nouns ($j =$ singular, dual, plural) |
| **Other Ratios** | Properties to Non-Properties, Non-Properties to Properties, Special Words to Properties, Special Words to Non-Properties |

Once each śloka is converted into a 42-dimensional feature vector, various classifiers and ensemble methods are used to classify into a synonym śloka or otherwise.

### 2.5.4 Identifying Synonymous Nouns

Once a synonym śloka is identified, the next task is to identify the synonyms from it. Given a synonym śloka, we first exclude all the property words from it. We next consider the list of all the nouns in the śloka: $\{n_1, n_2, \ldots, n_k\}$.

We call a pair of nouns $(n_i, n_j)$ a *synonym pair* if both $n_i$ and $n_j$ have the same case (विभक्ति) as well as the same number (वचन). We do not use the gender (लिङ्ग) information since there are examples of synonymous substance names that belong to different genders. For example, चव्य (neuter), चव्यिका (feminine) and ऊषणा (feminine) form a synonym group.

## 2.6 Experiments and Results

In this section, we present our experiments and discuss the results. The code is written in Python3. All experiments are done on Intel(R) Core(TM) i7-4770 CPU @

3.40GHz system with 16 GB RAM running Ubuntu 16.04.6 OS. RDF is used for storing the knowledge graph, and querying is done using SPARQL querying language. Python library RDFlib is used for working with RDF and SPARQL.

### 2.6.1 Datasets

We have worked with texts containing two types of relationships:

1. **Human Relationships:** The two well-known epics of ancient India, Rāmāyaṇa and Mahābhārata, contain numerous characters and relationships among them. We have, thus, used them as datasets for human relationships.

2. **Synonymous Relationships of Substances:** Āyurveda, the traditional Indian system of medicine, has a rich source of information about medicinal plants and substances. We considered Bhāvaprakāśanighaṇṭu, a glossary chapter of the Āyurveda text Bhāvaprakāśa as the dataset. It enlists numerous medicinal plants and substances along with their properties and inter-relationships. In this work, we only consider the relationship "is-synonym-of".

Table 2.3 shows the statistics about the datasets considered.

**Table 2.3:** Statistics of the various datasets used.

| **Dataset** | **Rāmāyaṇa** | **Mahābhārata** | **Bhāvaprakāśanighaṇṭu** |
|---|---:|---:|---:|
| Type | Classical | Classical | Technical |
| Chapters | 7 (kāṇḍa) | 18 (parvan) | 23 (adhyāya) |
| Documents | 606 | 2,327 | 23 |
| Śloka | 23,934 | 81,603 | 4,244 |
| Words (total) | 2,69,603 | 17,49,709 | 31,532 |
| Words (unique) | 16,083 | 55,366 | 5,976 |
| Nouns (total) | 1,52,878 | 6,36,781 | 19,689 |
| Nouns (unique) | 9,553 | 20,545 | 3,684 |

**Table 2.4:** Statistics of the knowledge graphs for the human relationships.

| | | Rāmāyaṇa | Mahābhārata |
|---|---|---|---|
| Time taken | Preprocessing | $\sim$ 3.5 days | $\sim$ 13 days |
| | Triplet Extraction | 14.18 sec | 57.19 sec |
| | Triplet Enhancement | 0.40 sec | 2.05 sec |
| Before enhancement | Entities (Nodes) | 1,711 | 3,552 |
| | Triplets (Edges) | 6,155 | 18,936 |
| | Type of Relations | 24 | 25 |
| After enhancement | Entities (Nodes) | 1,711 | 3,552 |
| | Triplets (Edges) | 16,367 | 48,395 |
| | Type of Relations | 27 | 27 |

### 2.6.2 Knowledge Graph from Rāmāyaṇa and Mahābhārata

Table 2.4 shows the various statistics about the knowledge graphs constructed from the datasets Rāmāyaṇa and Mahābhārata.

While pre-processing the text requires a large amount of time, the other steps are significantly faster. The querying times are in microseconds.

#### 2.6.2.1 Questions

To evaluate the performance of the question answering system, we collected $35$ questions from Rāmāyaṇa and $45$ questions from Mahābhārata from 12 different users, with each user contributing between 5-10 questions.

#### 2.6.2.2 Performance

We evaluate the performance of the system for three tasks.

- **QParse** refers to the query parsing task. If the query pattern is correctly formed from the natural language question, we count it as a success; otherwise, it is a failure.

- **QCond** is the conditional question answering task subject to correct query formation. A success is counted only if the answer to the question is completely correct.

**Table 2.5:** Performance of the question answering tasks.

| Text | Task | Total | Found | Correct | Precision | Recall | F1 |
|---|---|---|---|---|---|---|---|
| Rāmāyaṇa | QParse | 35 | 33 | 27 | 0.82 | 0.77 | 0.79 |
| | QCond | 27 | 19 | 09 | 0.47 | 0.33 | 0.39 |
| | QAll | 35 | 20 | 10 | 0.50 | 0.29 | 0.37 |
| Mahābhārata | QParse | 45 | 45 | 41 | 0.91 | 0.91 | 0.91 |
| | QCond | 41 | 36 | 22 | 0.61 | 0.54 | 0.57 |
| | QAll | 45 | 40 | 23 | 0.58 | 0.51 | 0.54 |
| Combined | QParse | 80 | 78 | 68 | 0.87 | 0.85 | 0.86 |
| | QCond | 60 | 55 | 31 | 0.56 | 0.46 | 0.50 |
| | QAll | 80 | 60 | 33 | 0.55 | 0.41 | 0.47 |

- **QAll** is the overall question answering task.

Table 2.5 demonstrates the performance of our system on the collected questions. The query parsing task is fairly accurate. However, the accuracy of question answering has a lot of scope for improvement. We next analyse some of the reasons for failure.

### 2.6.3 Analysis of Wrong Answers

We analyse the wrong answers in two phases: parsing errors and answering errors.

#### 2.6.3.1 Parsing Errors

Following are some examples of queries that got incorrectly parsed.

- गान्धार्याः पुत्राणाम् नामानि कानि → [गान्धारी, पुत्र, किम्]

  The question expects all the names of sons of gāndhārī गान्धारी but the parsed query only asks for the name of 'a son' of गान्धारी. This error originates from the fact that we have not considered the number (वचन) of the relationship word while parsing the question. However, the number can be considered, and all triplets that satisfy the criteria can be returned.

- कर्णार्जुनयोः कः सम्बन्धः → [किम्, किम्, सम्बन्ध]

  There are patterns in the question set that are not handled by our algorithm.

For example, the algorithm did not handle the way of asking the relationship between two people using the word सम्बन्ध and, thus, results in a triplet that does not make sense. If the same question was phrased as कर्णः अर्जुनस्य कः, our algorithm would be able to parse the question to give [अर्जुन, किम्, कर्ण]. Questions like कर्णः अर्जुनस्य कः, अर्जुनस्य कर्णः कः, अर्जुनस्य कः कर्णः and कर्णः कः अर्जुनस्य also get parsed correctly to [अर्जुन, किम्, कर्ण].

- विवाहः अर्जुनस्य अभवत् कया सह → [अर्जुन, किम्, विवाह]

  The question parsing algorithm, while tolerant to some extent, is not fully robust to free word order. An occurrence of विवाह word needs to be followed by the instrumental case (तृतीया) word, followed by सह for it to be parsed correctly. Thus, if the question is changed to अर्जुनस्य विवाहः कया सह अभवत्, it will get parsed correctly to yield [अर्जुन, पत्नी, किम्].

#### 2.6.3.2 Answering Errors

Out of the queries that correctly get parsed, following are the queries which we cannot find the answer due to the inability of performing path queries.

- ऊर्मिला दशरथस्य का → [दशरथ, किम्, ऊर्मिला]

  This question would have got answered only if there is a direct edge between दशरथ and ऊर्मिला. If there is no direct edge, but an edge between दशरथ and लक्ष्मण exists along with the edge between लक्ष्मण and ऊर्मिला, then this answer should have been found. Our inability to pose it as a graph path searching query is the cause of this failure.

- हनुमतः पिता कः → [हनुमत्, पितृ, किम्]

  We correctly parse this question and there exists a triplet [मारुति, पितृ, पवन]. However, as the information that मारुति is another name of हनुमत् is not present in the knowledge graph, resulting in the failure to answer this question.

- पुरोः कः वंशजः यस्य पुत्रः अर्जुनः → [पुरु, वंशज, किम्], [यदु, पुत्र, अर्जुन]

  Again, despite getting correctly parsed, since we cannot follow the “has-son”

**Table 2.6:** Śloka 25, 26, 27 from Adhyāya 67 of Ādi Parvan in Mahābhārata.

| Śloka | Sandhi-Samāsa split |
|---|---|
| अनिलस्य शिवा भार्या तस्याः पुत्रो मनोजवः। | अनिलस्य शिवा भार्या तस्याः पुत्रः मनोजवः। |
| अविज्ञातगतिश्चैव द्वौ पुत्रावनिलस्य तु॥२५॥ | अविज्ञात-गतिः-च-एव द्वौ पुत्रौ=-अनिलस्य तु॥२५॥ |
| प्रत्यूषस्य विदुः पुत्रमृषिं नाम्नाऽथ देवलम्। | प्रत्यूषस्य विदुः पुत्रम्-ऋषिम् नाम्ना-अथ देवलम्। |
| द्वौ पुत्रौ देवलस्यापि क्षमावन्तौ मनीषिणौ। | द्वौ पुत्रौ देवलस्य-अपि क्षमावन्तौ मनीषिणौ। |
| बृहस्पतेस्तु भगिनी वरस्त्री ब्रह्मवादिनी॥२६॥ | बृहस्पतेः-तु भगिनी वर-स्त्री ब्रह्म-वादिनी॥२६॥ |
| योगसिद्धा जगत्कृत्स्नमसक्ता विचचार ह। | योग-सिद्धाः जगत्-कृत्स्नम्-असक्ता विचचार ह। |
| प्रभासस्य तु भार्या सा वसूनामष्टमस्य ह॥२७॥ | प्रभासस्य तु भार्या सा वसूनाम्-अष्टमस्य ह॥२७॥ |

relationship arbitrary number of times, this query cannot be answered.

#### 2.6.3.3 Correct Answers despite Wrong Parsing

Interestingly, there are cases when despite the query being parsed incorrectly, the correct answer exists in the result set. The following examples highlight two such cases.

- रावणस्य कनिष्ठतमः भ्राता कः → [रावण, भ्रातृ, किम्]

  The triplet is incorrectly formed, since we did not capture the information कनिष्ठतमः (youngest). However, the correct answer, विभीषण, being a brother of रावण, is captured in the result set. The question is, thus, deemed to be answered correctly.

- भीमस्य अग्रजः कः आसीत् → [भीम, भ्रातृ, किम्]

  Similar to the previous question, we classify the formed triplet as incorrect, for missing the quality ‘elder’. However, answers found do contain the correct answers युधिष्ठिर and कर्ण.

### 2.6.4 Analysis of Errors in KG Triplets

We now take a look at in-depth analysis of some incorrect triplets retrieved by our method and investigate the reasons behind the failure. For this purpose, we consider a small extract from the corpus and follow the entire pipeline of forming the triplets.

**Table 2.7:** Analysis of Śloka 25.

| Word | Root | Analysis | Is-Noun | Is-Verb | Error |
|---|---|---|---|---|---|
| अनिलस्य | अनिल | [‘g.’, ‘sg.’, ‘m.’] | True | False | |
| शिवा | शिव | [‘nom.’, ‘sg.’, ‘f.’] | True | False | |
| भार्या | भारि | [‘i.’, ‘sg.’, ‘f.’] | True | False | AnalysisError |
| तस्याः | तद् | [‘g.’, ‘sg.’, ‘f.’] | False | False | |
| पुत्रः | पुत्र | [‘nom.’, ‘sg.’, ‘m.’] | True | False | |
| मनो जवः | मनोजव | [‘nom.’, ‘sg.’, ‘m.’] | True | False | Corrected |
| अविज्ञा | अविज्ञ | [‘nom.’, ‘sg.’, ‘f.’] | True | False | OversplitError |
| आत | अत् | [‘pft.’, ‘ac.’, ‘pl.’, ‘2’] | False | True | OversplitError |
| गतिः | गति | [‘nom.’, ‘sg.’, ‘f.’] | True | False | OversplitError |
| च | च | [‘conj.’] | False | False | |
| एव | एव | [‘prep.’] | False | False | |
| द्वौ | द्व | [‘acc.’, ‘du.’, ‘m.’] | True | False | |
| पुत्रौ | पुत्र | [‘acc.’, ‘du.’, ‘m.’] | True | False | |
| अनिलस्य | अनिल | [‘g.’, ‘sg.’, ‘m.’] | True | False | |
| तु | तु | [‘conj.’] | False | False | |

Table 2.6 gives an extract containing three śloka (25, 26 and 27) from Adhyāya 67 of the Ādi Parvan in Mahābhārata. Table 2.7, Table 2.8 and Table 2.9 contain the detailed analysis of these śloka as well as a classification of the errors in the analysis.

#### 2.6.4.1 Types of Errors

We now discuss the possible errors, as exemplified in the analysis Tables 2.7 to 2.9.

- **AnalysisError**:

  This is an error in the analysis obtained from *The Sanskrit Heritage Parser*. For example, the word भार्या in śloka 25 is analysed as a form of भारि instead of a form of भार्या. Thus, the prātipadika identified is wrong. This also results in the other analysis details such as case, gender and number, being wrong. It should be noted that words can be analysed differently in different contexts. For example, the word भार्या, if analysed standalone as a word, can get analysed correctly; however, in the current context, it results in an erroneous analysis.[14] Such errors may be fixed in the future versions of the parser.

[14]Erroneous analysis of भार्या: `https://sanskrit.inria.fr/cgi-bin/SKT/sktreader.cgi?lex=SH&st=t&us=f&cp=t&text=anilasya+zivaa+bhaaryaa+tasyaa.h+putra.h+manojava.h&t=VH&mode=p` (*Accessed on 10 Sep 2019*)

**Table 2.8:** Analysis of Śloka 26.

| Word | Root | Analysis | Is-Noun | Is-Verb | Error |
|---|---|---|---|---|---|
| प्रत्यूषस्य | प्रत्यूष | [‘g.’, ‘sg.’, ‘m.’] | True | False | |
| विदुः | विद् | [‘pft.’, ‘ac.’, ‘pl.’, ‘3’] | False | True | |
| पुत्रम् | पुत्र | [‘acc.’, ‘sg.’, ‘m.’] | True | False | |
| ऋषिम् | ऋषि | [‘acc.’, ‘sg.’, ‘m.’] | True | False | |
| नाम्ना | नामन् | [‘adv.’] | False | False | |
| अथ | अथ | [‘conj.’] | False | False | |
| देवलम् | देवल | [‘acc.’, ‘sg.’, ‘m.’] | True | False | |
| द्वौ | द्व | [‘acc.’, ‘du.’, ‘m.’] | True | False | |
| पुत्रौ | पुत्र | [‘acc.’, ‘du.’, ‘m.’] | True | False | |
| देवलस्य | देवल | [‘g.’, ‘sg.’, ‘m.’] | True | False | |
| अपि | अपि | [‘conj.’] | False | False | |
| क्षमावन्तौ | क्षमावत् | [‘acc.’, ‘du.’, ‘m.’] | True | False | |
| मनीषिणौ | मनीषिन् | [‘acc.’, ‘du.’, ‘m.’] | True | False | |
| बृहस्पतेः | बृहस्पति | [‘g.’, ‘sg.’, ‘m.’] | True | False | |
| तु | तु | [‘conj.’] | False | False | |
| भगिनी | भगिनी | [‘nom.’, ‘sg.’, ‘f.’] | True | False | |
| वर | वर | [‘voc.’, ‘sg.’, ‘m.’] | True | False | OversplitError |
| स्त्री | स्त्री | [‘nom.’, ‘sg.’, ‘f.’] | True | False | OversplitError |
| ब्रह्म | ब्रह्मन् | [‘acc.’, ‘sg.’, ‘n.’] | True | False | OversplitError |
| वा | वा | [‘conj.’] | False | False | OversplitError |
| आदिनी | आदिन् | [‘acc.’, ‘du.’, ‘n.’] | True | False | OversplitError |

**Table 2.9:** Analysis of Śloka 27.

| Word | Root | Analysis | Is-Noun | Is-Verb | Error |
|---|---|---|---|---|---|
| योग | योग | [‘voc.’, ‘sg.’, ‘m.’] | True | False | OversplitError, AnalysisError |
| सिद्धाः | सिद्ध | [‘acc.’, ‘pl.’, ‘f.’] | True | False | OversplitError, SandhiSamaasaError |
| जगत् | जगत् | [‘acc.’, ‘sg.’, ‘n.’] | True | False | |
| कृत्स्नम् | कृत्स्न | [‘acc.’, ‘sg.’, ‘m.’] | True | False | |
| असक्ता | असक्त | [‘nom.’, ‘sg.’, ‘f.’] | True | False | |
| विचचार | वि-चर् | [‘pft.’, ‘ac.’, ‘sg.’, ‘3’] | False | True | |
| ह | ह | [‘part.’] | False | False | |
| प्रभासस्य | प्रभास | [‘g.’, ‘sg.’, ‘m.’] | True | False | |
| तु | तु | [‘conj.’] | False | False | |
| भार्या | भार्य | [‘nom.’, ‘sg.’, ‘f.’] | True | False | |
| सा | तद् | [‘nom.’, ‘sg.’, ‘f.’] | False | False | |
| वसूनाम् | वसु | [‘g.’, ‘pl.’, ‘m.’] | True | False | |
| अष्टमस्य | अष्टम | [‘g.’, ‘sg.’, ‘m.’] | True | False | |
| ह | ह | [‘part.’] | False | False | |

- **OversplitError**:

  This is an error in the sandhi and samāsa splitter, where a word that should not have been split is split. For example, in śloka 26, वरस्त्री is wrongly oversplit as वर and स्त्री, and ब्रह्मवादिनी as ब्रह्म and वादिन्. Sometimes a word is erroneously oversplit by the analyser as well. Again, in śloka 26, for example, वादिन् is erroneously split as वा and आदिन्.

- **SandhiSamaasaError**:

  There can be error in analyzing the correct sandhi and samāsa in a word. In other words, when a word is broken, the constituent words can be erroneous. For example, in śloka 27, योगसिद्धा जगत् is split as योग, सिद्धाः and जगत्, where योगसिद्धा, in addition to being oversplit, is also changed into plural form.

#### 2.6.4.2 Extracting Triplets

After obtaining the analysis, when we proceed to extract triplets as mentioned, we tried using 4 different filters for extracting triplets. In every filter, the case of the subject word must be sixth (षष्ठी) and the gender of the object word must match with the gender of the predicate word. Filters differ in the allowed positions of subject and object words relative to the predicate word as well whether the number (वचन) of the object is matched or not.

**Table 2.10:** Filters for extracting triplets.

| Filter | Position of subject | Position of object | Number (वचन) of object |
|---|---|---|---|
| 1 | Either side of predicate | Either side of predicate | Does not matter |
| 2 | Either side of predicate | Either side of predicate | Must match predicate |
| 3 | Before predicate | After predicate | Must match predicate |
| 4 | After predicate | Before predicate | Must match predicate |

Table 2.10 describe the different filters. Filter 1 is the superset of other filters and Filter 2 is the superset of Filter 3 and Filter 4.

Through empirical evidence, we found that Filter 2, although being stricter than Filter 1, still captures roughly the same number of triplets while reducing the errors.

Filter 3 and Filter 4, while exhibiting fewer mistakes, find fewer correct triplets as well. While we acknowledge that such an analysis is required on a larger scale to decide among the filters, for our purposes, we choose Filter 2 based on the empirical evidence, and proceed further.

#### 2.6.4.3 Analysis of Incorrect Triplets

In this section, we take a look at some wrong triplets that were retrieved and the reasons behind their retrieval.

- (प्रत्यूष, पुत्र, मनीषिन्)

  śloka 26, listed in Table 2.6 contains two relationship words, पुत्रम् and पुत्रौ. The first one is used in relation to देवल who is the son of प्रत्यूष, and the triplet (प्रत्यूष, पुत्र, देवल) is found correctly. However, because of the presence of the second word पुत्रौ, which is actually used with देवलस्य, a wrong triplet (प्रत्यूष, पुत्र, मनीषिन्) is formed. Due to the same reason, (प्रत्यूष, पुत्र, क्षमावत्) is also found. Since the context for finding relationships covers the full śloka, when a single śloka contain multiple relationships, such errors occur. If sentences were instead used, the error could have been reduced. However, there do not exist clear sentence boundaries.

- (बृहस्पति, भगिनी, स्त्री)

  As discussed in Section 2.6.4.1, the word वरस्त्री gets oversplit wrongly into वर and स्त्री, and the split words are analysed separately, resulting in the wrong triplet. Even if this split did not occur, we would have got वरस्त्री as the object in this triplet. This is wrong since this is actually an adjective used for the sister of बृहस्पति. Since we currently do not have any mechanism of distinguishing between nouns and adjectives, it would have resulted in incorrect triplets.

We next examine some triplets that should have been found but were not found and the reasons behind their non-retrieval.

Table 2.11: Training and testing scenarios on Bhāvaprakāśanighaṇṭu.

| Scenario | Training Set | Testing Set |
|---|---|---|
| S1 | First 20% of Adhyāya 1 | Rest 80% of Adhyāya 1 |
| S2 | First 20% of Adhyāya 2 | Rest 80% of Adhyāya 2 |
| S3 | Adhyāya 1 | Adhyāya 2 |
| S4 | Adhyāya 2 | Adhyāya 1 |

- (अनिल, पत्नी, शिवा)

  The relationship word that occurs in śloka 25 in Table 2.6 is भार्या, which suffers an AnalysisError and is identified as तृतीया of भारि instead of प्रथमा of भार्या. Due to the root word (प्रातिपदिक) itself being misidentified, it is not recognized as a relationship word and thus, does not satisfy the filtering criterion. Consequently, the triplet (अनिल, पत्नी, शिवा) is missed.

- (प्रभास, पत्नी, ब्रह्मवादिनी)

  In śloka 27, भार्या of प्रभास is referred to with a pronoun सा, which is connected to a noun in the previous śloka. To correctly identify the triplet (प्रभास, पत्नी, ब्रह्मवादिनी), we would need a mechanism to connect pronouns to their proper subjects. We do not handle this currently.

### 2.6.5 Synonym Identification from Bhāvaprakāśanighaṇṭu

Questions for the Bhāvaprakāśa are implicit, as we are considering only the synonymous relationship. Therefore, the evaluation is performed on the *synonym groups* and *synonym pairs* identification. We created ground truth for the first two adhyāya of Bhāvaprakāśanighaṇṭu. Adhyāya 1 contains 261 śloka, while Adhyāya 2 contains 131 śloka. For each of these śloka, we first identified if it is a *synonym* śloka. If it is so, we next extracted the list of synonymous words contained in it.

#### 2.6.5.1 Classification

Using the feature vectors obtained for each śloka, we used various classifiers to classify each śloka as a synonym śloka or otherwise. We tried four practical scenarios of training and testing set choices as described in Table 2.11.

**Table 2.12:** Performance of classifiers in identifying synonym śloka.

| Scenario | Train Size | Test Size | $P$ | $P'$ | $TP$ | Accuracy | Precision | Recall | F1 |
|---|---|---|---|---|---|---|---|---|---|
| S1 | 52 | 209 | 84 | 56 | 42 | 0.73 | 0.75 | 0.50 | 0.60 |
| S2 | 26 | 105 | 44 | 43 | 31 | 0.76 | 0.72 | 0.71 | 0.71 |
| S3 | 261 | 131 | 54 | 45 | 36 | 0.79 | 0.80 | 0.67 | 0.73 |
| S4 | 131 | 261 | 90 | 99 | 66 | 0.78 | 0.67 | 0.73 | 0.70 |

**Table 2.13:** Examples of errors in classification (scenario S3).

| **False Positives** (9) | **False Negatives** (18) |
|---|---|
| कामरूपोद्भवा कृष्णा नैपाली नीलवर्णयुक्<br>काश्मीरी कपिलच्छाया कस्तूरी त्रिविधा स्मृता ॥६॥ | श्रीखण्डं चन्दनं न स्त्री भद्र श्रीस्तैलपर्णिकः<br>गन्धसारो मलजयस्तथा चन्द्र द्युतिश्च सः ॥११॥ |
| महिषाक्षो महानीलः कुमुदः पद्म इत्यपि<br>हिरण्यः पञ्चमो ज्ञेयो गुग्गुलोः पञ्च जातयः ॥३३॥ | भद्र मुस्तञ्च गुन्द्रा च तथा नागरमुस्तकः<br>मुस्तं कटु हिमं ग्राहि तिक्तं दीपनपाचनम् ॥९३॥ |

The size of training sets were chosen to be smaller than those of test sets to resemble the real-world scenario where the ground truth can be created for only a small portion of the text, and predictions are needed to be made on the rest.

Table 2.12 shows the performance of the best classifier under various scenarios in identifying the śloka containing synonyms.

Table 2.13 shows some examples of wrongly classified śloka for the best performing scenario S3.

#### 2.6.5.2 Synonym Identification

We next evaluate the performance of finding synonymous pairs from a synonym śloka. Table 2.14 shows the performance in identifying groups of synonymous substances. We say that a group of substances is *covered* even if a single pair in the group is identified. The result shows that even this has a scope for improvement.

**Table 2.14:** Group coverage in synonym pair identification.

| | Synonym Śloka | Groups present | Groups found | Group coverage |
|---|---|---|---|---|
| Adhyāya 1 | 90 | 87 | 60 | 0.69 |
| Adhyāya 2 | 54 | 53 | 39 | 0.74 |

Table 2.15 shows an example of a synonym śloka where none of the pairs are extracted correctly. The correct synonyms are चन्द्रिका, चर्महन्त्री, पशुमेहनकारिका, नन्दिनी, कारवी, भद्रा, वासपुष्पा, सुवासरा. We find the pairs (कारिका, हन्तृ), (कारिका, भद्र), (कारिका, सपुष्प), (नन्दिन्, रवि), (भद्र, हन्तृ), (भद्र, सपुष्प), (सपुष्प, हन्तृ), none of which are correct. The reasons for the errors are shown in Table 2.16. Almost all the nouns are analysed incorrectly, resulting in the group being completely missed.

**Table 2.15:** Śloka 96 from Adhyāya 1 of Bhāvaprakāśanighaṇṭu and its sandhi-samāsa split.

| Synonym Śloka | Sandhi-Samāsa Split |
|---|---|
| चन्द्रि का चर्महन्त्री च पशुमेहनकारिका।<br>नन्दिनी कारवी भद्रा वासपुष्पा सुवासरा ॥९६॥ | चन्द्रि का चर्महन्त्री च पशुमेहन-कारिका।<br>नन्दिनी कारवी भद्रा वासपुष्पा सु-वासराः ॥९६॥ |

In addition to the errors discussed in Section 2.6.4.1, an additional error occurs here, that of **TextError**. This refers to an error in the text corpus that we are working with. In particular, the original śloka contains the word चन्द्रिका while the corpus we are working with, has that word split as चन्द्रि and का, which results in this word not being analysed correctly. After correcting this error manually, we now obtain a valid pair (चन्द्रिका, भद्रा), thus covering this group.

We next analyse the finer errors that occur when some members of a synonymous group are identified correctly, but not all. Table 2.17 shows the performance.

Table 2.18 shows a synonym śloka from Adhyāya 1 (हरीतक्यादिवर्गः).

This śloka contains a total of 11 synonyms. We find pairs of synonyms involving 9 out of these, synonym pairs involving 8 of which are correct. We show examples of some of the false negatives and false positives among the pairs of synonyms identified.

- **False Positive:** (अमृता, अवी)

  The word अव्यथा is split wrongly as अवी and अथा, and are then analysed separately. This results in both अमृता and अवी being in the same case (प्रथमा) and same number (एकवचन), thus getting wrongly marked as a synonymous pair.

- **False Negative:** (अभया, अमृता)

  The word अभया gets analysed as instrumental (तृतीया) case of अभा instead of nominative (प्रथमा) case of अभया. This results in a case mismatch with अमृता and the pair is not extracted as a synonymous pair.

**Table 2.16:** Analysis of Śloka 96.

| Word | Root | Analysis | Is-Noun | Is-Verb | Error |
|---|---|---|---|---|---|
| चन्द्रि | चन्द्रि | ['?'] | False | False | TextError |
| का | किम् | ['nom.', 'sg.', 'f.'] | False | False | TextError |
| चर्म | चर्मन् | ['acc.', 'sg.', 'n.'] | True | False | OversplitError |
| हन्त्री | हन्तृ | ['nom.', 'sg.', 'f.'] | True | False | OversplitError |
| च | च | ['conj.'] | False | False | |
| पशुमेहन | पशुमेहन | ['voc.', 'sg.', 'n.'] | True | False | OversplitError |
| कारिका | कारिका | ['nom.', 'sg.', 'f.'] | True | False | OversplitError |
| नन्दिनी | नन्दिन् | ['acc.', 'du.', 'n.'] | True | False | AnalysisError |
| का | किम् | ['nom.', 'sg.', 'f.'] | False | False | OversplitError |
| रवी | रवि | ['acc.', 'du.', 'm.'] | True | False | OversplitError |
| भद्रा | भद्र | ['nom.', 'sg.', 'f.'] | True | False | |
| वा | वा | ['conj.'] | False | False | OversplitError |
| सपुष्पा | सपुष्प | ['nom.', 'sg.', 'f.'] | True | False | OversplitError |
| सु | सु | ['?'] | False | False | OversplitError |
| वासराः | वासर | ['voc.', 'pl.', 'm.'] | True | False | OversplitError |

**Table 2.17:** Performance of finding synonym pairs.

| | Śloka | Synonym Śloka | $P$ | $P'$ | $TP$ | Precision | Recall | F1 |
|---|---|---|---|---|---|---|---|---|
| Adhyāya 1 | 231 | 90 | 534 | 562 | 369 | 0.66 | 0.69 | 0.67 |
| Adhyāya 2 | 161 | 54 | 300 | 348 | 214 | 0.62 | 0.71 | 0.66 |

**Table 2.18:** Example of wrong pairs from Adhyāya 1 of Bhāvaprakāśanighaṇṭu.

| Synonym Śloka | Sandhi-Samāsa split | $P$ | $P'$ | $TP$ |
|---|---|---|---|---|
| हरीतक्यभया पथ्या कायस्था पूतनाऽमृता<br>हैमवत्यव्यथा चापि चेतकी श्रेयसी शिवाः ॥६॥ | हरीतकी-अभया पथ्या कायस्था पूतना-अमृता<br>हैमवती-अव्यथा च-अपि चेतकी श्रेयसी शिवाः ॥६॥ | 11 | 9 | 8 |

## 2.7 Summary

In this chapter, we have described a framework to build a knowledge graph (KG) directly from Sanskrit texts, and use it for question answering in Sanskrit. Our framework has multiple components and is based on rules and heuristics developed using the knowledge of grammar of Sanskrit language and structure of the text.

One of the primary outcomes of this effort is the realization that for almost all the components, the accuracy can be improved. Improvements on any of these components will make the system better. A word analyser that produces all possible analyses can benefit from a disambiguator which chooses from the generated options. Usage of dictionaries, thesauri (such as Amarakoṣa) and Sanskrit WordNet needs to be explored to see if they can help in understanding the structure of a word better. Crowdsourcing tools as well as human experts can also help refine some of the steps.

This effort serves as a step towards the ultimate aim of automatically building a full-fledged knowledge graph from a Sanskrit corpus and paves the way for a more generic question answering framework.

# Chapter 3

# *Sangrahaka*: Annotation and Querying Tool for Knowledge Graphs

We present a web-based tool *Sangrahaka* for annotating entities and relationships from text corpora towards construction of a knowledge graph and subsequent querying using templatized natural language questions. The application is language and corpus agnostic, but can be tuned for specific needs of a language or a corpus. The application is freely available for download and installation. Besides having a user-friendly interface, it is fast, supports customization, and is fault tolerant on both client and server side. It outperforms other annotation tools in an objective evaluation metric. The framework has been successfully used in two annotation tasks. The code is available from `https://github.com/hrishikeshrt/sangrahaka/`.

We also describe our efforts on manual annotation of Sanskrit text for the purpose of knowledge graph (KG) creation. We choose three chapters: Dhānyavarga, Śākavarga and Māṃsavarga from Bhāvaprakāśanighaṇṭu portion of the Āyurveda text Bhāvaprakāśa for annotation. The constructed knowledge graph contains 1606 entities and 1707 relationships. Since Bhāvaprakāśanighaṇṭu is a technical glossary text that describes various properties of different substances, we develop an elaborate ontology to capture the semantics of the entity and relationship types present in the text. To query the knowledge graph, we design 31 query templates that cover most of the common question patterns. For both manual annotation and querying, we customize the *Sangrahaka* framework previously developed by us. The entire system including the dataset is available from `https://sanskrit.iitk.ac.in/ayurveda/`. We hope that the knowledge graph that we have created through manual annotation and subsequent curation will help in development and testing of NLP tools in future as well as studying of the Bhāvaprakāśanighaṇṭu text.

**Table 3.1:** Feature Comparison *Sangrahaka* with Various Annotation Tools

| Feature | WebAnno | GATE | BRAT | FLAT | doccano | *Sangrahaka* |
|---|---|---|---|---|---|---|
| Distributed Annotation | ✓ | ✓ | ✓ | ✓ | ✓ | ✓ |
| Simple Installation | | | ✓ | ✓ | ✓ | ✓ |
| Intuitive | ✓ | ✓ | ✓ | | ✓ | ✓ |
| Entity and Relationship | ✓ | ✓ | ✓ | ✓ | | ✓ |
| Query Support | | | | | | ✓ |
| Crash Tolerance | | | | | | ✓ |

## 3.1 *Sangrahaka* Software

For the purpose of knowledge graph focused annotation, it is important to have capabilities for multi-label annotations and support for annotating relationships.

Several text annotation tools are readily available to handle a wide range of text annotation tasks including classification, labeling and sequence-to-sequence annotations.. These tools include *WebAnno* [Yimam et al., 2013], *FLAT* [van Gompel, 2014], *BRAT* [Stenetorp et al., 2012], *GATE Teamware* [Bontcheva et al., 2013], and *doccano* [Nakayama et al., 2018].

While *WebAnno* is extremely feature rich, it compromises on simplicity. Further, its performance deteriorates severely as the number of lines displayed on the screen increases. *GATE* also has the issue of complex installation procedure and dependencies. *FLAT* has a non-intuitive interface and non-standard data format. Development of *BRAT* has been stagnant, with the latest version being published as far back as 2012. The tool *doccano*, while simple to setup and use, does not support relationship annotation. Thus, unfortunately, none of these tools supports all the desired features of an annotation framework for the purpose of knowledge graph annotation. Further, none of the above frameworks provide an integration with a graph database, a querying interface, or server and client side crash tolerance.

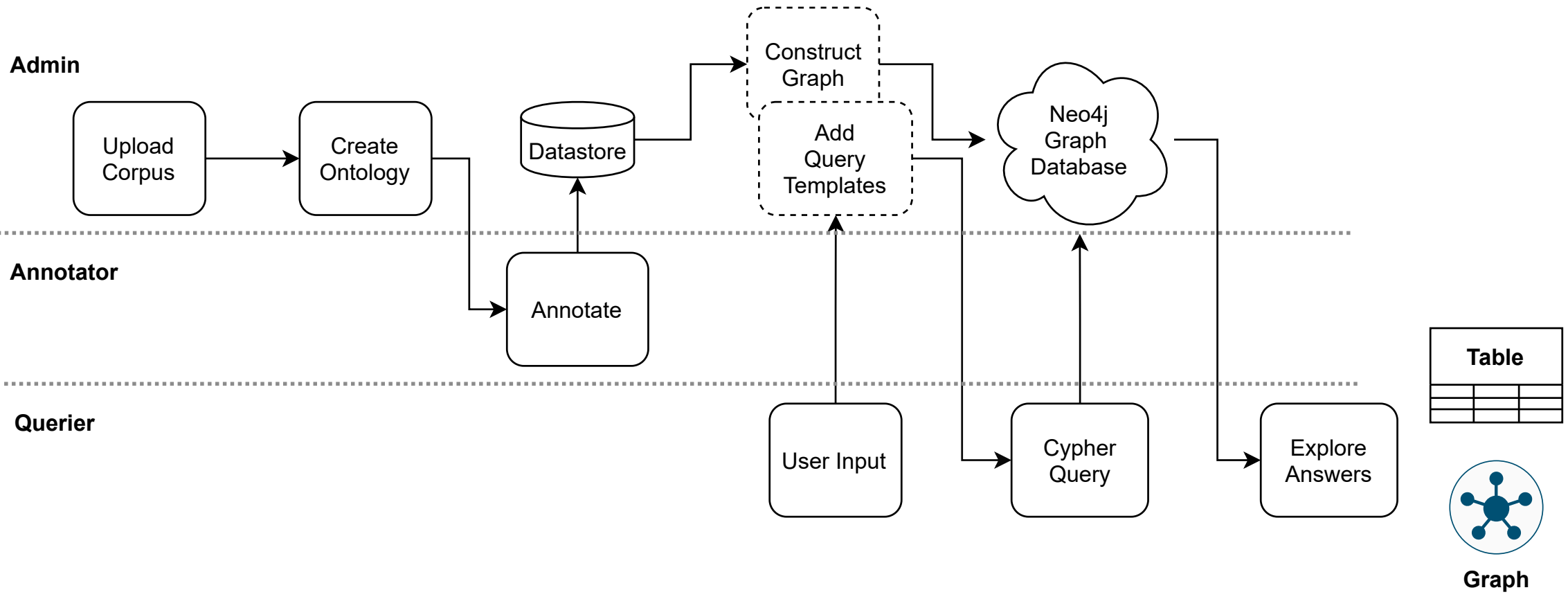

**Figure 3.1:** Workflow of Admin, Annotator and Querier roles and their interaction with each other. Corpus creation, ontology creation, annotation, graph creation, graph querying are the principal components.

Thus, to satisfy the need of an annotation tool devoid of these pitfalls, we present *Sangrahaka*. It allows users to annotate and query through a single platform. The application is language and corpus agnostic, but can be customized for specific needs of a language or a corpus. Table 3.1 provides a high-level feature comparison of these annotation tools including *Sangrahaka*.

A recently conducted extensive survey [Neves and Ševa, 2021] evaluates 78 annotation tools and provides an in-depth comparison of 15 tools. It also proposes a scoring mechanism by considering 26 criteria covering publication, technical, function and data related aspects. We evaluate *Sangrahaka* and other tools using a the same scoring mechanism, albeit with a modified set of criteria. The details are in Section 3.1.3.

### 3.1.1 Architecture

*Sangrahaka* is a language and corpus agnostic tool. Salient features of the tool include an interface for annotation of entities and relationships, and an interface for querying using templatized natural language questions. The results are obtained by querying a graph database and are depicted in both graphical and tabular formats. The tool is also equipped with an administrators' interface for managing user access levels, uploading corpora and ontology creation. There are utility scripts for language-specific or corpus-specific needs. The tool can be deployed on the Web for distributed annotation by multiple annotators. No programming knowledge is expected from an annotator.

**Table 3.2:** Roles and Permissions

| Permissions | Roles | | | |
|---|---|---|---|---|
| | Querier | Annotator | Curator | Admin |
| Query | ✓ | ✓ | ✓ | ✓ |
| Annotate | | ✓ | ✓ | ✓ |
| Curate | | | ✓ | ✓ |
| Create Ontology | | | | ✓ |
| Upload Corpus | | | | ✓ |
| Manage Access | | | | ✓ |

The project relies on several key tools and technologies for its implementation. These include *Python 3.8* [Van Rossum and Drake, 2009], *Flask 1.1.2* [Ronacher, 2011, Grinberg, 2018], *Neo4j Community Server 4.2.1* [Webber, 2012], *SQLite 3.35.4* [Hipp, 2022] for the backend and *HTML5*, *JavaScript*, *Bootstrap 4.6* [boo, 2021], and *vis.js* [vis, 2021] for the frontend.

#### 3.1.1.1 Workflow

Figure 3.1 shows the architecture and workflow of the system.

The tool is presented as a web-based full-stack application. To deploy it, one first configures the application and starts the server. A user can then register and login to access the interface. The tool uses a role based access system. Roles are Admin, Curator, Annotator, and Querier. Permissions are tied to roles. Table 3.2 enlists the roles and the permissions associated with them. A user can have more than one role. Every registered member has permission to access user control panel and view corpus.

An administrator creates a corpus by uploading the text. She also creates a rele-vant ontology for the corpus and grants annotator access to relevant users. The ontology specifies the type of entities and relationships allowed. An annotator signs-in and opens the corpus viewer interface to navigate through lines in the corpus. For every line, an annotator then marks the relevant entities and relationships. A curator can access annotations by all annotators, and can make a decision of whether to keep or discard a specific annotation. This is useful to resolve conflicting annotations. An administrator may customize the graph generation mechanism based on the semantic task and semantics of the ontology. She then imports the generated graph into an independently running graph database server. A querier can then access the querying interface and use templatized natural language questions to generate graph database queries. Results are presented both in graphical as well as tabular formats and can be downloaded as well.

### 3.1.2 Data Format

The tool relies on the widely-accepted JSON data format as its core method for managing a range of tasks, encompassing corpus input and query template definitions. For a comprehensive understanding of these formats, we present a detailed description along with sample examples. Further information can be found at `https://github.com/hrishikeshrt/sangrahaka/tree/main/examples`.

#### 3.1.2.1 Corpus Format

The top-level structure of the corpus data is represented as a list, wherein each element corresponds to a line in the text. This hierarchical organization allows for easy access and manipulation of individual lines within the overall structure.

Each line object within the list contains several key-value pairs that provide specific information about that particular line. The *text* key stores the actual text of the line, capturing the textual content in its original form.

Additionally, the *split* key, if present, contains the line's text with word segmentation applied. This segmentation breaks down the line into individual words or tokens, facilitating further analysis and processing at the word level.

For lines belonging to poetic verses, the *verse* key, when available, provides the verse identifier associated with that particular line. This allows for the identification and organization of lines within the broader context of the poem or verse structure.

Furthermore, the *analysis* key, if provided, contains linguistic information pertaining to the line. This information is stored as a list of key-value pairs, with each pair representing a specific token within the line. The analysis may include details such as part-of-speech tags, morphological analysis, syntactic dependencies, or any other relevant linguistic annotations associated with the tokens in the sentence.

The following is an example of a corpus file containing a single sentence.

```
[{
  "verse": 1,
  "text": "To sainted Nárad, prince of those",
  "analysis": {
    "source": "spacy",
    "tokens": [
      {"Word": "Nárad", "Lemma": "Nárad", "POS": "PROPN"},
      {"Word": "prince", "Lemma": "prince", "POS": "NOUN"}
    ]
  },
},]
```

#### 3.1.2.2 Query Template

The top-level structure consists of a list of query objects, each representing a specific query template. These query objects contain the following keys: *gid*, *groups*, *texts*, *cypher*, *input*, and *output*. *gid* is used for grouping similar queries together in the frontend. *groups* and *texts* are objects that store language names as keys, with corresponding group names and query texts in those languages as their respective values. If a query requires user input, it is indicated by placeholders such as `{0}` or {1} within the query text. The *input* key contains a list of objects that provide information for populating user-input elements in the frontend. Each object in this list should have a unique *id* for the element and specify the *type* of input element. Valid types include *entity*, *entity_type*, *relation*, and *relation_detail*.

Here is an example of a query template file that includes a single query:

```
[{
  "gid": "1",
  "cypher": "MATCH (p1)-[r:IS_FATHER_OF]->(p2) "
            "WHERE p2.lemma =~ \"{0}\" RETURN *",
  "input": [{"id": "p", "type": "entity"},],
  "output": ["p1", "r", "p2"],
  "texts": {"english": "Who is the father of {0}?",},
  "groups": {"english": "Kinship",}
},]
```

#### 3.1.2.3 Backend

The backend is written in *Python*, using *Flask*, a micro-webframework. Pluggable components of the backend are a relational database and a Neo4j graph database server.

**3.1.2.3.1 Web Framework** The web framework manages routing, templating, user-session management, connections to databases, and other backend tasks. A *Web Server Gateway Interface (WSGI)* HTTP server runs the Flask application. We use *Gunicorn* [gun, 2021] running behind an *NGINX* [ngi, 2021] reverse proxy for this purpose. However any WSGI server, including the Flask's in-built server, can be used.

**3.1.2.3.2 Data** Data related to user accounts, roles as well as corpus text, ontology, entity annotations and relationship annotations are stored in a relational database. This choice is made due to the need of cross-references (in database parlance, *joins*) across user, corpus and annotation related information. Any relational database compatible with *SQLAlchemy* [sql, 2021] can be used. We have used *SQLite*. An administrator uploads various chapters in a corpus as JSON files using a pre-defined format. Each JSON object contains text of the line and optional extra information such as word segmentation, verse id, linguistic information etc. The structure of JSON file corresponding to a chapter is explained in Section 3.1.2.1. This information is then organized in a hierarchical structure with 4 levels: *Corpus*, *Chapter*, *Verse* and *Line*. Additionally, there is an *Analysis* table that stores the linguistic information for each line. The system is equipped to deal with morphologically rich languages. Lemmas (i.e., word roots) are stored in a separate table and referenced in entity and relationship annotations. Every entity annotation consists of a lemma, an entity type, a line number and user-id of the annotator. Every relationship annotation consists of a source (lemma), a target (lemma), a relationship type, an optional detail text, a line number and user-id of the annotator.

**Table 3.3:** List of important configuration options and their explanation

| Option | Explanation |
|---|---|
| Admin user | Username, Password and E-mail of owner |
| Roles | Configuration of Roles and Permissions |
| SQL config | *SQLAlchemy* compatible Database URI |
| Neo4j config | Server URL and Credentials |

**3.1.2.3.3 Knowledge Graph** A knowledge graph is constructed using the entity and relationship annotations. *Neo4j* is used as the graph database server to store and query the KG. Connection to it is made using the *Bolt* protocol [bol, 2021]. Hence, the graph database can exist independently on a separate system. *Cypher* query language [cyp, 2021] is used to query the graph database and produce results.

**3.1.2.3.4 Natural Language Query Templates** Templates for natural language questions are added by an administrator. A query template has two essential components, a natural language question with placeholder variables and a *Cypher* equivalent of the query with references to the same placeholder variables. Placeholder variables represent values where user input is expected. Query templates are provided in a JSON file whose structure is given in Section 3.1.2.1. The natural language query template, combined with user input, forms a valid natural language question, and the same replacement in *Cypher* query template forms a valid *Cypher* query.

**3.1.2.3.5 Configuration** The application contains several configurable components. The entire configuration setting is stored in a settings file. Table 3.3 explains some important configuration options.

**3.1.2.3.6 Utility Scripts** Utility Python scripts are provided for tasks that need to be performed in the background. The primary among these is a graph generation script to generate JSONL [jso, 2021] formatted data suitable for direct import in the *Neo4j Graph Database*. Sample scripts are also provided for generation of corpus file and query template file. These can be easily customized to suit corpus specific or application specific needs.

#### 3.1.2.4 Frontend

The frontend is in form of a web application. HTML5 webpages are generated using *Jinja* template engine [jin, 2021], styled using *Bootstrap 4.6* and made interactive using *JavaScript*. The web-based user interface has several components that are accessible to users based on their roles. Some of these components are shown in Figure 3.2.

**3.1.2.4.1 Corpus Viewer Interface** The corpus viewer interface consists of a row-wise display of lines in a corpus. For such languages such as Sanskrit, German, Finnish, Russian, etc. that exhibit a large number of compound words, the corpus viewer can display the word-split output added by the administrator. Further, an administrator may run other language specific tools to obtain any kind of semantic and syntactic information about the components of the sentence as a list of key-value pairs. The corpus viewer displays this information in a tabular format whenever a line is selected.

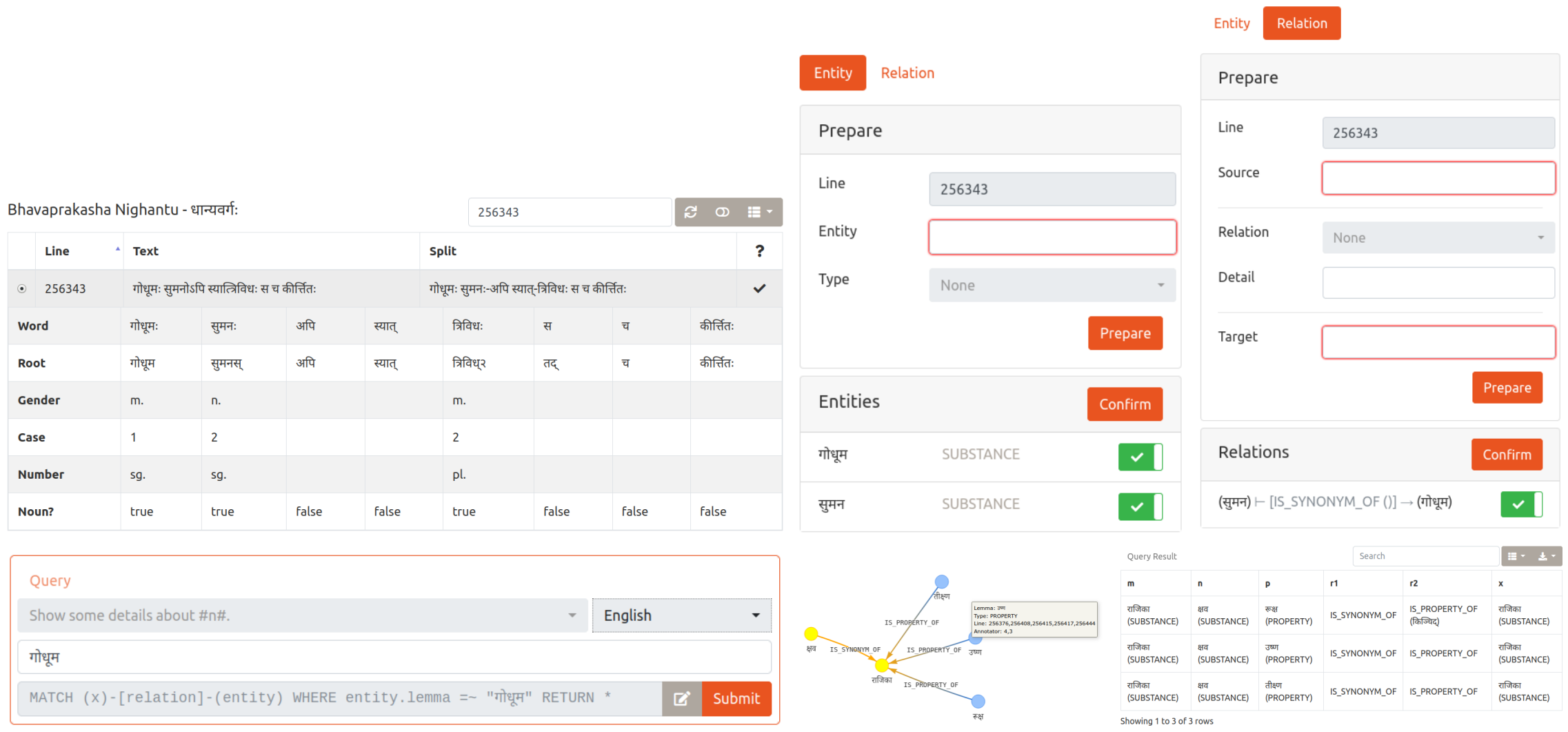

**Figure 3.2:** Corpus Viewer, Entity Annotator, Relation Annotator, Query Interface, Graphical Result Interface, Tabular Result Interface

**3.1.2.4.2 Annotator Interface** The annotator interface is interlinked with the corpus viewer interface. It contains two views, one for entity annotation and the other for relation annotation. Adaptive auto-complete suggestions are offered based on previously added lemmas and lemmas present in the line being annotated.

**3.1.2.4.3 Query Interface** The query interface makes use of pre-defined natural language query templates and combines them with user input to form *Cypher* queries. A user may directly edit the *Cypher* query as well if she so desires. These are communicated to the graph database using *Bolt* protocol and results are fetched. Result of a *Cypher* query is a subgraph of the knowledge graph and is presented in an interactive interface that allows users to zoom-in to specific areas of the graph, rearrange nodes and save the snapshot of the graph as an image. Results are also displayed in a tabular manner and can be exported in various file formats including CSV, JSON, text, etc.

**3.1.2.4.4 Graph Query Builder Interface** Templatized querying interface, although extremely easy to use, is unable to answer complex factual questions. For example, while the query *Which substance increases* vāta*?* can be covered by a single-edge query, a query with multiple relations like *Which variant of* mudga *increases* vāta*?* cannot be handled by a single-edge query template. Note that the intent of the question is to find a substance that is (1) a variant of mudga and (2) increases vāta, and, hence, both the edges are important. Another complex question, *Which substances of red colour and soft texture increase* vāta *and decrease* pitta*?* also falls into the same category. While it is possible to construct templates to fit these questions, it is infeasible to try and imagine all possible question templates that may be posed.

Hence, we built a *graph query builder interface*. This interface allows the user to add nodes and edges and construct an arbitrary graph structure. The user can also specify node and edge specific properties. After the construction of the graph structure, the interface lets user generate the corresponding *Cypher* query for querying the KG system.

**3.1.2.4.5 Graph Browser Interface** The graph browser interface offers an intuitive means of exploring the knowledge graph by navigating through the neighboring nodes of any displayed node. Whenever a user clicks on a visible node, they are presented with two options: (1) to view all the neighbors of the clicked node *in addition to* the current graph being shown, or (2) to view the neighbors *in place of* the current graph. This functionality provides users with flexibility and control over their exploration, allowing them to seamlessly expand their understanding of the graph's interconnected information. Figure 3.3 showcases this interface.

**3.1.2.4.6 Admin Interface** The administrator frontend allows an administrator to perform tasks such as change users' access levels, create corpus, upload chapters in a corpus, and create ontology. Adding a new corpus requires two steps: corpus creation and chapter upload. The corpus creation step refers to creating a new entry in the *Corpus* table along with a description. Once a corpus has been added, chapters associated with the corpus can be uploaded.

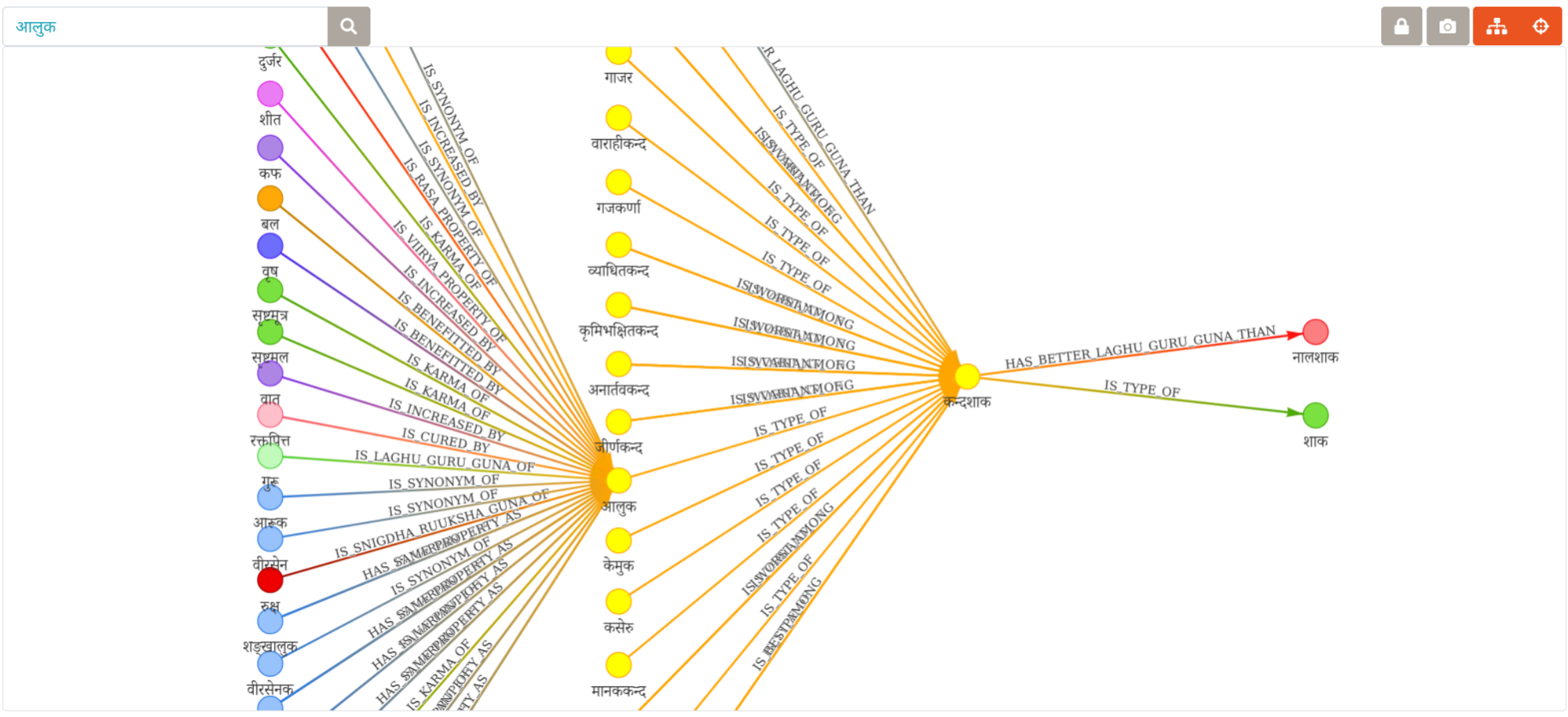

**Figure 3.3:** Graph Browser Interface

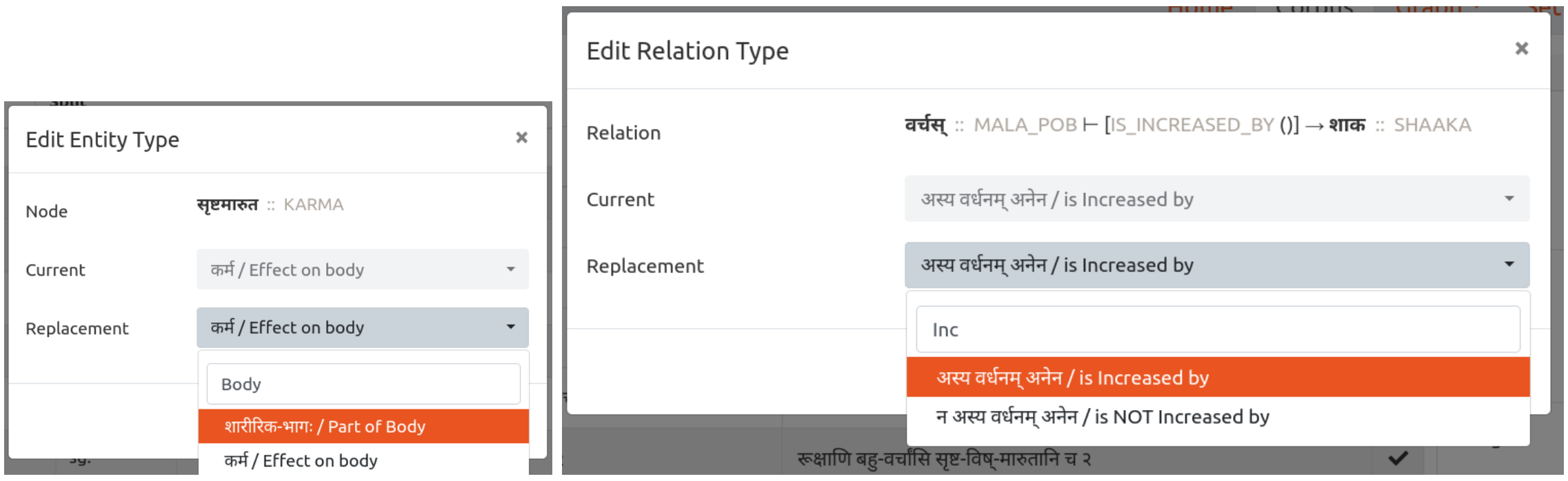

**Figure 3.4:** Curation Interfaces for Editing Node Label and Relation Label

**3.1.2.4.7 Ontology Creation** The ontology creation interface allows an administrator to add or remove node types and relation types. If an entity or relation type is being used in an annotation, removal of the same is prevented.

**3.1.2.4.8 Curation** Curation is performed through the annotation interface. A curator can see annotations by all annotators and can choose to keep or remove them as well as add new annotations. We have created some special interfaces for commonplace curation tasks such as changing a node's category, or changing the type of relation between two nodes. These interfaces can be accessed through the context menu (i.e. by right clicking a node or a relation in the list). They are illustrated in Figure 3.4.

**Table 3.4:** Annotation tasks performed using *Sangrahaka*

| Corpus | Lines | Annotators | Ontology | | Annotations | | Progress |
|---|---|---|---|---|---|---|---|
| | | | Nodes | Relations | Nodes | Relations | |
| BPN [bpn, 2016] | 180 | 5 | 25 | 30 | 602 | 778 | 100% |
| VR [val, 2021] | 17655 | 9 | 107 | 132 | 1810 | 2087 | 54% |

#### 3.1.2.5 Fault Tolerance

Corpus viewer and annotation interface act as a *single-page application* [wik, 2021b] and make use of *AJAX* [wik, 2021a] calls to communicate with the server. Entity and Relation annotation processes have two steps, ‘Prepare’ and ‘Confirm’. Once an entity or a relation is prepared, it is stored in a browser based `localStorage` [Hickson, 2021] that persists across browser sessions and is, thus, preserved even if the browser crashes. Once a user clicks ‘Confirm’, an attempt to contact the server is made. If the attempt is successful and the data is inserted in the database successfully, the server returns success and the data associated with that annotation is cleared from the local storage. If the server returns failure, the data persists. Thus, a server crash does not affect the user’s unconfirmed annotations. Further, if a page is already loaded in the browser and the server crashes, a user can still continue to annotate. The unconfirmed entities and relations are color coded and can be easily located later to confirm once the server is restored. Thus, the application is fault-tolerant on both client and server side. This is an important feature that distinguishes *Sangrahaka* from other tools.

### 3.1.3 Evaluation

The tool has been used for two distinct annotation tasks: (1) a chapter from a medical text (*Ayurveda*) corpus in Sanskrit (*BPN*) [bpn, 2016], and (2) full text of the epic *Ramayana* in English (*VR*) [val, 2021]. Table 3.4 presents details of these tasks.

Due to the nature of semantic annotation, where an annotator usually has to spend more time on mentally processing the text to decide the entities and relationships than the actual mechanical process of annotating the text, and the fact that annotations are done over several sessions of various lengths over an extended period of time, time taken for annotation is not an adequate metric of evaluation.

**Table 3.5:** Annotator Ratings for *Sangrahaka* across various metrics

| Metric | Score |
|---|---|
| Looks and feel | 4.7 |
| Ease of use | 4.4 |
| Annotation Interface | 4.5 |
| Querying Interface | 4.6 |
| Administrative Interface | 4.7 |
| Overall | 4.5 |

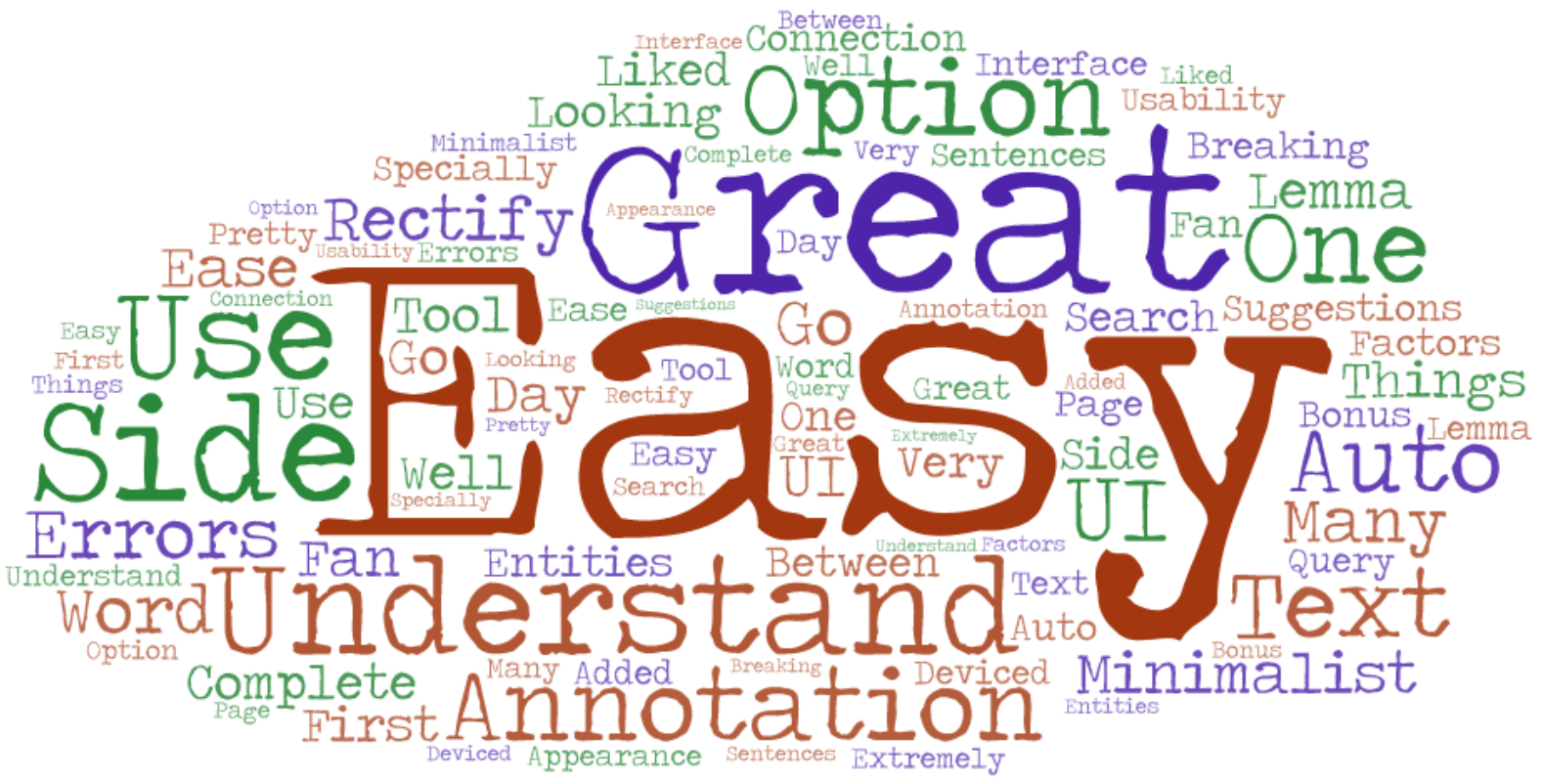

**Figure 3.5:** Wordcloud created using comments provided by users in the survey

#### 3.1.3.1 Subjective Evaluation

As a subjective evaluation, a survey was conducted among the annotators from two annotation tasks. They were asked to rate the tool on a scale of 5 in several metrics. The survey also asked them to describe their experience with *Sangrahaka*. A total of 10 annotators participated in the survey. We received an overall rating of 4.5 from 10 annotators.

Table 3.5 shows the ratings given by the users. Figure 3.5 shows a word-cloud representation of the testimonials provided by the users.

#### 3.1.3.2 Objective Evaluation

To perform the objective evaluation, we adopted the methodology employed in the study conducted by [Neves and Ševa, 2021]. They utilized a set of 26 criteria, catego-rized into four groups: publication, technical, data, and functional. For our evaluation, we excluded parameters related to publication and citations. Additionally, we removed criteria (F2 and F6) that were not met by any of the tools in the comparison, as they were not applicable for score calculation. Instead, we introduced four new criteria: (A1) support for querying, (A2) server-side crash tolerance, (A3) client-side crash tolerance, and (A4) distributed annotation support.

| Criteria | | | Tools | | | | |
|---|---|---|---|---|---|---|---|
| **ID** | **Description** | **Weight** | **WebAnno** | **doccano** | **FLAT** | **BRAT** | **Sangrahaka** |
| P1 | Year of the last publication | 0 | 1 | 0 | 0 | 1 | 1 |
| P2 | Citations on Google Scholar | 0 | 1 | 0 | 0 | 1 | 0 |
| P3 | Citations for Corpus Development | 0 | 1 | 0 | 0 | 1 | 0 |
| T1 | Date of the last version | 1 | 1 | 1 | 1 | 0.5 | 1 |
| T2 | Availability of the source code | 1 | 1 | 1 | 1 | 1 | 1 |
| T3 | Online availability for use | 1 | 0 | 0 | 1 | 0 | 0 |
| T4 | Easiness of Installation | 1 | 0 | 1 | 1 | 0.5 | 1 |
| T5 | Quality of the documentation | 1 | 1 | 1 | 1 | 1 | 0.5 |
| T6 | Type of license | 1 | 1 | 1 | 1 | 1 | 1 |
| T7 | Free of charge | 1 | 1 | 1 | 1 | 1 | 1 |
| D1 | Format of the schema | 1 | 1 | 1 | 1 | 0.5 | 1 |
| D2 | Input format for documents | 1 | 1 | 0.5 | 1 | 1 | 1 |
| D3 | Output format for annotations | 1 | 1 | 1 | 1 | 0.5 | 0 |
| F1 | Allowance of multi-label annotations | 1 | 1 | 0 | 1 | 1 | 1 |
| F2 | Allowance of document level annotations | 0 | 0 | 0 | 0 | 0 | 0 |
| F3 | Support for annotation of relationships | 1 | 1 | 0 | 0 | 1 | 1 |
| F4 | Support for ontologies and terminologies | 1 | 1 | 0 | 1 | 1 | 1 |
| F5 | Support for pre-annotations | 1 | 0.5 | 0 | 0.5 | 0.5 | 0 |
| F6 | Integration with PubMed | 0 | 0 | 0 | 0 | 0 | 0 |
| F7 | Suitability for full texts | 1 | 0.5 | 0.5 | 1 | 1 | 1 |
| F8 | Allowance for saving documents partially | 1 | 1 | 1 | 1 | 1 | 1 |
| F9 | Ability to highlight parts of the text | 1 | 1 | 1 | 1 | 1 | 1 |
| F10 | Support for users and teams | 1 | 0.5 | 0.5 | 1 | 0.5 | 0.5 |
| F11 | Support for inter-annotator agreement | 1 | 1 | 0.5 | 0 | 0.5 | 0.5 |
| F12 | Data privacy | 1 | 1 | 1 | 1 | 1 | 1 |
| F13 | Support for various languages | 1 | 1 | 1 | 1 | 1 | 1 |
| A1 | Support for querying | 1 | 0 | 0 | 0 | 0 | 1 |
| A2 | Server side crash tolerance | 1 | 0 | 0 | 0 | 0 | 1 |
| A3 | Client side crash tolerance | 1 | 0 | 0 | 0 | 0 | 1 |
| A4 | Web-based/distributed annotation | 1 | 1 | 1 | 1 | 1 | 1 |
| | **Total** | 25 | 18.5 | 15.0 | 19.5 | 17.5 | **20.5** |
| | **Score** | | 0.74 | 0.60 | 0.78 | 0.70 | **0.82** |

**Table 3.6:** Evaluation of *Sangrahaka* in comparison with other annotation tools using objective evaluation criteria

Using this modified set of $25$ criteria, we re-evaluated the top-scoring tools identified in the study by [Neves and Ševa, 2021] (WebAnno, BRAT, and FLAT), along with *Sangrahaka*. The evaluation results, illustrated in Table 3.6, highlight that *Sangrahaka* surpassed the other tools with a score of $0.82$, outperforming FLAT ($0.78$), WebAnno ($0.74$), and BRAT ($0.70$).

## 3.2 Semantic Annotation of Semi-structured Āyurveda Text

We now describe the application of *Sangrahaka* for a real-world semantic annotation task on the Āyurveda text Bhāvaprakāśanighaṇṭu.

### 3.2.1 Introduction

Sanskrit is one of the most prolific languages in the entire world, and text in Sanskrit far outnumber other classical languages. Consequently, with the advancement of natural language processing with the aid of computers, there has been a surge in the field of computational linguistics for Sanskrit over the last couple of decades. This has resulted in development of various tools such as the *Samsaadhanii* by [Kulkarni, 2016], *The Sanskrit Heritage Platform (SHP)* by [Goyal et al., 2012], *Sanskrit Sandhi and Compound Splitter (SSCS)* by [Hellwig and Nehrdich, 2018], *Sanskrit WordNet (SWN)* by [Kulkarni et al., 2010], etc. for linguistic tasks such as word segmentation, lemmatization, morphological generation, dependency parsing, etc. Despite this, many fundamental processing tasks such as anaphora resolution and named entity recognition that are needed for higher-order tasks such as discourse processing, are either not available or have a long way to go. Combined with the fact that Sanskrit is a morphologically rich language, for tasks such as machine translation, question answering, semantic labeling, discourse analysis, etc. there are no ready-to-use tools available.

A standard way of capturing knowledge from a text is through the use of *knowledge bases* (KB). It is a form of data repository that stores knowledge in some structured or semi-structured form. A *knowledge graph* (KG) is a particular form of knowledge base that uses the graph data structure to store knowledge. In a KG, nodes represent real-world *entities*, and edges represent *relationships* between these entities. Knowledge about these entities and relationships is typically stored in the form of triplets (*subject, predicate, object*) denoting the relationship predicate a subject has with an object. For example, (Pāṇini, Is-Author-Of, Aṣṭādhyāyī) captures the knowledge nugget 'Pāṇini is the author of Aṣṭādhyāyī'.

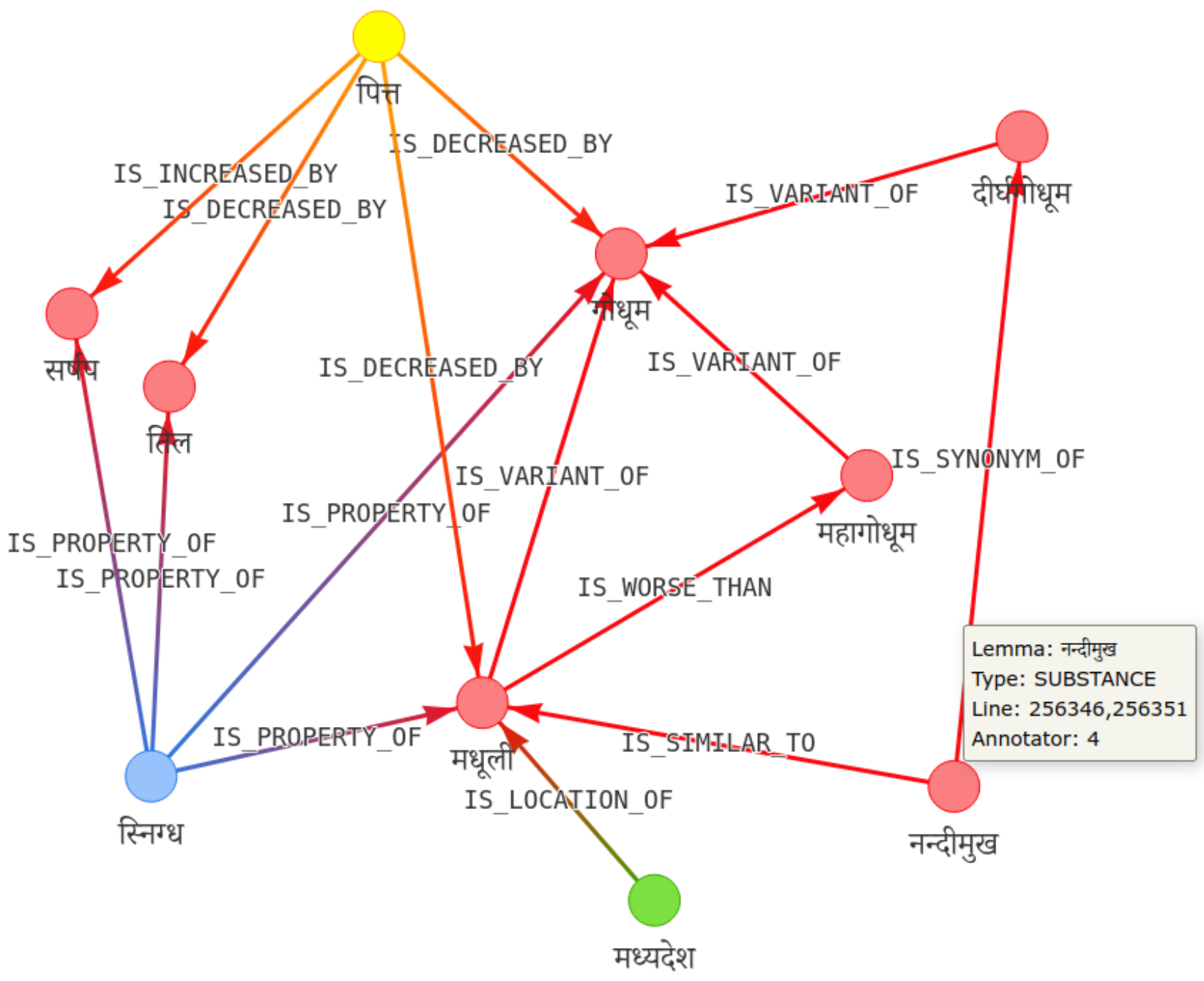

**Figure 3.6:** Example of a Knowledge Graph (KG).

An important usage of KGs is automated *question answering* (QA) where the task is to automatically find answers to questions posed in a natural language. It is an important high-level task in the fields of Information Retrieval (IR) and Natural Language Processing (NLP). Questions can be either from a specific closed domain (such as, say, manuals of certain products) or from the open domain (such as what Google and many other search engines attempt to do). Also, they can be factual (phrase-based or objective) or descriptive (subjective, such as 'Why' questions). Since its introduction by [Voorhees, 1999], one of the main approaches for the question answering task has been through use of knowledge bases [Hirschman and Gaizauskas, 2001, Kiyota et al., 2002, Yih et al., 2015].

Figure 3.6 shows an example of (a snippet of) a knowledge graph. The triplets are depicted visually. It contains several nodes (entities) such as madhūlī (मधूली), nandīmukha (नन्दीमुख), pitta (पित्त), snigdha (स्निग्ध), etc. and edges (relationships) including

'is Decreased by', 'is Property of', 'is Variant of', etc. The graph also shows properties associated with the entities and the relationships.

There exist various automated approaches for constructing knowledge bases from a corpus of text [Dong et al., 2014, Pujara and Singh, 2018, Mitchell et al., 2018, Wu et al., 2019]. These attempts are fairly successful for languages such as English where the state-of-the-art in NLP tools is more advanced. However, due to paucity of such tools in Sanskrit, automated construction of knowledge bases in Sanskrit, to the best of our understanding and knowledge, is only moderately successful. As described in Chapter 2 we previously attempted to automatically extract all human relationships from Itihāsa texts (Rāmāyaṇa and Mahābhārata) and synonym relationships from Bhāvaprakāśa. We reported that for an objective natural language query, the correct answer was present in the reported set of answers 50% of the times. We, however, could not report how accurately a triplet is automatically extracted, due to the lack of ground truth for the evaluation.

A more viable and accurate alternative of constructing knowledge bases is through the route of human annotation. In addition to information extracted from a corpus, the knowledge base may use information that is not directly mentioned in the corpus, such as world knowledge (for example, a person is a living being) or an ontology or a fact specific to the domain of the corpus. Human annotators are typically aware of the domain; although, depending on the task, they need not always be experts in the subject. For example, vāta (वात) has a general meaning as 'wind', but in the context of Āyurveda, it refers to the tridoṣa (त्रिदोष) by the name of vāta. This is not directly mentioned in every Āyurveda text, but any domain expert is aware of this fact.

In this work, we follow the human annotation process of creating a knowledge graph. We annotate three chapters: Dhānyavarga, Śākavarga and Māṃsavarga from the Bhāvaprakāśanighaṇṭu (भावप्रकाशनिघण्टु) portion from the Āyurveda text Bhāvaprakāśa (भावप्रकाश) as the corpus. Bhāvaprakāśa is one of the most prominent texts in Āyurveda, which is an important medical system developed in ancient India and is still in practice. A nighaṇṭu (निघण्टु) in the Indic knowledge system is a list of words, grouped into semantic and thematic categories and accompanied by relevant information about these words such as meanings, explanations or other annotations. It is analogous to a glossary in purpose, but differs in structure. In particular, the Bhāvaprakāśanighaṇṭu text, like most of Sanskrit literature, is in padya (verse) form. The text, while loosely following a theme or a structure, is still free flowing. Sanskrit literature contains a large number of such nighaṇṭu texts either as stand-alone books or as parts of other books.

The nighaṇṭu texts, owing to their partial structure are, therefore, amenable to construction of knowledge bases using human annotation. Further, since they contain a wealth of information, they are important resources for building knowledge bases that can be automatically questioned. A benefit of the presence of structure in nighaṇṭu texts is that annotators need not be domain experts as long as the structure is clear.

#### 3.2.1.1 Contributions

The contributions of this work are three-fold.

First, we describe a process of constructing a knowledge graph (KG) through manual annotation. This helps to capture the *semantic information* present in the text that is extremely difficult to do otherwise using automated language and text processing methods. The proposed annotation process also enables capturing relationships with entities that are not named directly in the text. We further discuss the curation process and the optimizations performed during the process of knowledge graph creation from the perspective of querying efficiency.

Second, through careful examination of the different types of entities and relationships mentioned in Bhāvaprakāśanighaṇṭu, we create a suitable ontology for annotating the text. We believe that this can be a good starting point for building an ontology for other Āyurveda texts, and in particular, glossaries.

Third, we annotate three complete chapters from the text (Dhānyavarga, Śākavarga and Māṃsavarga), and create a KG from the annotations. For this purpose, we de-ploy a customized instance of *Sangrahaka*, an annotation and querying framework developed by us previously [Terdalkar and Bhattacharya, 2021a]. We also create 31 query templates in English and Sanskrit to feed into the templatized querying interface, that aids users in finding answers for objective questions related to the corpus.

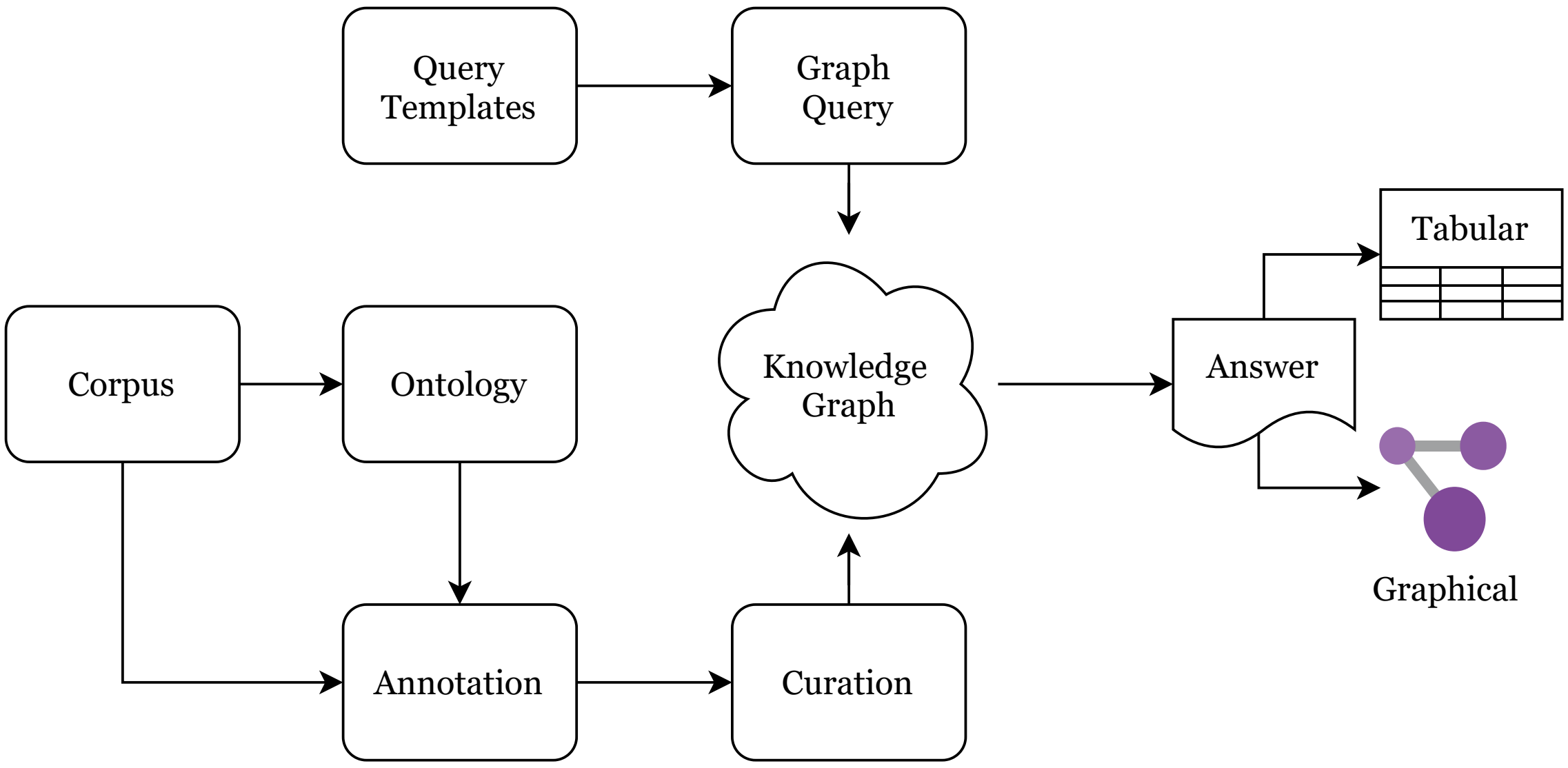

**Figure 3.7:** Workflow of semantic annotation for KG construction and querying

The system and the dataset can be accessed at `https://sanskrit.iitk.ac.in/ayurveda/`[1].

#### 3.2.1.2 Outline

Figure 3.7 shows the workflow of our proposed method. The first step after inspection of the corpus is of *ontology* creation. After creating a relevant ontology, i.e., specifying what kinds of relationships and entity types are there in the corpus, annotation is performed. Using the entities and relationships captured through annotation, a knowledge graph is constructed. The knowledge graph can be queried with the help of query templates to retrieve answers for templatized natural language questions. Answers are presented in both tabular and graphical formats.

The rest of the chapter is organized as follows. Section 3.2.2 motivates the problem of creating a knowledge graph using manual annotations. Section 3.2.5 de-scribes the annotation and curation process along with the construction of knowledge graph. Section 3.2.6 explains the querying mechanism.

[1]Please create an account and contact authors requesting access to annotation or querying interface.

### 3.2.2 Motivation for Manual Annotation

To the best of our knowledge, the state-of-the-art in Sanskrit NLP and IR is not advanced enough for automatic construction of knowledge bases from text. One of the first efforts towards automatic construction of knowledge graphs from Sanskrit text was made by [Terdalkar and Bhattacharya, 2019a]. The framework described does not yield results comparable to state-of-the-art models for English due to errors in various stages of the construction pipeline. As mentioned earlier, the success rate for even single relationships was not very high.

In this work, we discuss the issues with the Sanskrit state-of-the-art linguistic tools and the need for manual annotation for a *semantic* task such as automatic creation of knowledge graphs.

#### 3.2.2.1 Word Segmentation

Sanskrit texts make heavy use of compound words in the form of sandhi and samāsa. *Word segmentation*, that splits a given compound word into its constituents is, therefore, an important need in Sanskrit. Notable works in this area are *The Sanskrit Heritage Platform* (SHP) [Huet, 2009, Goyal et al., 2012], *Sanskrit Sandhi and Compound Splitter* (SSCS) [Hellwig and Nehrdich, 2018, Krishna et al., 2016, Krishna et al., 2021].

Treating the segmentation task as a combined splitting of both sandhi and samāsa, while useful, does not fit well into the pipeline described in Chapter 2, where the split output is then passed to a morphological analyser. An example is splitting of the word maharṣi as mahat + ṛṣiḥ, which while correct as a samāsa-split, if passed to a morphological analyser as two separate words, produces an analysis[2] of the word mahat independently that does not fit the semantics of the word or the context. Earlier [Terdalkar and Bhattacharya, 2019a], we applied *SSCS* followed by *SHP*. This results in a word such as rāmalakṣmaṇau getting split into two words rāma and lakṣmaṇau due to *SSCS* resulting in the word rāma getting assigned the vocative case by *SHP*. We had applied a heuristic to resolve these errors, where the grammatical analysis of the second word was copied to the first word as well. However, this heuristic would change the semantics of the word rāmalakṣmaṇau.

[2]Analysis: mahat `(n. sg. acc. | n. sg. nom.)`

#### 3.2.2.2 Morphological Analysis

Sanskrit is a highly inflectional language. In Sanskrit, words are categorized as subanta (noun-like)[3] and tiṅanta (verb-like). *Morphological analysis* is the task of identifying the *stem* (prātipadika or dhātu) of the given word form, along with other relevant linguistic information. Notable works in this area are by [Goyal et al., 2012] and [Kulkarni, 2016]. These tools perform the best when the input given is without sandhi. If, however, the input also contains splits of samāsa as generated by tools described in the previous section (Section 3.2.2.1), the morphological analysers treat it as a separate word, resulting in an analysis of the word that may be correct on the syntactic level, but not so in the context of the sentence. In the case of samāsa, if the morphological analyser is provided the text with hyphens ('-') separating the components, the analyser produces correct analysis. However, the treatment of word segmentation as a joint task, and the lack of differentiation between a sandhi and samāsa splits in the *SSCS* output makes this a difficult task.

#### 3.2.2.3 Other Linguistic Tasks

A dependency parser for Sanskrit from *Samsaadhanii* [Kulkarni, 2016] performs best when the sentences are in an anvaya order (prose form). Further, it is based on a fixed vocabulary and, therefore, when inflected forms of words from outside the vocabulary are encountered, it fails to parse the sentence. For example, a word śālidhānya is not present in the vocabulary, so a sentence containing that word does not get parsed successfully.

[3]Words that behave like nouns insomuch as that they exhibit gender, case, and number.

[Krishna et al., 2021] in their recent work claim to be able to perform poetry-to-prose linearization and dependency parsing. However, we have not been able to obtain the source code or a functional interface to evaluate it for our data (we contacted the authors).

Another hurdle in the poetry-to-prose linearization is that the sentence boundaries are often not clearly marked. In general, a semantically complete sentence may span over multiple verses. On the other hand, at times a verse may contain multiple sentences as well. This can be seen in the sample of 10 verses given in Section 3.2.3.1. Thus, extracting sentences with proper sentence boundaries is also a difficult task.

#### 3.2.2.4 Semantic Information Extraction

Extracting the semantics of a sentence is a very important step in the construction of a knowledge graph. Automatic KG construction frameworks for English such as [Auer et al., 2007, Suchanek et al., 2007] extract semantic information from various information sources including Wikipedia articles and info-boxes. One of the challenges faced in this task is that the same concept can be expressed in English in numerous ways, such as “birthplace” or “place of birth”. The issue of expressing a concept in more than one ways is extremely significant and much more severe for Sanskrit due to its semantic richness. In particular, the process of samāsa creates long and semantically rich words.

Table 3.7 highlights this phenomenon. The first column contains the concept while the second column enlists the words used in Dhānyavarga to express that concept. A “concept” captures the semantics of a word or a phrase.

It can be noted that even in the span of 90 verses, there are more than 10 different ways used to express the same concept ‘(a substance) decreases vāta’. In addition to that, the word vāta itself can be part of another compound, coupled with other words as can be seen in the example ‘decreases vāta and pitta’. There are more than 5 usages of this complex concept, which is a superset of the earlier concept. Moreover, these are not the only ways in which the concept of increasing vāta is expressed.

**Table 3.7:** Semantic variations due to richness of Sanskrit through examples from Dhānyavarga.

| Concept | Words or Phrases |
|---|---|
| increases bala | balya, balada, balāvaha, balaprada, balakara, balakṛt |
| increase vāta | vātala, vātakṛt, vātakara, vātajanaka, vātajananī, vātātikopana, vātaprakopaṇa, vātavardhana, vātakopana |
| decreases pitta | pittaghna, pittapraṇāśana, pittapraśamana, pittahara, pittaghnī, pittāpaha, pittajit, pittahṛt, pittanut, pittavināśinī |
| decreases vāta and pitta | vātapittaghna, pittavātaghna, pittavātavibandhakṛt, vātapittahara, vātapittahṛt |

There are numerous other words that can combine with the word vāta in the form of samāsa to indicate the concept of decrement for multiple entities at the same time. Moreover, in such cases, where a samāsa is used, the order of vāta and pitta could be reversed as well. Further, this list is not exhaustive for a specific concept, and the number of ways to denote the same concept can become computationally intractable.

One can observe that there are some common suffixes used in similar concepts. However, firstly, there is no exhaustive list of suffixes available associated with a particular concept. Second, the suffixes have different concepts in different contexts. For example, the suffix -ghna (-घ्न) in the context of Ayurveda, or specifically of tridoṣa, means 'to decrease', e.g., pittaghna (पित्तघ्न). The same suffix in the context of a person may mean 'to kill', e.g., śatrughna (शत्रुघ्न) – one who kills his enemies (śatru).

Thus, using a fixed set of suffixes may not be a feasible solution.

To the best of our knowledge, there is no existing system for Sanskrit that can extract such semantic information in either a generic sense or in a specific context. Amarakoṣa Knowledge Net [Nair and Kulkarni, 2010] and Sanskrit WordNet [Kulkarni et al., 2010] are also limited in their scope. For example, none of the words listed in the Table 3.7 to express the concept of 'increasing bala' can be directly found in either of these two resources. These are examples of samāsa or words formed from bala through the application of taddhita or kṛdanta suffixes. Due to the recursively generative grammar of Sanskrit, it is impossible to produce an exhaustive list of all possible derivations from a word.

#### 3.2.2.5 Need for Annotation

Issue of compounding errors is relevant to any NLP pipeline, where individual parts of the pipeline have their own error rates. The success rate of the entire pipeline, being a multiplicative factor of the individual success rates (since all the parts have to be accurate for the entire task to be accurate), is significantly lower. Thus, the pipeline for the automated question answering task that requires modules such as word segmentation, morphological analysis, part-of-speech tagging, dependency parsing, etc. has a very low accuracy. Further, the lack of semantic analysis tools or systems is a major hurdle in semantic tasks such as construction of knowledge graphs. Thus, even if the accuracy of the individual parts are improved significantly, the final semantic labeling remains a bottleneck.

We highlight this fact by taking an example of the first $\sim 10\%$ of the Dhānyavarga, i.e., 10 verses corresponding to 21 lines. We have manually segmented the words in these lines and also converted the sentences to anvaya order. The first 10 verses correspond to a total of 14 prose sentences. The original text in verse format, in the sandhi-split format, and in anvaya format, is given in Section 3.2.3.1.

There are a total of 35 occurrences of sandhi and 50 occurrences of samāsa in the text. *SSCS* is able to identify 34 of the sandhi (with an accuracy of 0.97) and 34 occurrences of samāsa correctly (with an accuracy of 0.68). However, the tool does not differentiate a sandhi from a samāsa. Therefore, when passed to the *SHP* it is likely to obtain incorrect analysis.

A single word-form in Sanskrit can have numerous valid morphological analyses. If there are $N$ words in a sentence, and every word has $a_i$ analyses possible, then there are $\Pi_{i=1}^{N} a_i$ possible combinations for the correct analysis of the sentence. *SHP* and *Samsaadhanii* both rank these solutions based on various linguistic features, and after pruning the unlikely ones, present the feasible solutions. For automatic processing pipelines, a particular choice of the solution is required, and the solution presented as the *best* by the tools, i.e., the first solution, is a natural choice. Thus, we present the evaluation by choosing the first solution.

We pass the manually created sandhi-split corpus through *SHP* for morphological analysis.[4] There are an average of $9$ solutions per line (ranging from $0$ to $72$) reported. We evaluate based on the first reported solution.

There are $103$ words, after manually splitting sandhi. *SHP* could not analyse $21$ words, and wrongly analysed $14$ words, resulting in an overall accuracy of $0.66$. Further, *SHP* split $34$ words, of which $8$ were incorrect splits, resulting in an accuracy of $0.76$ for samāsa-split.

Additionally, we pass the anvaya-order sentences to the dependency parser tool by [Kulkarni, 2016]. We also manually add missing verbs (adhyāhāra forms such as asti, santi, etc.) due to that being a requirement of the parser. Without samāsa split markers, the dependency parser manages to parse only 1 out of 14 sentences, while with the samāsa markers, it can parse 6 out of 14. Out of the 6 sentences that produce a dependency parse tree, 4 are simple 3-word sentences (sentences 2, 3, 4, 6 in Section 3.2.3.1). In the other 2 instances (sentences 7, 10), errors were found in the dependency parse trees.

For example, consider a line from śloka 2:

> Sanskrit: ⟨ कङ्ग्वादिकं क्षुद्रधान्यं ⟩ ⟨ तृणधान्यञ्च तत्स्मृतम् ⟩
>
> IAST: ⟨ kaṅgvādikaṃ kṣudradhānyaṃ ⟩ ⟨ tṛṇadhānyañca tatsmṛtam ⟩
>
> Meaning: kaṅgu etc. are types of kṣudradhānya. It (kṣudradhānya) is also called tṛṇadhānya.
>
> Anvaya: क्षुद्रधान्यम् कङ्ग्वादिकम् (अस्ति)। तत् तृणधान्यम् च स्मृतम् (अस्ति)।

There are two sentences in this line, as can be seen by the boundary markers and anvaya text. Figure 3.8 shows the dependency parse for these two sentences. The dependency parse for the first dependency tree is correct. However, even for the sentences that get a correct dependency parse, the dependency relations we get are kartā and kartṛsamānādhikaraṇa which, while grammatically correct, still do not help in capturing the *semantic* concept of the sentence that kaṅgu is a *type* of kṣudradhānya. Also, tat in the second sentence is an *anaphora* of kṣudradhānya from the first sentence. Thus, the intended relation that tṛṇadhānya is a synonym of kṣudradhānya cannot be extracted without a module for *anaphora resolution*. Yet again, to the best of our knowledge, there is no such co-reference resolution system for Sanskrit.

[4] We keep a timeout of 60 seconds, within which, if the analysis is not found, we report the analysis as missing, i.e., $0$ solutions for that line.

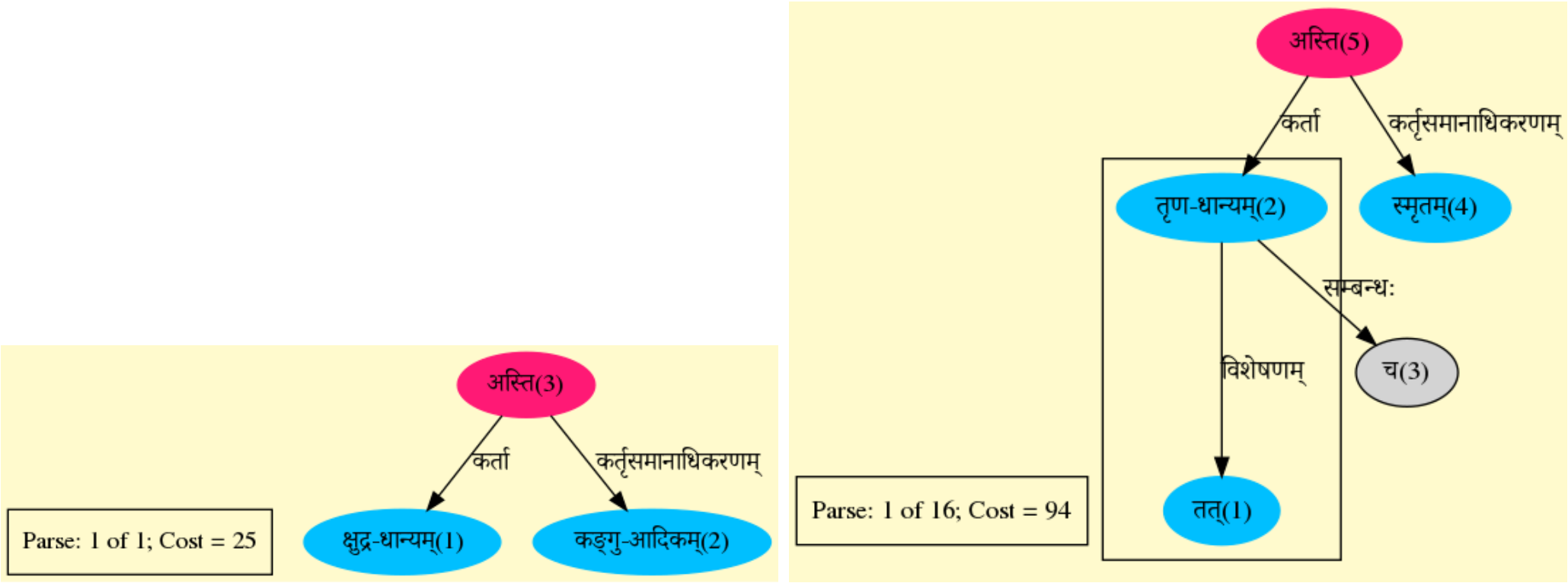

**Figure 3.8:** Dependency parse trees for sentences from śloka 2.

More importantly, no existing tool has a capability of performing *semantic* tasks, which are a requirement for knowledge extraction. Manual annotation, therefore, is the only way to capture the semantic relations. In addition, it bypasses the entire NLP pipeline and, thus, has a high potential for creating a question answering system that is much more accurate and reliable than a system based on automatically created knowledge graphs.

Another prevalent issue is the lack of datasets for training and evaluation of tasks such as question answering or creation of knowledge bases. Creation of knowledge bases through manual annotation is, thus, of utmost importance both for the actual task of question answering and for further research in the field, including automated knowledge base construction since these may act as ground-truth benchmark datasets for evaluation of future automated tools.

### 3.2.3 Corpus

Bhāvaprakāśanighaṇṭu is the nighaṇṭu portion of Bhāvaprakāśa. It contains a list and description of various medicinally relevant plants, flowers, fruits, animals, grains, animal products, metals, prepared substances, etc. These are divided into 23 chapters called vargas.

A general structure followed in the Bhāvaprakāśanighaṇṭu is as follows,

- Substances are semantically grouped in chapters. For example, all grains appear in the chapter Dhānyavarga, all vegetables appear in the chapter Śākavarga and all types of meats appear in the chapter Māṃsavarga.
- Each chapter contains several virtual sections pertaining to a single substance. Only when a substance has been described in entirety, discussion about another substance starts.[5]
- Each section about a substance has the following information:
  - Synonyms, if any, of the substance
  - Properties, e.g., color, smell, texture
  - Effects, e.g., effect on tridoṣa (vāta, pitta, and kapha)
  - Symptoms and diseases treated or cured by the substance
  - Variants, if any, of the substance
  - Properties of each variant, and their distinguishing characteristics
  - Comparison between the variants, if possible
  - Time and location where the substance is found or grown, if relevant
- The order of information components about a substance within a section may vary.

The entire Bhāvaprakāśanighaṇṭu contains 2087 verses, corresponding to 4201 lines. We have chosen three chapters: Dhānyavarga, a chapter about grains, Śākavarga, a chapter about vegetables and Māṃsavarga, a chapter about meats. Together, these contain a wide variety of entity types and relations. In total, these three chapters contain 340 verses, corresponding to 690 lines.

[5]There is, however, no indication in the original text that a section/substance has ended, and a new one has started. It must be inferred by reading the text.

#### 3.2.3.1 Sample of Text from Dhānyavarga

We present here, an extract from Dhānyavarga used in Section 3.2.2. Table 3.8 contains the first 10 verses from Dhānyavarga and a version with sandhi resolved. The sentence boundaries are denoted using '⟨' and '⟩' markers.

**Table 3.8:** First 10 verses from Dhānyavarga of Bhāvaprakāśanighaṇṭu

| **Original Text** | **Sandhi Split** |
|---|---|
| ⟨ शालिधान्यं व्रीहिधान्यं शूकधान्यं तृतीयकम्<br>शिम्बीधान्यं क्षुद्रधान्यमित्युक्तं धान्यपञ्चकम् १ ⟩ | शालिधान्यम् व्रीहिधान्यम् शूकधान्यम् तृतीयकम्<br>शिम्बीधान्यम् क्षुद्रधान्यम् इति उक्तम् धान्यपञ्चकम् १ |
| ⟨ शालयो रक्तशाल्याद्या ⟩ ⟨ व्रीहयः षष्टिकादयः ⟩<br>⟨ यवादिकं शूकधान्यं ⟩ ⟨ मुद्गाद्यं शिम्बिधान्यकम् ⟩<br>⟨ कङ्ग्वादिकं क्षुद्रधान्यं ⟩ ⟨ तृणधान्यञ्च तत्स्मृतम् २ ⟩ | शालयः रक्तशाल्याद्याः व्रीहयः षष्टिकादयः<br>यवादिकम् शूकधान्यम् मुद्गाद्यम् शिम्बिधान्यकम्<br>कङ्ग्वादिकम् क्षुद्रधान्यम् तृणधान्यम् च तत् स्मृतम् २ |
| ⟨ कण्डनेन विना शुक्ला हैमन्ताः शालयः स्मृताः ३ ⟩ | कण्डनेन विना शुक्लाः हैमन्ताः शालयः स्मृताः ३ |
| ⟨ रक्तशालिः सकलमः पाण्डुकः शकुनाहृतः<br>सुगन्धकः कर्दमको महाशालिश्च दूषकः ४ | रक्तशालिः सकलमः पाण्डुकः शकुनाहृतः<br>सुगन्धकः कर्दमकः महाशालिः च दूषकः ४ |
| पुष्पाण्डकः पुण्डरीकस्तथा महिषमस्तकः<br>दीर्घशूकः काञ्चनको हायनो लोध्रपुष्पकः ५ | पुष्पाण्डकः पुण्डरीकः तथा महिषमस्तकः<br>दीर्घशूकः काञ्चनकः हायनः लोध्रपुष्पकः ५ |
| इत्याद्याः शालयः सन्ति बहवो बहुदेशजाः ⟩<br>⟨ ग्रन्थविस्तरभीतेस्ते समस्ता नात्र भाषिताः ६ ⟩ | इत्याद्याः शालयः सन्ति बहवः बहुदेशजाः<br>ग्रन्थविस्तरभीतेः ते समस्ताः न अत्र भाषिताः ६ |
| ⟨ शालयो मधुराः स्निग्धा बल्या बद्धाल्पवर्चसः<br>कषाया लघवो रुच्याः स्वर्या वृष्याश्च बृंहणाः<br>अल्पानिलकफाः शीताः पित्तघ्ना मूत्रलास्तथा ७ ⟩ | शालयः मधुराः स्निग्धाः बल्याः बद्धाल्पवर्चसः<br>कषायाः लघवः रुच्याः स्वर्याः वृष्याः च बृंहणाः<br>अल्पानिलकफाः शीताः पित्तघ्नाः मूत्रलाः तथा ७ |
| ⟨ शालयो दग्धमृज्जाताः कषाया लघुपाकिनः<br>सृष्टमूत्रपुरीषाश्च रूक्षाः श्लेष्मापकर्षणाः ८ ⟩ | शालयः दग्धमृज्जाताः कषायाः लघुपाकिनः<br>सृष्टमूत्रपुरीषाः च रूक्षाः श्लेष्मापकर्षणाः ८ |
| ⟨ कैदारा वातपित्तघ्ना गुरवः कफशुक्रलाः<br>कषायाश्चाल्पवर्चस्का मेध्याश्चैव बलावहाः ९ ⟩ | कैदाराः वातपित्तघ्नाः गुरवः कफशुक्रलाः<br>कषायाः च अल्पवर्चस्काः मेध्याः च एव बलावहाः ९ |
| ⟨ स्थलजाः स्वादवः पित्तकफघ्ना वातवह्निदाः<br>किञ्चित्तिक्ताः कषायाश्च विपाके कटुका अपि १० ⟩ | स्थलजाः स्वादवः पित्तकफघ्नाः वातवह्निदाः<br>किञ्चिद् तिक्ताः कषायाः च विपाके कटुकाः अपि १० |

#### 3.2.3.2 Poetry-to-Prose Conversion of Verses from Table 3.8

We next list the prose version of the verses listed in Table 3.8 above.

1. शालिधान्यम् व्रीहिधान्यम् तृतीयकम् शूकधान्यम् शिम्बीधान्यम् क्षुद्रधान्यम् इति धान्यपञ्चकम् उक्तम् (अस्ति)।
2. शालयः रक्तशाल्याद्याः (सन्ति)।
3. व्रीहयः षष्टिकादयः (सन्ति)।
4. शूकधान्यम् यवादिकम् (अस्ति)।
5. शिम्बिधान्यकम् मुद्गाद्यम् (अस्ति)।
6. क्षुद्रधान्यम् कङ्ग्वादिकम् (अस्ति)।
7. तत् तृणधान्यम् च स्मृतम् (अस्ति)।
8. कण्डनेन विना शुक्लाः हैमन्ताः (च) शालयः स्मृताः (सन्ति)।
9. शालयः रक्तशालिः सकलमः पाण्डुकः शकुनाहृतः सुगन्धकः कर्दमकः महाशालिः दूषकः पुष्पाण्डकः पुण्डरीकः महिषमस्तकः दीर्घशूकः काञ्चनकः हायनः लोध्रपुष्पकः च तथा इत्याद्याः बहवः बहुदेशजाः सन्ति।
10. ग्रन्थविस्तरभीतेः ते समस्ताः अत्र न भाषिताः (सन्ति)।
11. शालयः मधुराः स्निग्धाः बल्याः बद्धाल्पवर्चसः कषायाः लघवः रुच्याः स्वर्याः वृष्याः बृंहणाः अल्पानिलकफाः शीताः पित्तघ्नाः तथा मूत्रलाः च (सन्ति)।
12. दग्धमृज्जाताः शालयः कषायाः लघुपाकिनः सृष्टमूत्रपुरीषाः रूक्षाः श्लेष्मापकर्षणाः च (सन्ति)।
13. कैदाराः (शालयः) वातपित्तघ्नाः गुरवः कफशुक्रलाः कषायाः अल्पवर्चस्काः मेध्याः बलावहाः च एव (सन्ति)।
14. स्थलजाः (शालयः) स्वादवः पित्तकफघ्नाः वातवह्निदाः किञ्चिद् तिक्ताः कषायाः विपाके कटुकाः च

### 3.2.4 Ontology

Through meticulous examination of various chapters from the Bhāvaprakāśanighaṇṭu, including Dhānyavarga, Śākavarga, and Māṃsavarga, we have meticulously constructed an ontology. This ontology serves as a comprehensive framework that adheres to the fundamental principles of Āyurveda. By incorporating semantic intricacies, our ontology will facilitate the precise annotation of the entire Bhāvaprakāśanighaṇṭu.

Ontology is structured in a hierarchical manner. The top-level categories for node types cover broad concepts like properties of substances, substances, diseases, treatments, preparations, location, time and so on. The granularity of subcategories varies in each top-level category. Figure 3.9 illustrates the richness of the hierarchy for a single top-level category, namely, 'Property' of a medicinal substance. The ontology exemplifies the meticulous thought and deliberation invested in its categorization and structure. We have made every effort to remain true to the authentic terminology and concepts derived from Āyurveda. One notable example is the creation of ten distinct categories to represent ten pairs of properties, commonly referred to as "Gurvādi" properties (literally, 'Guru and so on'), ensuring their accurate representation within the ontology.

| node label hierarchy | | | sanskrit | description |
|---|---|---|---|---|
| PROPERTY | | | गुणः | Property |
| | EXTERNAL_PROPERTY | | बाह्यगुणः | External Property |
| | | VARNA_PROPERTY | वर्ण-गुणः | Colour |
| | | GANDHA_PROPERTY | गन्ध-गुणः | Smell |
| | | SHABDA_PROPERTY | शब्द-गुणः | Sound |
| | | SPARSHA_PROPERTY | स्पर्श-गुणः | Touch |
| | | AARDRA_SHUSHKA_PROPERTY | आर्द्र-शुष्क-गुणः | Wet-Dry Property |
| | | SIZE_PROPERTY | आकारमान-गुणः | Size (Large, Small etc) |
| | | SHAPE_PROPERTY | आकार-गुणः | Shape |
| | | RUUPA_PROPERTY | रूप-गुणः | Ruupa Property |
| | INTERNAL_PROPERTY | | आन्तरिकगुणः | Internal Property |
| | | RASA_PROPERTY | रस-गुणः | Taste |
| | | ANURASA_PROPERTY | अनुरस-गुणः | Anurasa (Secondary Taste) |
| | | VIIRYA_PROPERTY | वीर्य-गुणः | Viirya (Potency) |
| | | VIPAAKA_PROPERTY | विपाक-गुणः | Vipaaka |
| | | DHARMA_PROPERTY | धर्म-गुणः | Acidic - Alkaline Property |
| | | LAGHU_GURU_GUNA | लघु-गुरु-गुणः | Heavy - Light Guna |
| | | MANDA_TIIKSHNA_GUNA | मन्द-तीक्ष्ण-गुणः | Dull - Sharp Guna |
| | | SHIITA_USHNA_GUNA | शीत-उष्ण-गुणः | Cold - Hot Guna |
| | | SNIGDHA_RUUKSHA_GUNA | स्निग्ध-रूक्ष-गुणः | Unctuous - Non-unctous Guna |
| | | SHLAKSHNA_KHARA_GUNA | श्लक्ष्ण-खर-गुणः | Smooth - Rough Guna |
| | | SAANDRA_DRAVA_GUNA | सान्द्र-द्रव-गुणः | Solid - Fluid Guna |
| | | MRUDU_KATHINA_GUNA | मृदु-कठिन-गुणः | Soft - Hard Guna |
| | | STHIRA_CHALA_GUNA | स्थिर-चल-गुणः | Immobile - Mobile Guna |
| | | SUUKSHMA_STHUULA_GUNA | सूक्ष्म-स्थूल-गुणः | Fine - Bulky Guna |
| | | VISHADA_PICCHILA_GUNA | विशद-पिच्छिल-गुणः | Clear - Slimey Guna |
| | EFFECT_PROPERTY | | प्रभाव-गुणः | Effect Property |
| | | BAAHYA_PRABHAAVA_PROPERTY | बाह्य-प्रभाव-गुणः | External Effect |
| | | AABHYANTARA_PRABHAAVA_PROPERTY | आभ्यन्तर-प्रभाव-गुणः | Internal Effect |

**Figure 3.9:** Example of hierarchical node ontology for a single top-level category 'Property' (Guṇaḥ).

The decision to add a certain entity type or relation type is made based on the importance of the concept, frequency of its occurrence, and nature of frequently asked questions.

For example, the concept of vāta, pitta and kapha, collectively referred to as tridoṣa or fundamental elements (humors) of the body, is central to Āyurveda. Consequently, queries such as "What effect does a substance X have on a (one of the) tridoṣa?" is one of the most fundamental and common information requirement about the substance. Therefore, we have a special category Śārīrika-Doṣa to represent the three entities vāta, pitta and kapha. Any occurrence of these words or their synonyms, e.g., śleṣman, a synonym of a kapha, results in the creation of an entity of type Śārīrika-Doṣa.

Similarly, there are three entities corresponding to the mental or psychological attributes, namely, satva, rajas and tamas. We have a category Mānasika-Doṣa to represent these three entities. Both of these categories are nested under the broader category Doṣa, which in turn is nested under the category 'Part of Body'.

The type of effect any substance has on each of the śārīrika-doṣa is broadly an increment or a decrement. However, it is important to note that within these categories, there exist nuances associated with the nature of increment or decrement. In particular, Prakopa of a particular doṣa means an increment that is harmful. Therefore, to capture these nuances, we have identified four relations is Increased by, is Decreased by, is Vitiated by and is Passivised by. Additionally, there are subtle differences between decrement and non-increment, so we also have negation of each of these categories.

Overall, the ontology encompasses 300 entity types and 320 relationship types. Once the ontology has been finalized, the next step is annotation.

### 3.2.5 Annotation Process

Annotation has been done with the purpose of building a knowledge graph (KG). The annotators encompass individuals with a basic understanding of Sanskrit and Ayurveda. We fix the unit for annotation to be a line from a verse (śloka). We collect annotations of two types — *entity* and *relation* — described in detail in Section 3.2.5.1 and Section 3.2.5.2 respectively.

The corpus interface from *Sangrahaka* is capable of displaying extra information about each line. We use this feature to display word segmentation and morphological analysis of the text produced by *SSCS* and *SHP*, which can potentially help the annotators. Figure 3.10 shows a sample text from Dhānyavarga with linguistic infor-

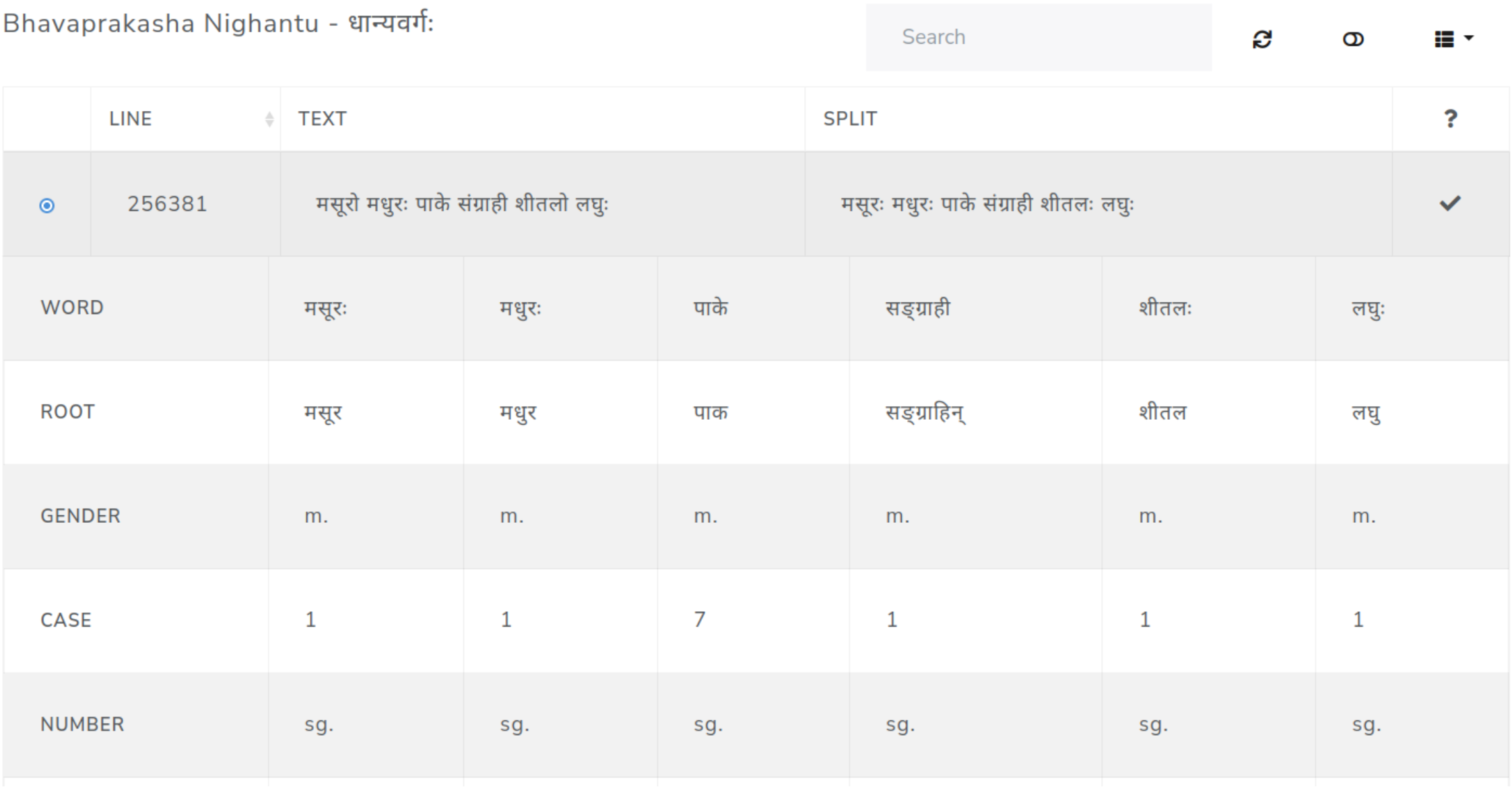

Bhavaprakasha Nighantu - धान्यवर्गः

Search

| | LINE | TEXT | SPLIT | ? |
|---|---|---|---|---|
| ⊙ | 256381 | मसूरो मधुरः पाके संग्राही शीतलो लघुः | मसूरः मधुरः पाके संग्राही शीतलः लघुः | ✓ |

| WORD | मसूरः | मधुरः | पाके | सङ्ग्राही | शीतलः | लघुः |
|---|---|---|---|---|---|---|
| ROOT | मसूर | मधुर | पाक | सङ्ग्राहिन् | शीतल | लघु |
| GENDER | m. | m. | m. | m. | m. | m. |
| CASE | 1 | 1 | 7 | 1 | 1 | 1 |
| NUMBER | sg. | sg. | sg. | sg. | sg. | sg. |

**Figure 3.10:** Sample text from Dhānyavarga with linguistic information

mation.

An annotator goes through the lines assigned to her and for each line, identifies the entities as well as the relationships between the entities appearing in it.

#### 3.2.5.1 Entity Annotation

Entities correspond to nodes in the knowledge graph. When a word that represents an entity is encountered, its *lemma* (prātipadika) and the entity type it belongs to are identified, and the entity is marked.

As an example, consider the following line from śloka-31 of Dhānyavarga:

Devanagari: गोधूमः सुमनोऽपि स्यात्त्रिविधः स च कीर्त्तितः।

IAST: godhūmaḥ sumano'pi syāttrividhaḥ sa ca kīrttitaḥ.

Meaning: Godhūma (wheat) is also called Sumana, and it is said to be of three kinds.

Here, there are two entities, godhūma and sumana, both of type "Substance". An entity needs to be added explicitly only the first time it is encountered.

In a case where a samāsa is used to indicate an effect on an entity, and the relation fits one of the relationship types, a relevant word (pada) from the segmentation (vigraha) of samāsa is used. For example, consider the following line from śloka-33:

Devanagari: गोधूमः मधुरः शीतो वातपित्तहरो गुरुः।

IAST: godhūmaḥ madhuraḥ śīto vātapittaharo guruḥ.

Meaning: Godhūma is sweet, cold, hard to digest and removes (decreases) vāta and pitta.

Here, vātapittaharaḥ is a single word, which uses samāsa, to indicate that vāta and pitta are reduced by godhūma. Therefore, vātapittahara will not be added as an entity; instead the entities vāta and pitta are recognized.

#### 3.2.5.2 Relation Annotation

Relations correspond to edges in the knowledge graph. A relation, which fits one of the relationship types from the ontology, is identified by interpreting the śloka. *Subject* and *Object* for this relation are then identified. Relations, where extra semantic information is known, such as madhura is known to be a rasa, are endowed with that extra information.

Consider the two examples of lines mentioned in the previous section (Section 3.2.5.1), śloka-31 and śloka-33. Following relations are added based on these two lines:

sumana $\vdash$ is Synonym of $\rightarrow$ godhūma

madhura $\vdash$ is (rasa) Property of $\rightarrow$ godhūma

śīta $\vdash$ is Property of $\rightarrow$ godhūma

vāta $\vdash$ is Decreased by $\rightarrow$ godhūma

pitta $\vdash$ is Decreased by $\rightarrow$ godhūma

guru $\vdash$ is Property of $\rightarrow$ godhūma

It should be noted that neither the subject nor the object may be present as words in the line that mentions a relationship about it. Consider, for example, the next line of the śloka-33:

Devanagari: कफशुक्रप्रदो बल्यः स्निग्धः सन्धानकृत्सरः।

IAST: kaphaśukraprado balyaḥ snigdhaḥ sandhānakṛtsaraḥ.

Meaning: (Godhūma) increases kapha, śukra, bala, is snigdha, sandhānakṛt (helps in joining broken bones) and laxative.

Here, the description of properties of godhūma (from previous line) is continued. Therefore, one of the relations added is

kapha ⊢ is Increased by → godhūma

This relation has godhūma as *Object*, although it is not present in the line itself.

#### 3.2.5.3 Unnamed Entities

On occasions, it may happen that an entity is referenced by its properties only, and it is not named at all in the text. Consider the following line from śloka-39:

Devanagari: मुद्गो बहुविधः श्यामो हरितः पीतकस्तथा।

IAST: mudgo bahuvidhaḥ śyāmo haritaḥ pītakastathā.

Meaning: Mudga is of various types – black, green, and yellow.

Thus, there are three colored variants of the substance mudga, but they are not named. In such a case, we create three *unnamed entities* (denoted by X-prefixed nodes) with entity type "Substance", same as that of mudga to refer to the three varieties. Each of these entities is given a unique identifier. The unique identifier is a combination of the unnamed entity number and the line number it occurs in. Thus, if the line number is 256358, the black variant is given the identifier X1-256358. Similarly, the green variant is identified as X2-256358 while the yellow variant is identified as X3-256358.

To describe these variants, three relations are added as well:

śyāma ⊢ is (varṇa) Property of → X1-256358

harita ⊢ is (varṇa) Property of → X2-256358

pīta ⊢ is (varṇa) Property of → X3-256358

The utility of such annotations becomes clear when these unnamed entities are later referred to in another line or another verse.

The next line of śloka-39 reads

Devanagari: श्वेतो रक्तश्च तेषान्तु पूर्वः पूर्वो लघुः स्मृतः ॥३९॥

IAST: śveto raktaśca teṣāntu pūrvaḥ pūrvo laghuḥ smṛtaḥ. ||39||

Meaning: ... white and red. Among them, each is successively easier to digest.

The word teṣām here refers to the five varieties of mudga, and gives a relation between them. So, we get two new unnamed entities in this line, `X1-256359` and `X2-256359` (note how `X1` and `X2` are re-used but with different line numbers).

We also get a total of four new relations to capture the successive ease in digestion properties:

`X1-256358` $\vdash$ is Better (in property laghu) than $\rightarrow$ `X2-256358`

`X2-256358` $\vdash$ is Better (in property laghu) than $\rightarrow$ `X3-256358`

`X3-256358` $\vdash$ is Better (in property laghu) than $\rightarrow$ `X1-256359`

`X1-256359` $\vdash$ is Better (in property laghu) than $\rightarrow$ `X2-256359`

For the purpose of querying, the anonymous nodes are treated like any other node.

#### 3.2.5.4 Auto-complete Suggestions

We have enhanced the annotation interface from *Sangrahaka* to improve user experience with Sanskrit text by adding transliteration-based suggestions. There are numerous standard schemes for *Devanagari* transliteration[6]. Whenever a *Devanagari* entity is annotated, we use *indic-transliteration* package [Sanskrit programmers, 2021] to transliterate it into various available schemes. We maintain an index with all the transliterations. Now, when a user enters any text, we query our index and return all suggestions that match with the lower-cased version of the user text. For example, consider a word in Devanagari 'माष', which transliterates into 'mASa' (HK), 'mASha' (ITRANS), 'māṣa' (IAST), 'maa.sa' (Velthuis), 'mARa' (WX) and 'mAza' (SLP1). Now, a user may enter at least 3 starting characters from any of the scheme, e.g., 'mas', 'maa', 'maz', 'mar' etc. and the Devanagari word 'माष' will be suggested. The index is maintained globally. So, once an entity is entered by any annotator, the completions for that entity become available to all annotators.

[6]https://en.wikipedia.org/wiki/Devanagari_transliteration

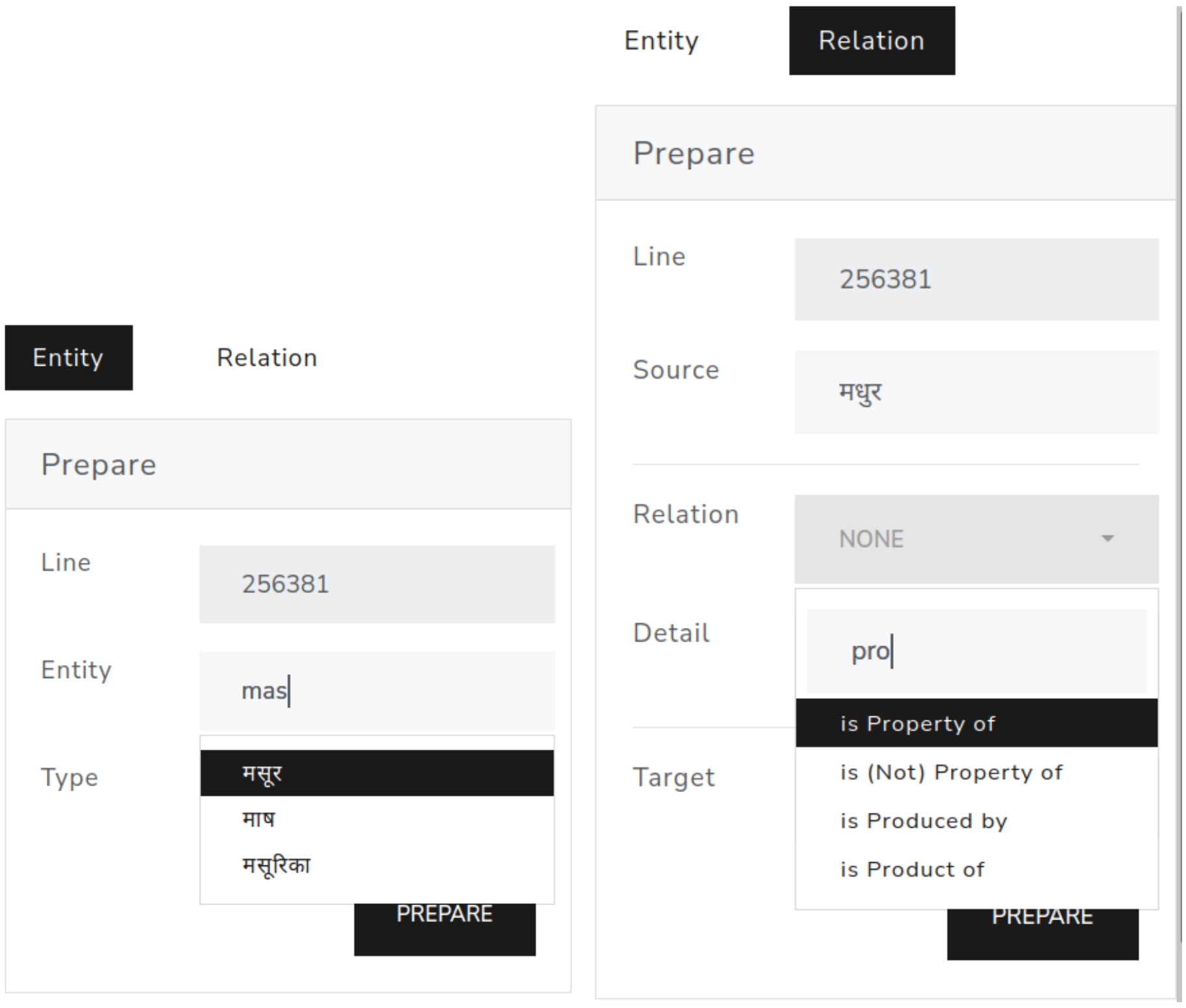

**Figure 3.11:** Modified annotation interface with multi-transliteration-based suggestions

These suggestions are enabled to all text input fields, namely, entity annotation, relationship annotations and querying interface.

Figure 3.11 shows the modified annotation interface with auto-complete suggestions.

#### 3.2.5.5 Curation

After the annotation step, before construction of the knowledge graph, a thorough curation step is required to resolve errors or inconsistencies that may inadvertently creep up during the annotation process.

**3.2.5.5.1 Equivalent Entities** The linguistic information that we have added with the corpus is supposed to serve as a guideline for the annotation. However, since this information is generated using automated tools, there might be errors. For example, the word grāhī (ग्राही) refers to substances that have a property of absorbing liquid and increasing digestive power. The reported prātipadika of this word, automatically generated as the first result by *SHP*, is grāha (ग्राह) instead of grāhin. An annotator, by oversight, may mark the incorrect lemma. Additionally, for substance names in feminine gender, which also have this property, an adjective grāhiṇī (ग्राहिणी) is used. The correct prātipadika in this case would be grāhiṇī. The node refers to the same property. So, semantically they are equivalent to each other, and ideally should be captured using a single name. These instances are common for properties of substances.

To address this issue, we add a relation is Synonym of between these entities. This, in conjunction with the optimization mechanism described in Section 3.2.5.6, tackles the issue of equivalent entities.

**3.2.5.5.2 Inconsistent Node Categories** There may be differences of opinions between annotators regarding which category a particular node should belong to. For example, an entity jvara (ज्वर) refers to fever. This entity was marked as a *Symptom* by some annotators and as a *Disease* by the others. Such cases were resolved through discussion among the curators.

**3.2.5.5.3 Missing Node Categories** The framework allows entities to be mentioned in the relationships without being added as entities. While care was taken to always add entities before marking relationships involving those entities, there may still be instances of human error, where an annotator may forget to mark an entity. We created a set of inference rules to infer as many instances of such occurrences as possible. For example, if an entity is marked as a *source* of the relation is Property of, without having been added as an entity, we can automatically create that entity by assigning the category '*Property*' to it.

#### 3.2.5.6 Symmetric Relationships

The relation is Synonym of is symmetric, i.e., if A is a synonym of B, then, by definition, B is also a synonym of A. A query can be made using any of the synonyms, and the system should still be able to return the correct answer.

Suppose, $S_1, S_2 \ldots, S_N$ are $N$ synonyms of a substance. If the synonym group is completely captured, then a user should be able to query using any synonym and still get the desired result. Properties of the substance corresponding to this synonym group can also be scattered across $S_i$'s. Say, there are $M$ properties $P_1, P_2, \ldots, P_M$, and some of the relations are $P_1$ is Property of $S_1$, $P_2$ is Property of $S_4$, $P_3$ is Property of $S_3$, and so on. Now, if we want to query whether substance $S_2$ has property $P_1$ but we search using the name of the substance as $S_2$, a direct query will not work, as there is no direct relation between the nodes $S_2$ and $P_1$. For the correct answer, we would have to find every synonym of $S_2$ and check if any of them has the property $P_1$. This requires a *path query*. A path query involving $N$ synonyms may require as many as $N-1$ edge traversals. Path queries are NP-hard [Mendelzon and Wood, 1995] and are, therefore, computationally expensive.

For example, rājikā, kṣava, kṣutābhijanaka, kṛṣṇīkā, kṛṣṇasarṣapa, rājī, kṣujjanikā, āsurī, tīkṣṇagandhā, cīnāka are all names of the same substance. A relation is added as follows,

uṣṇa $\vdash$ is Property of $\rightarrow$ rājikā

Now, suppose that we want to query for a property of the substance kṣava, which while referring to the same entity as rājikā, does not have a property edge incident upon it.

We, therefore, are forced to use a path query, and the query has to explore all the synonym paths from kṣava to find out if kṣava itself or one of its synonyms has any property edge. The number of such paths can be impractically large, especially for large knowledge graphs.

We perform a simple optimization heuristic to tackle this issue. We first identify a synonym $S_K$ among all the synonyms having the highest degree, i.e., $K \in \{1, .., N\}$, such that $K = \underset{i}{\text{argmax}}\ degree(S_i)$. We treat this as the *canonical name* for that synonym group, and we add a relation is Synonym of from every $S_i, i \neq K$ to $S_K$. Further, we transfer all the edges (other than the is Synonym of edge) from every $S_i, i \neq K$ to $S_K$. In other words, if $S_i$ was connected to a node $V$ by a relation $R$, after optimization, $S_K$ will be connected to node $V$ by relation $R$. Now, every synonym has a direct edge to the canonical name, with all the properties getting attached to the canonical name only. Thus, a query on any synonym has to traverse *at most 1 edge* before reaching the desired node.

At the end of curation and optimization steps, there were $410$ nodes and $764$ relationships that constitute our knowledge graph.

### 3.2.6 Querying

Although the ideal way of question answering is by posing queries in natural language, unfortunately, the state-of-the-art in Sanskrit NLP tools does not allow that. Hence, to simulate natural language queries, we use query templates.

The annotation and querying platform that we use, *Sangrahaka*, uses *Neo4j* graph database[7] for the purpose of storing and querying the knowledge graph. *Cypher*[8] is *Neo4j*'s graph query language inspired by SQL, but optimized for graph querying, and it makes use of intuitive ASCII-art syntax for querying. The platform utilizes the power of *Cypher* for connecting to the graph database. Natural language queries are simulated using query templates.

#### 3.2.6.1 Query Templates

A query template consists of a set of natural language templates and an equivalent graph query template. Each of these templates contain *placeholders*. Values of these placeholders can be filled by choosing the required entity, entity type or relation, to convert the query template into a valid natural language query. The same replacement in the graph query template yields a valid graph query which can be directly used to fetch results from the graph database.

For example, consider a sample query template:

- Sanskrit: `के पदार्थाः {0} इति दोषस्य वर्धनं कुर्वन्ति।`
- English: `Which entities increase the dosha {0}?`

[7] https://neo4j.com/
[8] https://neo4j.com/developer/cypher/

- Cypher:

```
MATCH (dosha:TRIDOSHA)-[relation:IS_INCREASED_BY]->(entity)
WHERE dosha.lemma = "{0}"
RETURN entity
```

The variable `{0}` here is a word representing an entity of type `TRIDOSHA`. The valid values for the variable in this query are vāta or pitta or kapha or one of their synonyms. So, natural language questions such as "Which substances increase kapha?", etc. can be realized using this query template.

In order to increase the number of the questions that can be answered, we have created a set of *generic* queries which help model *any* query up to a single relation. It contains the following three query templates:

- Which entity is related to entity `{0}` by relation `{1}`?
- How is entity `{0}` related to entity `{1}`?
- Show all matches where an entity of type `{0}` has relation `{1}` with an entity of type `{2}`.

We have a total of 31 *natural language query templates* in Sanskrit[9] to represent the most relevant queries. We have classified these templates semantically into 12 categories. Classification helps to locate an intended query template faster. An exhaustive list of these query templates and their categories is in Table 3.9.

#### 3.2.6.2 Query Answers

Result of graph queries are also graphs. The querying interface from *Sangrahaka* consisted of a graphical and a tabular display. Figure 3.12 shows a sample output using the query interface. In the graph, hovering over a node lets the user see the properties associated with that node. *Lemma* (word-stem) associated with that node is visible as the label of the node in the graph. In addition, provenance of the node such as which line from the corpus does that node correspond to, and the identifier of the annotator(s) who added that entity are also mentioned. The nodes are color-coded in such a way that nodes referring to entities of same type get the same color[10].

[9]We also have their English translated versions in the system.

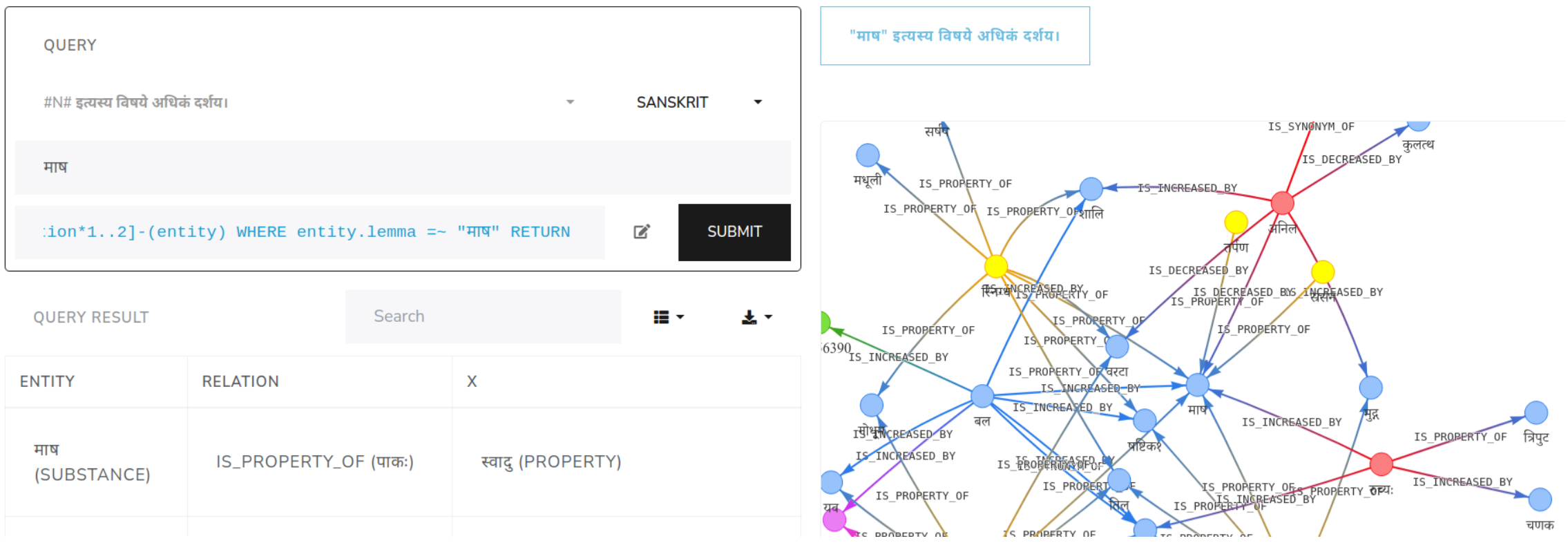

**Figure 3.12:** Sample output using query interface featuring Sanskrit query templates

## 3.3 Summary

Current state of Sanskrit NLP makes manual annotation a necessity for semantic tasks. In this chapter, we proposed the construction of a knowledge graph (KG) through manual annotation process with a special focus on capturing semantic information. We described a web-based tool *Sangrahaka* for annotation and querying of knowledge graphs. As a proof-of-concept, we selected three chapters from the nighaṇṭu text Bhāvaprakāśanighaṇṭu, carefully created an ontology, and performed semantic annotation to construct a knowledge graph. Our methodology is extensible to other nighaṇṭu texts.

The source code of *Sangrahaka* is freely available in the open-source domain at `https://github.com/hrishikeshrt/sangrahaka/`, allowing users to access and utilize it. The customized instance of the tool, deployed for the annotation and querying of Bhāvaprakāśanighaṇṭu, can be accessed at `https://sanskrit.iitk.ac.in/ayurveda/`.

[10]Colors are not fixed. Thus, the color *yellow* is not indicative of a specific entity type. It only ensures that for a particular query answer, all yellow nodes will have the same entity type.

**Table 3.9:** Natural Language Query Templates

| Category | Sanskrit Template | English Template | Input Type |
|---|---|---|---|
| सूचि (Contents) | पदार्थानां के प्रकाराः। | What are all the entity types? | |
| सूचि (Contents) | पदार्थेषु के सम्बन्धाः। | What are all the relationships? | |
| सूचि (Contents) | के के द्रव्याः। | What are all the substances? | |
| सूचि (Contents) | के के पदार्थाः। | What are all the entities? | |
| वर्णन (Detail) | `{0}` इत्यस्य विषये दर्शय। | Show some details about `{0}`. | Entity |
| वर्णन (Detail) | `{0}` इत्यस्य विषये अधिकं दर्शय। | Show some more details about `{0}`. | Entity |
| प्रकार (Type) | `{0}` इत्यस्य प्रकारः कः। | What is the type of `{0}`? | Entity |
| प्रकार (Type) | सर्वे `{0}` इति प्रकारस्य पदार्थाः चिनु। | Find all the entities of type `{0}`. | Entity-Type |
| गुण (Property) | केषां द्रव्याणां `{0}` इति गुणः अस्ति। | Which substances have a property `{0}`? | Entity |
| द्रव्य (Substance) | `{0}` इत्यस्य गुणाः के। | What are the properties of `{0}`? | Entity |
| द्रव्य (Substance) | `{0}` इति द्रव्यस्य प्रकाराः के। | What are the types/variants of the substance `{0}`? | Entity |
| समानार्थक (Synonym) | `{0}` इत्यस्य अन्यानि नामानि कानि। | What are the synonyms of `{0}`? | Entity |

**Table 3.9:** Natural Language Query Templates

| Category | Sanskrit Template | English Template | Input Type |
|---|---|---|---|
| सम्बन्ध (Relation) | {0} इति सम्बन्धेन बद्धं सर्वं दर्शय। | Find all entities related by the relation {0}. | Relation |
| सम्बन्ध (Relation) | {0} {1} एतयोः मध्ये कः सम्बन्धः। | What is the relation between {0} and {1}? | Entity, Entity |
| त्रिदोष (Tridoṣa) | के पदार्थाः {0} इति दोषस्य वर्धनं कुर्वन्ति। | Which entities increase the dosha {0}? | Entity |
| त्रिदोष (Tridoṣa) | के पदार्थाः {0} इति दोषस्य ह्रासं कुर्वन्ति। | Which entities decrease the dosha {0}? | Entity |
| त्रिदोष (Tridoṣa) | के पदार्थाः {0} इति दोषस्य वर्धनं {1} इति दोषस्य ह्रासं च कुर्वन्ति। | Which entities increase the dosha {0} and decrease the dosha {1}? | Entity, Entity |
| त्रिदोष (Tridoṣa) | के पदार्थाः {0} {1} एतयोः दोषयोः वर्धनं {2} इति दोषस्य ह्रासं च कुर्वन्ति। | Which entities increase the doshas {0} and {1} and decrease the dosha {2}? | Entity, Entity, Entity |
| त्रिदोष (Tridoṣa) | के पदार्थाः {0} इति दोषस्य वर्धनं {1} {2} एतयोः दोषयोः ह्रासं च कुर्वन्ति। | Which entities increase the dosha {0} and decrease the doshas {1} and {2}? | Entity, Entity, Entity |
| रोग (Disease) | के पदार्थाः {0} इति रोगं कुर्वन्ति। | Which entity causes the disease {0}? | Entity |
| रोग (Disease) | के पदार्थाः {0} इति रोगं हरन्ति। | Which entity cures the disease {0}? | Entity |

**Table 3.9:** Natural Language Query Templates

| Category | Sanskrit Template | English Template | Input Type |
|---|---|---|---|
| प्रभाव (Effect) | के पदार्थाः {0} एतं विकुर्वन्ति। | Which entities affect {0}? | Entity |
| प्रभाव (Effect) | के पदार्थाः {0} एतस्मै लाभप्रदाः। | Which entities benefit {0}? | Entity |
| प्रभाव (Effect) | के पदार्थाः {0} एतस्मै क्षतिप्रदाः। | Which entities harm {0}? | Entity |
| प्रभाव (Effect) | के पदार्थाः {0} इत्यस्य वर्धनं कुर्वन्ति। | Which entities increase {0}? | Entity |
| प्रभाव (Effect) | के पदार्थाः {0} इत्यस्य ह्रासं कुर्वन्ति। | Which entities decrease {0}? | Entity |
| अधिकरण (Space-Time) | {0} इति पदार्थः कदा जायते। | When does {0} grow? | Entity |
| अधिकरण (Space-Time) | {0} इति पदार्थः कुत्र लभ्यते। | Where is {0} found? | Entity |
| साधारण (Generic) | के पदार्थाः {0} इति पदार्थेन सह {1} इति सम्बन्धेन सम्बन्धिताः। | Which entity is related to {0} by relation {1}? | Entity, Relation |
| साधारण (Generic) | {0} इति पदार्थः {1} इति पदार्थेन सह कथं सम्बन्धितः। | How is {0} related to {1}? | Entity, Relation |
| साधारण (Generic) | {0} इति प्रकारस्य पदार्थैः सह {1} इति सम्बन्धेन बद्धाः {2} इति प्रकारस्य पदार्थान् दर्शय। | Show all matches where an entity of type {0} has relation {1} with an entity of type {2}. | Entity-Type, Relation, Entity-Type |

# Chapter 4

# *Antarlekhaka*: Comprehensive Natural Language Annotation Tool

*Sangrahaka* excels in the annotation for the construction of knowledge graphs. However, there is often a need for general-purpose annotation for various linguistic tasks. Additionally, to support the progress of natural language processing (NLP) and facilitate research, datasets for various NLP tasks are required.

We now introduce *Antarlekhaka*[1], a tool for distributed annotation that offers user-friendly interfaces to facilitate the annotation process of various common NLP tasks in a straightforward and efficient way. We propose a sequential annotation model, where an annotator carries out multiple annotation tasks relevant for a small text unit, such as a verse, before proceeding to the next. The tool has full Unicode support and is designed to be language-agnostic, meaning it can be used with corpora from many different languages, making it highly versatile.

The tool sports eight task-specific user-friendly annotation interfaces corresponding to eight general categories of NLP tasks: sentence boundary detection, canonical word ordering, free-form text annotation of tokens, token classification, token graph construction, sentence classification and sentence graph construction.

The goal of the tool is to streamline the annotation process, making it easier and more efficient for annotators to complete multiple NLP tasks on the same corpus.

[1]*Antarlekhaka* is a Sanskrit word meaning 'annotator'.

**Table 4.1:** Comparison of *Antarlekhaka* with various annotation tools based on primary features and supported tasks

| Feature | WebAnno | GATE | BRAT | FLAT | doccano | Sangrahaka | *Antarlekhaka* |
|---|---|---|---|---|---|---|---|
| Distributed Annotation | ✓ | ✓ | ✓ | ✓ | ✓ | ✓ | ✓ |
| Easy Installation | | | ✓ | ✓ | ✓ | ✓ | ✓ |
| Sequential Annotation | | | | | | | ✓ |
| Querying Interface | | | | | | ✓ | |
| Token Text Annotation | ✓ | ✓ | ✓ | ✓ | | | ✓ |
| Token Classification | ✓ | ✓ | ✓ | ✓ | ✓ | | ✓ |
| Token Graph | ✓ | ✓ | ✓ | ✓ | | ✓ | ✓ |
| Token Connection | ✓ | ✓ | ✓ | ✓ | | ✓ | ✓ |
| Sentence Boundary | | | | | | | ✓ |
| Word Order | | | | | | | ✓ |
| Sentence Classification | | | | | | | ✓ |
| Sentence Graph | | | | | | | ✓ |

## 4.1 *Antarlekhaka* Software

There are several text annotation tools available that target specific annotation tasks. However, each of these tools falls short in fulfilling all the requirements of an ideal annotation tool. For example, WebAnno [Yimam et al., 2013] is rich in features but becomes complex to use and experiences performance issues as the number of lines displayed on the screen increases. GATE Teamware [Bontcheva et al., 2013] is difficult to install. FLAT [van Gompel, 2014] lacks an intuitive interface and uses a non-standard data format. BRAT [Stenetorp et al., 2012] has not been actively[2] developed. doccano [Nakayama et al., 2018], although simple to set up and intuitive, only supports labeling tasks. None of these tools address the important tasks of sentence boundary detection or canonical word ordering. Thus, there is a need for an annotation tool that is user-friendly, easy to install and deploy, and covers all the necessary tasks for NLP annotation.

*Sangrahaka* [Terdalkar and Bhattacharya, 2021a], while being easy to setup and use, focuses only on the annotation towards creation of knowledge graphs. It also sports a querying interface for template-based natural language queries. But it lacks support towards general-purpose NLP annotation tasks.

[2]The latest version was published in 2012

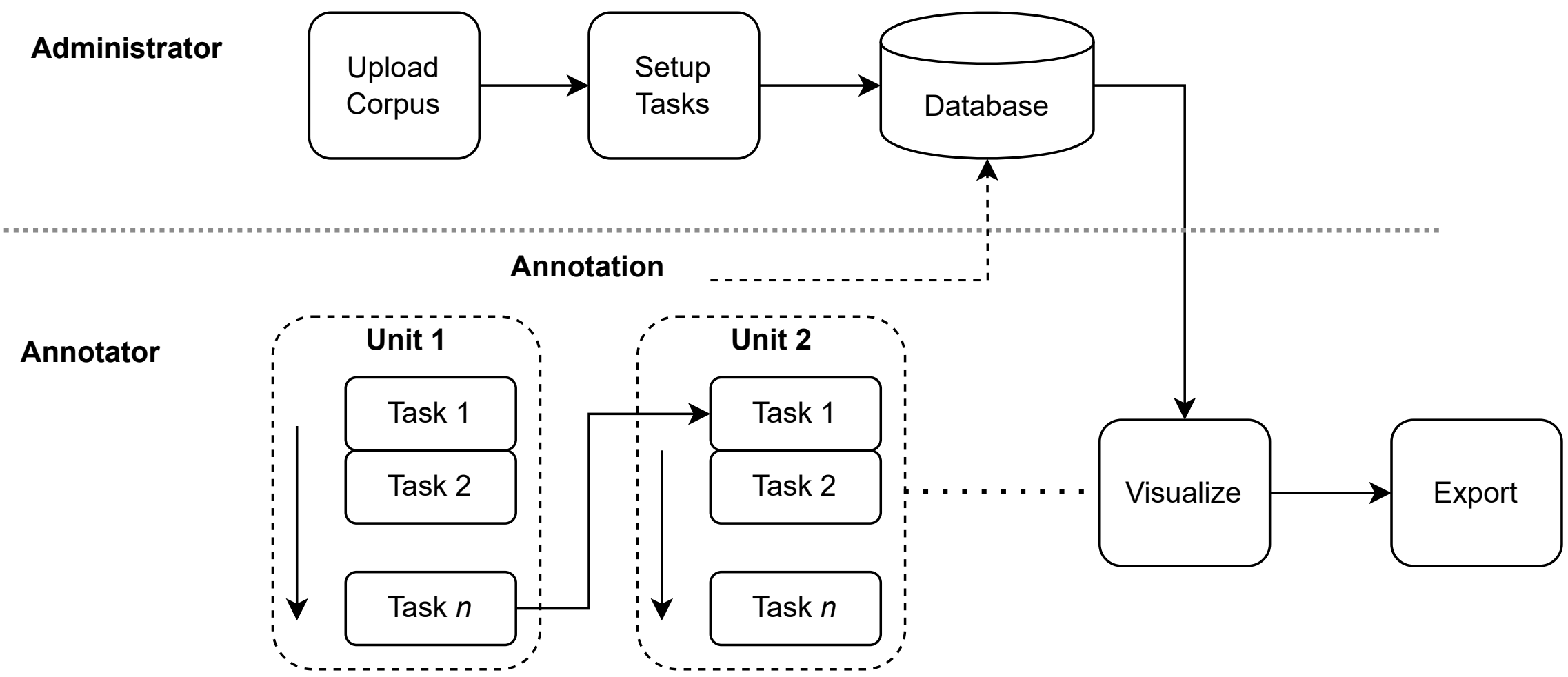

**Figure 4.1:** Workflow of Administrator and Annotator roles and their interaction with each other. Corpus upload, task setup, annotation and visualization are the principal components.

Thus, for the general purpose multi-task annotation of NLP tasks, we present *Antarlekhaka*. The annotation is performed in a sequential manner on small units of text (e.g., verses in poetry). The application is language and corpus agnostic. The tool is able to process data in two different formats: the standard CoNLL-U[3] format and plain text format. Regular-expressions based tokenizer is applied when using the data in plain text format.

Table 4.1 shows a comparison of the prominent annotation tools. We also conduct an objective evaluation of *Antarlekhaka* using the scoring methodology proposed by [Neves and Ševa, 2021]. We modify the criteria suitable to the domain of NLP annotation. Details of the evaluation are described in Section 4.1.2. It is important to note that while some of the existing tools, in theory, have the capability to support certain NLP tasks, they may not be designed with user-friendly interfaces.

### 4.1.1 Architecture

*Antarlekhaka* is a language-agnostic, multi-task, distributed annotation tool that is presented as a Web-deployable software. The tool leverages various technologies for its implementation, such as *Python 3.8* [Van Rossum and Drake, 2009], *Flask 2.0.1* [Ronacher, 2011, Grinberg, 2018], and *SQLite 3.38.3* [Hipp, 2022] for the backend,

[3]https://universaldependencies.org/format.html

and *HTML5, JavaScript*, and *Bootstrap 4.6* [boo, 2021] for the frontend.

*Flask* web framework powers the backend of *Antarlekhaka* providing a robust and scalable infrastructure. A web framework is responsible for a range of backend tasks, including routing, templating, managing user sessions, connecting to databases and others. The recommended way to run the tool in a production environment is using a *Web Server Gateway Interface* (WSGI) HTTP server, such as *Gunicorn* [gun, 2021], which can operate behind a reverse proxy server such as *NGINX* [ngi, 2021] or *Apache HTTP Server* [apa, 2023]. However, any WSGI server, including the built-in server of Flask, can be utilized to run the application.

*SQLite* is used as the database management system to store and manage the data and metadata associated with the annotation tasks. An Object Relational Mapper (ORM) *SQLAlchemy* [sql, 2021] is used to interact with the relational database. This allows the user to choose any supported dialect of traditional SQL, such as *SQLite*, *MySQL* [mys, 2023], *PostgreSQL* [pos, 2023], *Oracle* [ora, 2023], *MS-SQL* [mss, 2023], *Firebird* [fir, 2023], *Sybase* [syb, 2023] and others[4].

The frontend of the tool, built using *HTML5, JavaScript*, and *Bootstrap*, provides user-friendly interfaces for annotators and administrators. The tool provides a feature-rich administrative interface to manage user access, corpus, tasks and ontology. The tool also includes eight types of intuitive annotation interfaces. These are further explained in detail in Section 4.1.1.3.

By combining these technologies, *Antarlekhaka* offers a powerful and flexible solution for large-scale annotation projects.

#### 4.1.1.1 Workflow

The workflow of the system is demonstrated in Figure 4.1.

The application is presented as a full-stack web-based software. It follows a single-file configuration system. An administrator may configure the tool and deploy it to web, making it immediately available for use. User registration is supported.

[4] `https://docs.sqlalchemy.org/en/20/dialects/`

User access is controlled by a 4-tier permission system, namely User, Annotator, Curator and Admin.

The tool has eight annotation interface templates corresponding to eight generic categories of NLP annotation tasks: sentence boundary detection, canonical word order, free-form token annotation, token classification, token graph construction, token connection, sentence classification, and sentence graph construction. Various NLP tasks can be modelled using each of these categories. More than one tasks of same category may be required for a specific annotation project. For example, named entity recognition (NER) and parts-of-speech (POS) tagging are both examples of token classification. To facilitate this, the administrative interface allows an administrator to create multiple tasks of each category. Additionally, an administrator can also control the set of active tasks, order of the tasks, ontology for the relevant tasks, corpus management and user access management.

We propose a streamlined sequential mode of annotation where an annotator completes multiple annotation tasks for a single unit of text before moving on to the next unit. While sequential annotation is suggested, it is not strictly enforced, allowing an annotator to perform only a subset of tasks, as well as go back and make changes. The set and order of tasks is customizable through an administrative interface. We consider a small logical block of text as a unit for the annotation, e.g., a verse from the poetry corpus.

#### 4.1.1.2 Data

The data for corpus can be in either of two formats: *CoNLL-U* format or *plain text* format and can contain Unicode text. *CoNLL-U* is a widely used format for linguistic annotation, and it is based on the column format for treebank data. The format is designed to store a variety of linguistic annotations, including part-of-speech tags, lemmas, morphological features, and dependencies between words in a sentence. Figure 4.2 shows a sample of *CoNLL-U* data taken from Digital Corpus of Sanskrit [Hellwig, 2021].

```
# text = tapaḥsvādhyāyaniratam̩ tapasvī vāgvidām̩ varam
# sent_id = 110529
# sent_counter = 1
# sent_subcounter = 1
1-3	tapaḥsvādhyāyaniratam̩	_	_	_	_	_	_	_	_
1	tapas	tapas	NOUN	_	Case=Cpd	_	_	_	LemmaId=96401|OccId=31
2	svādhyāya	svādhyāya	NOUN	_	Case=Cpd	_	_	_	LemmaI
3	niratam	niram	VERB	_	Case=Acc|Gender=Masc|Number=Sing|VerbForm=Part	_	_
4	tapasvī	tapasvin	NOUN	_	Case=Nom|Gender=Masc|Number=Sing	_	_
5-6	vāgvidām̩	_	_	_	_	_	_	_	_
5	vāc	vāc	NOUN	_	Case=Cpd	_	_	_	LemmaId=76023|OccId=31
6	vidām	vid	ADJ	_	Case=Gen|Gender=Masc|Number=Plur	_	_	_
7	varam	vara	ADJ	_	Case=Acc|Gender=Masc|Number=Sing	_	_	_

```

**Figure 4.2:** Example of CoNLL-U Data from Digital Corpus of Sanskrit. The columns displaying word index, word form, lemma, universal parts-of-speech tag, language-specific parts-of-speech tag, and morphological features are visible.

Data in *CoNLL-U* format can be obtained directly from treebanks such as Universal Dependencies [De Marneffe et al., 2021], which is a project that aims to develop cross-linguistically consistent treebank annotation. In addition, NLP tools such as Stanza [Qi et al., 2020] are capable of processing a general corpus of text and producing data in *CoNLL-U* format, making it easier to obtain data in this format.

On the other hand, plain text data is processed using a regular-expression based tokenizer, which is a process that splits the text into individual units of meaning, such as verses, lines and tokens using patterns defined in the form of regular expressions to identify the respective separators. This process allows the system to convert plain text data into a more structured format. The default tokenizer uses '\n\n' (two newlines, i.e., one blank line) as verse delimiter, '\n' (single newline) as line delimiter and '\s' (whitespace) as token delimiter. The plain text processor module is a pluggable component. An administrator may reimplement it using any language specific features or tools as long as the data output by the module meets the current format specifications.

After the data has been imported, it is organized in a five-level hierarchy structure consisting of: Corpus, Chapter, Verse, Line, and Token. The structure is designed to provide a clear and systematic way to categorize and access the data, making it easier to locate and analyse specific pieces of information.

The first level of the structure, Corpus, refers to the entire collection of text data.

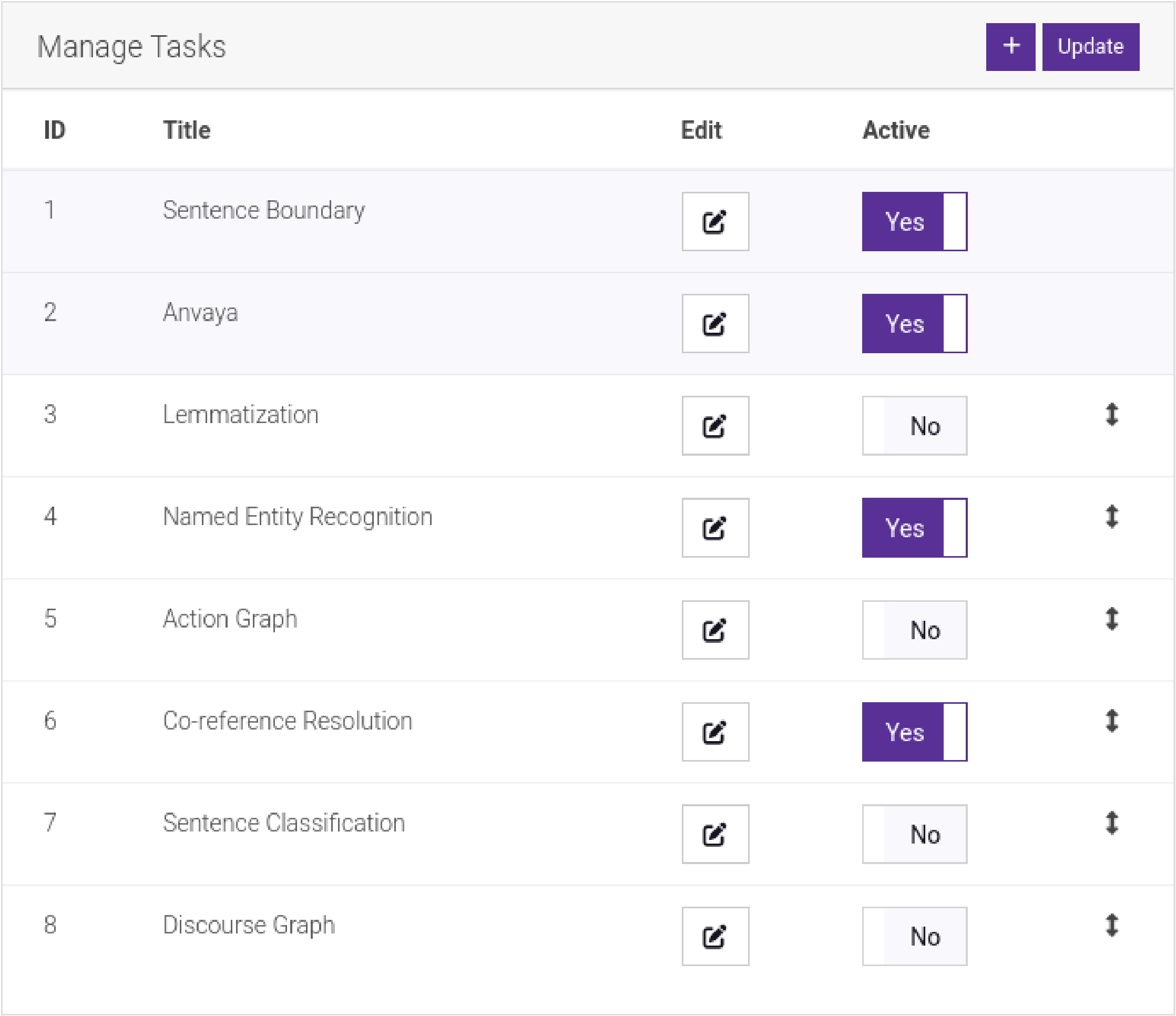

**Figure 4.3:** Task management interface, a part of administrative interface. Tasks can be added, edited, activated, deactivated and reordered here.

The second level, Chapter, refers to a specific division of the text, such as a book or section of a book. The third level, Verse, is the default unit of annotation, and it refers to a specific verse or passage within the chapter. The fourth level, Line, refers to the specific line of text within the verse, and the fifth and final level, Token, refers to the individual units of meaning within the line, such as words, numbers, or punctuation marks.

The hierarchical structure of the data provides a clear and organized framework for annotating and analyzing the data, making it easier to capture the relationships between different elements of the data.

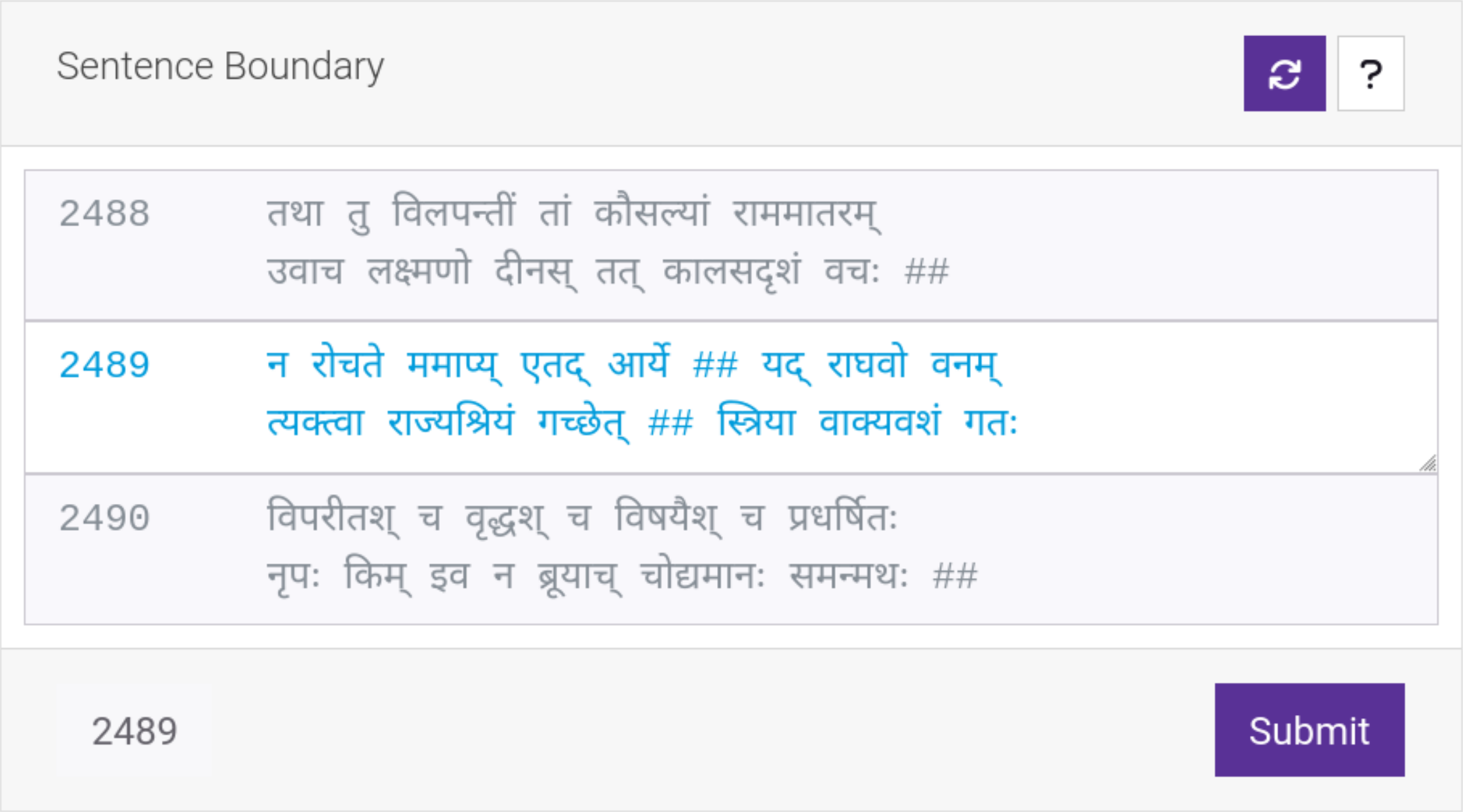

**Figure 4.4:** Sentence Boundary Annotation Interface

#### 4.1.1.3 Interfaces

The annotation tool provides eight intuitive annotation interfaces to support various NLP annotation tasks. These interfaces are, Sentence Boundary Detection, Canonical Word Order, Token Annotation, Token Classification, Token Graph, Token Connection, Sentence Classification, and Sentence Graph. There may be multiple tasks of each category. The information such as task title, task-specific instructions to annotators, set of active tasks and the task order can be configured by an administrator from the administrative interface as illustrated in Figure 4.3.

The tool proposes a sequential model of annotation. An annotator is shown the corpus in the form of small text units (e.g., verses) on the left side, and an annotation area on the right side of the screen. Upon submitting the annotations for a particular task, the interface automatically takes the annotator to the next task. An annotator is expected to complete all the tasks associated with a text unit before moving on to the next unit. This, however, is not strictly enforced, allowing annotator to still go back to make changes to the annotation.

We now describe each task category and the corresponding frontend interface.

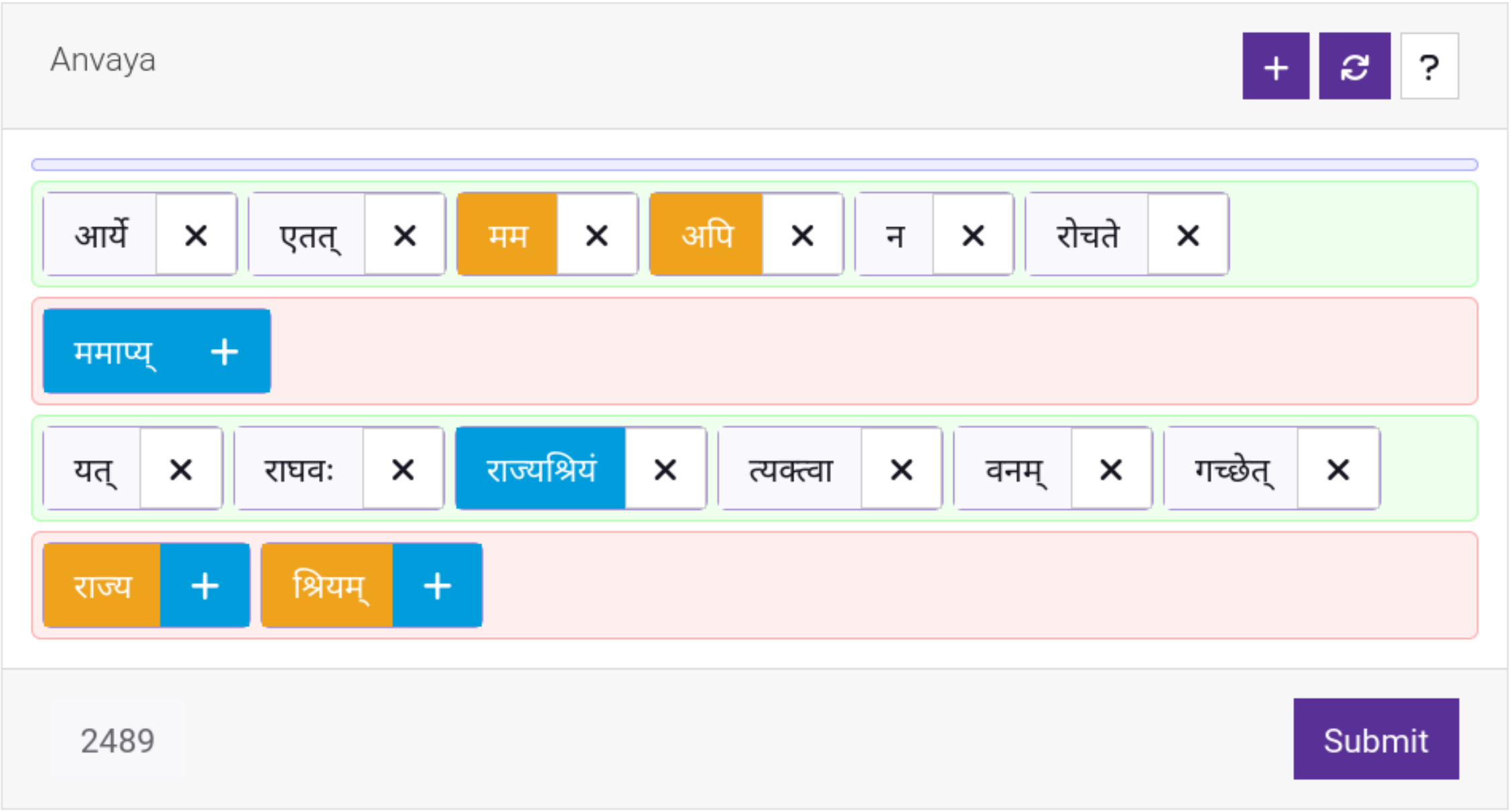

**Figure 4.5:** Word Order Annotation Interface

**4.1.1.3.1 Sentence Boundary Detection** The annotator is presented with a unit of text in an editable text area prefilled with the original text. The user is tasked with identifying the sentence boundaries and placing the delimiter '##' (two 'hash' symbols) at the end of every sentence, effectively marking the sentence boundaries. If the sentence does not end in the displayed unit, the user does not add any delimiters. Once the sentence boundaries have been marked, the user can move on to the next annotation task. An illustration of this annotation task is shown in Figure 4.4.

The importance of the sentence boundary task is not limited to languages without distinct sentence markers; it also pertains to poetry text, making it relevant to all languages.

It's worth mentioning that although the sentence boundary task is given primary citizen treatment, it can still be turned off for languages where it's not applicable. In such instances, the boundaries of annotation text units (e.g., verses) are treated as sentence boundaries.

**4.1.1.3.2 Canonical Word Order** All sentences that end in the current unit of text are displayed to the annotator as a list of sortable tokens. The annotator has the ability to rearrange these tokens into the correct canonical word order by dragging them into place. Additionally, if any tokens are missing, the annotator can add them as well. A visual representation of this task can be seen in Figure 4.5. The sorting capability is made possible through the use of the *jQuery UI (Sortable plugin)*[5].

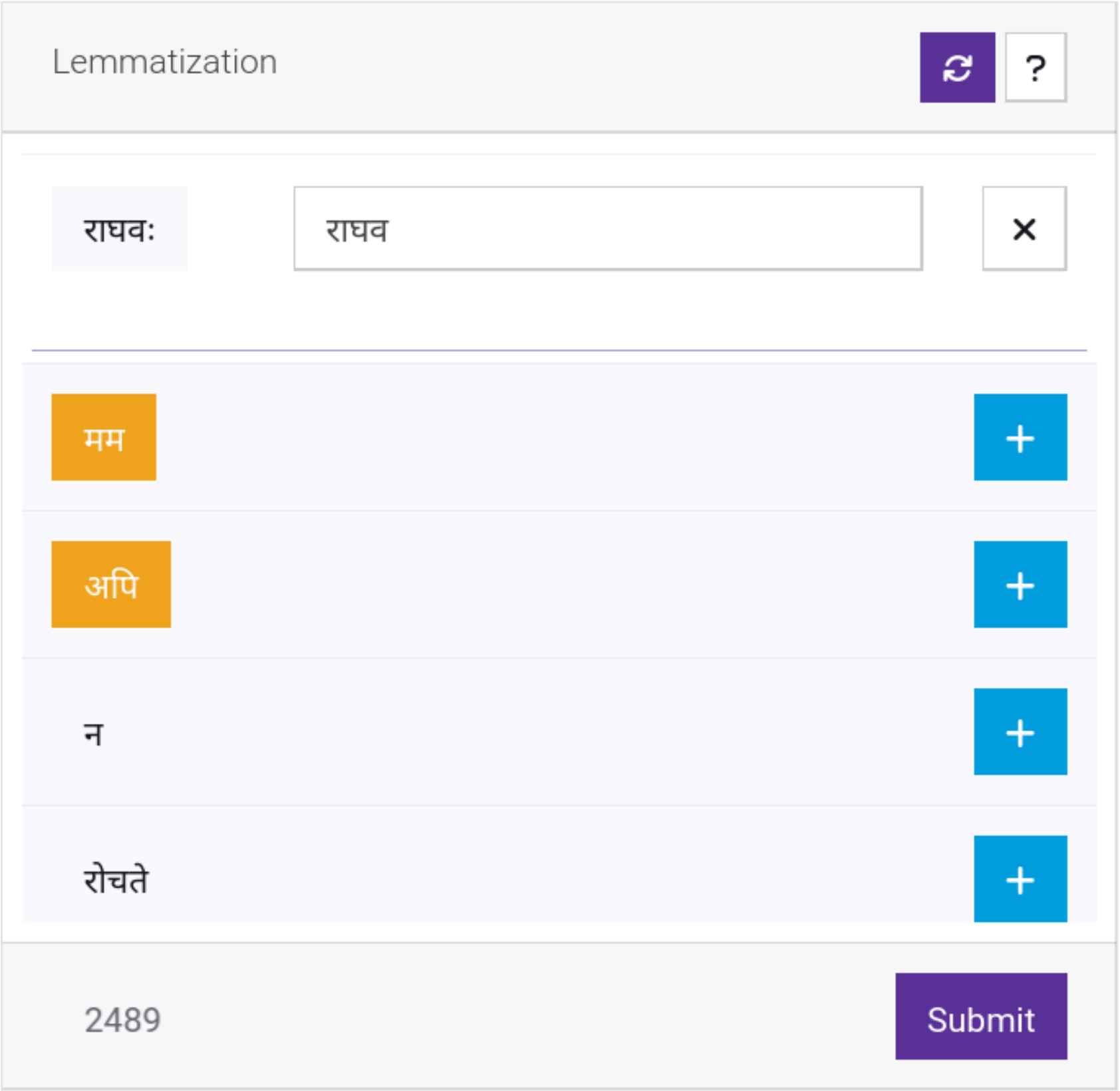

**Figure 4.6:** Token Annotation Interface: Lemmatization

**4.1.1.3.3 Token Annotation** The token annotation interface allows an annotator to add free-form text associated with every token. This free-form text can have different purposes, such as to identify the root word of a word (lemmatization), to separate multi-word expressions into individual words (compound splitting), to analyse the morphological structure of a word (morphological analysis), etc. The token annotation interface is shown in Figure 4.6.

**4.1.1.3.4 Token Classification** Token classification is a process of assigning predefined categories to individual tokens in text data. It is a special case of free-form token annotation, wherein the annotations are guided by an ontology. For such an annotation task, an administrator must create an ontology at the time of setting up the task. During the annotation process, an annotator is provided with a list of tokens, each accompanied by a dropdown menu, from which they can select the appropriate category for relevant tokens. Some common examples of token classification tasks include named entity recognition, dependency tagging, part-of-speech tagging, and compound classification. Figure 4.7 illustrates the token classification interface.

[5] `https://api.jqueryui.com/sortable/`

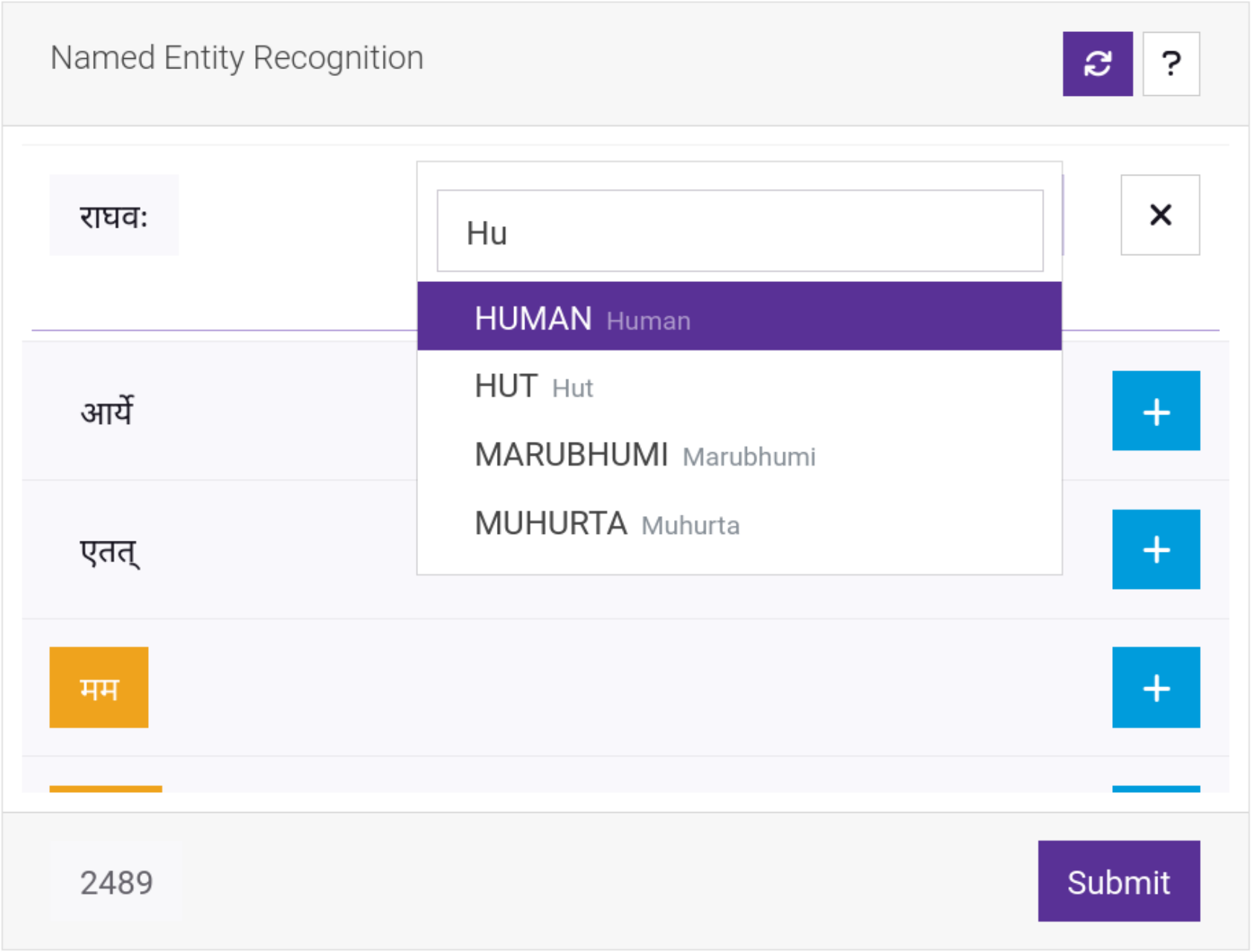

**Figure 4.7:** Token Classification Interface: Named Entity Recognition

**4.1.1.3.5 Token Graph** A token graph is a graph representation of the sentence, where the nodes are tokens belonging to a single sentence and the relations are based on an ontology. Tasks such as dependency parse tree, constituency graph, action graph are examples of tasks belonging to this category.

Semantic triple[6] is a standard format to represent and store graph-structured information in a relational database in a systematic manner. The interface allows an annotator to add multiple relations per sentence in the form of subject-predicate-object triples, where subject and object are tokens from the sentence and the predicate is a relation from the task specific ontology. The valid values of subject, object and predicate appear in individual dropdown menu elements for the annotator to choose from. Erroneous triples may also be removed. During the annotation process, an annotator can view the current status of the token graph at any time by clicking the 'Show Graph' button.

[6]https://en.wikipedia.org/wiki/Semantic_triple

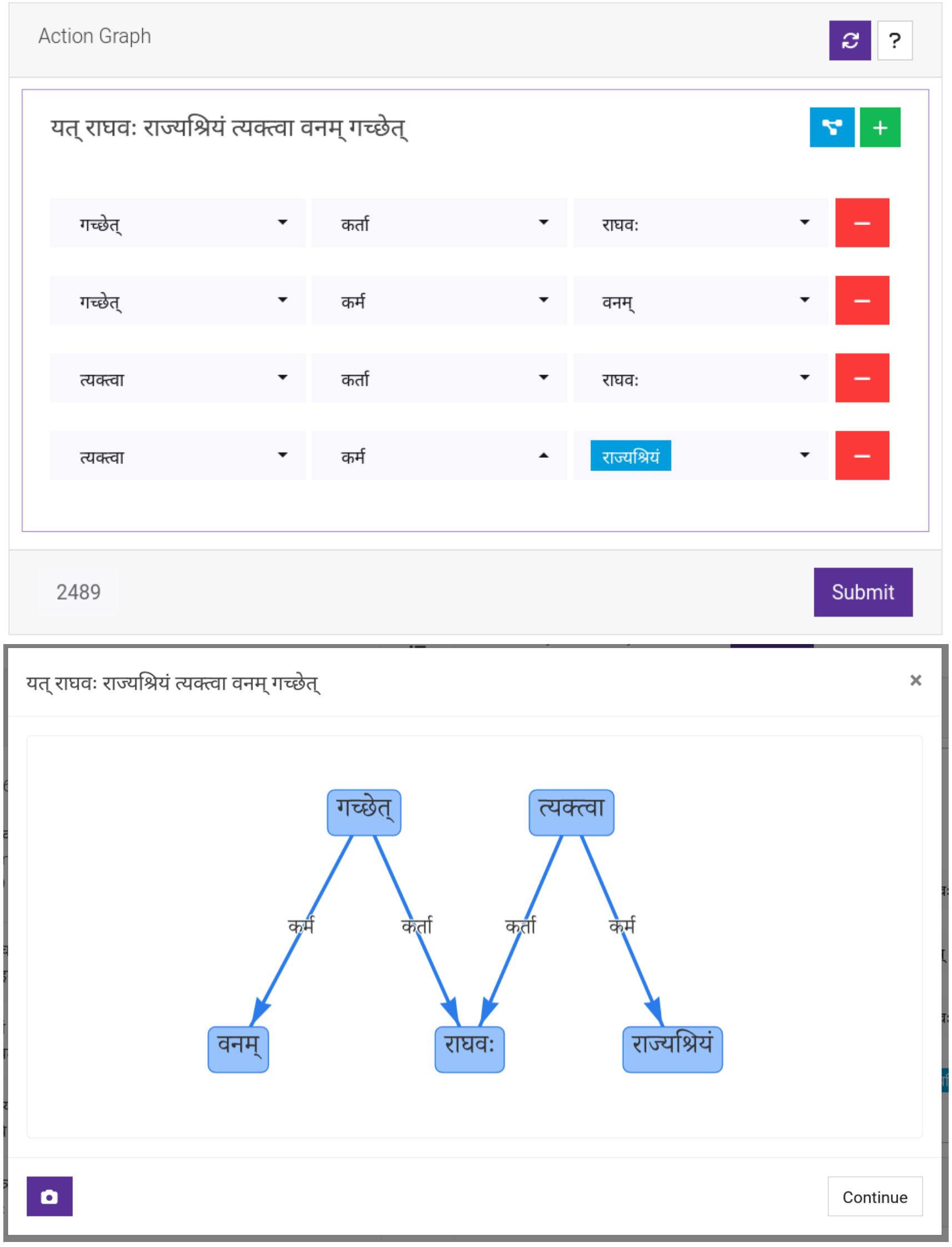

**Figure 4.8:** Token Graph Interface for a sample task of Action Graph. The second interface shows the visualization of the graph

Figure 4.8 shows the token graph interface with graph visualization.

**4.1.1.3.6 Token Connection** Token connection is similar to token graph, however, there is a single type of relation to be captured. For example, when marking co-references, only connecting the two tokens to each other is sufficient, while the relationship 'is-coreference-of' is implicit. The tool provides a special simplified interface for this scenario. In addition to implicit relations, token connections can also be established across sentences. The annotator is presented with a list of clickable tokens from the current sentence as well as tokens from a context window of previous $n = 5$ sentences. The annotator can add a connection by clicking on the source token and the target token one after the other and confirming the connection. If a connection is added in error, it can be removed as well. The token connection interface is shown in Figure 4.9.

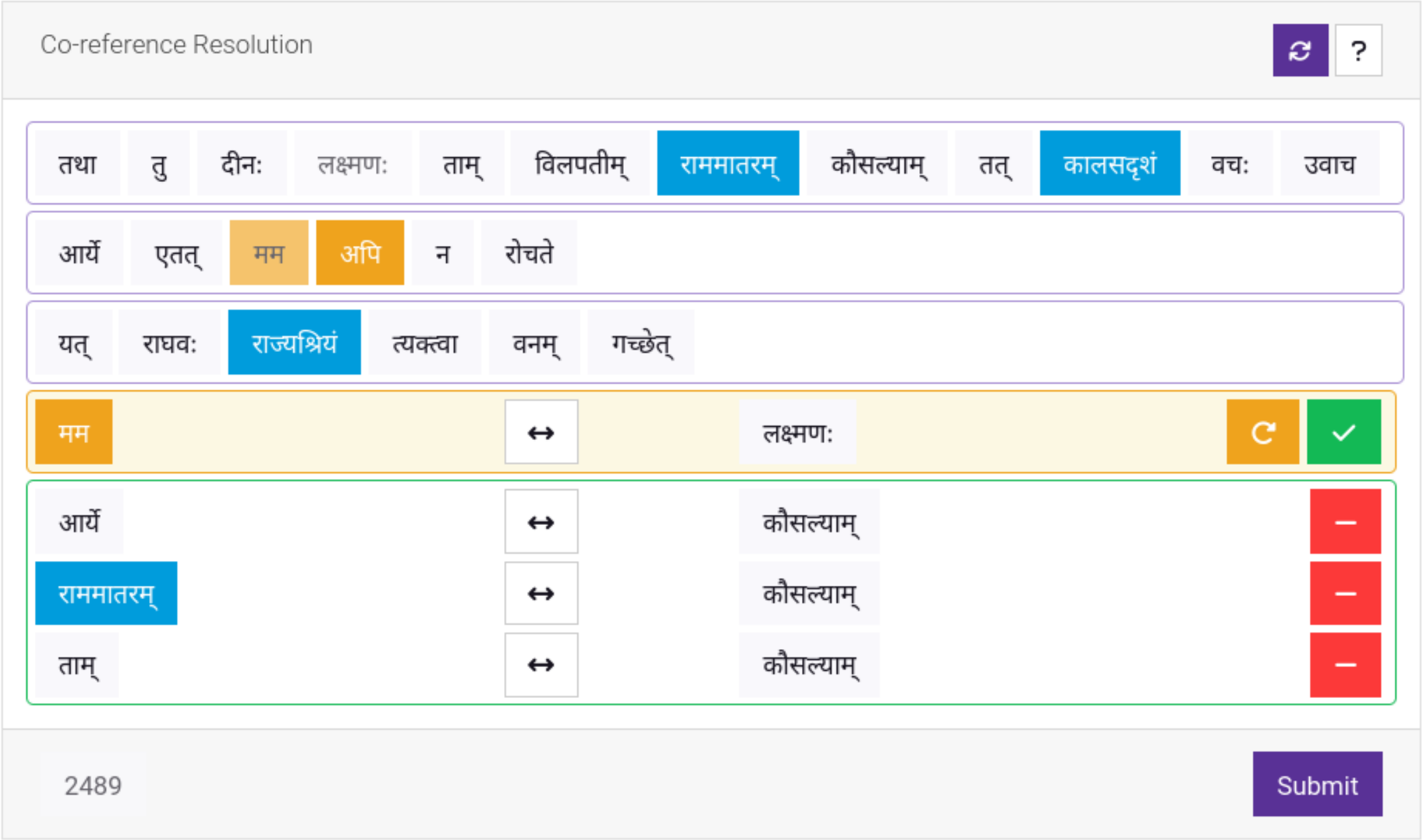

**Figure 4.9:** Token Connection Interface for a sample task of Co-reference Resolution

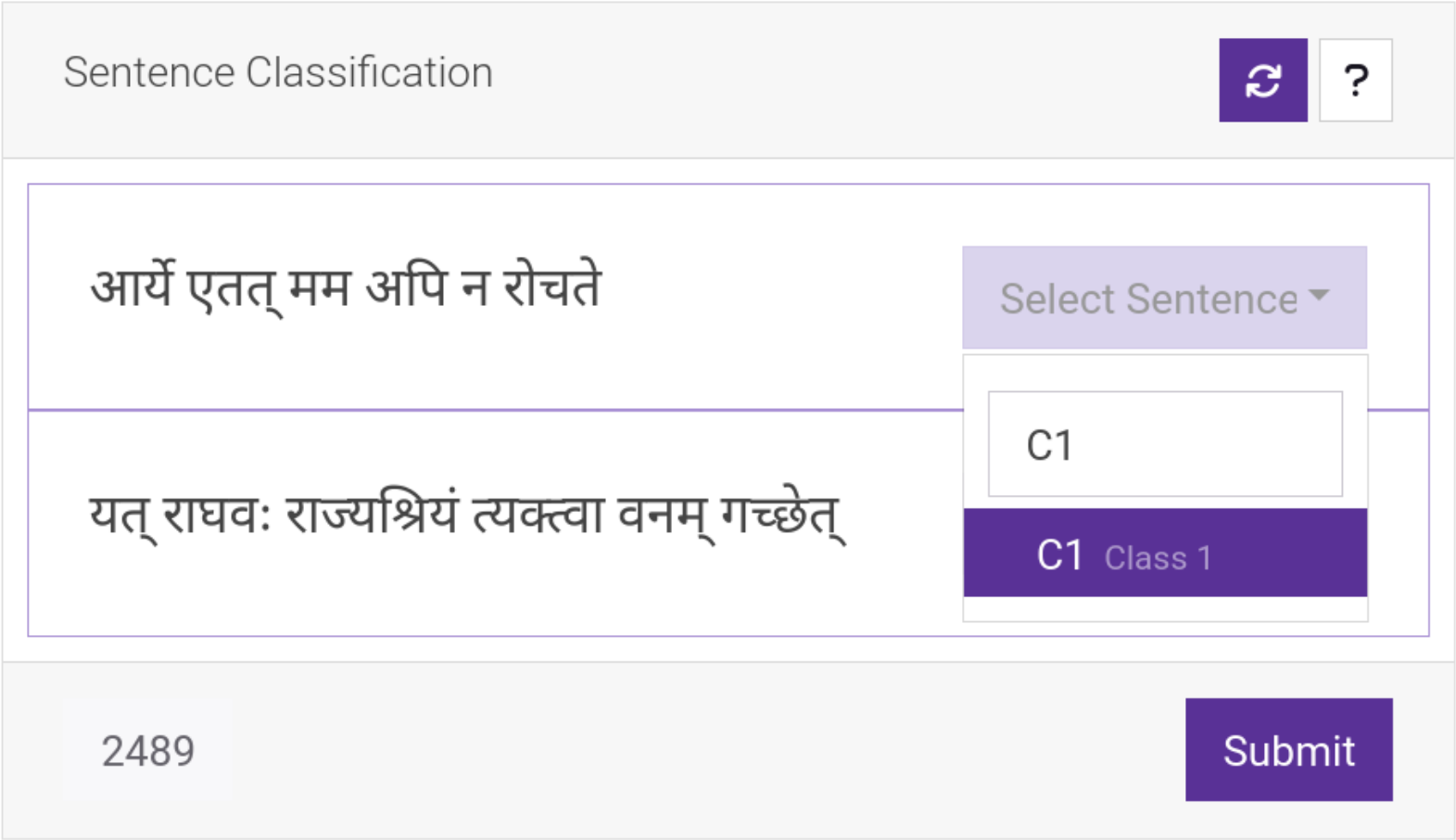

**Figure 4.10:** Sentence Classification Interface

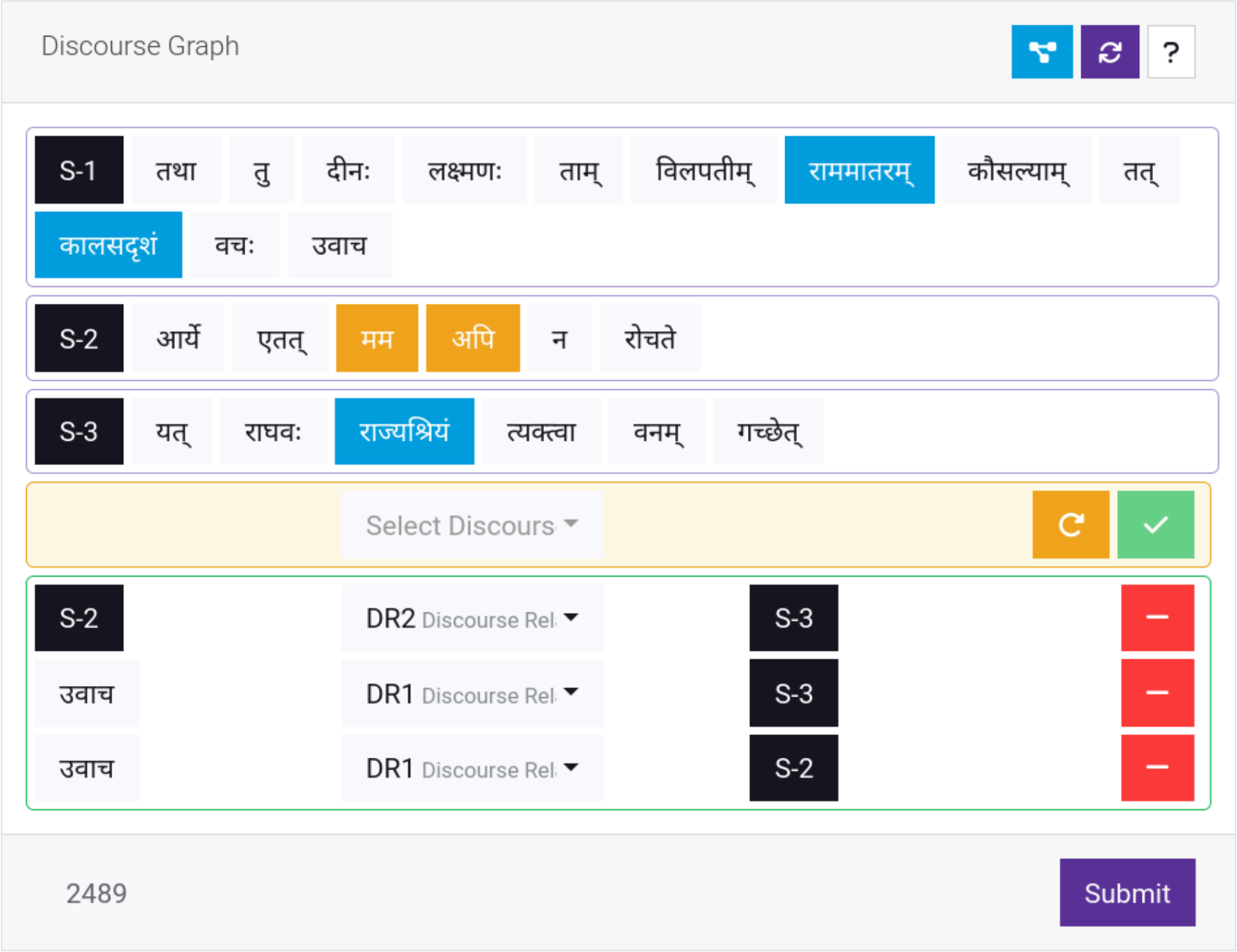

**Figure 4.11:** Sentence Graph Interface

**4.1.1.3.7 Sentence Classification** Sentence classification is a task where sentences are classified into different categories. This task is similar to ontology-driven token classification, with the difference being that classes are associated with sentences rather than tokens. The ontology is predefined by the administrator while setting up the task. The annotator can select the category for a sentence from a dropdown menu. Tasks such as sentiment classification and sarcasm detection are examples of sentence classification tasks. Figure 4.10 illustrates the sentence classification interface.

**4.1.1.3.8 Sentence Graph** Sentence graph is a graph representation of relationships between sentences. The connections can be between tokens or complete sentences, and the relationships are captured as subject-predicate-object triples. Tokens from the previous $n = 5$ sentences are presented as buttons arranged in the annotated word order. An annotator can create connections by clicking on the source and target tokens and selecting the relationship from a dropdown menu based on an ontology. A special token is provided to denote the entire sentence as an object. Tasks such as timeline annotation and discourse graphs are examples of tasks belonging to this category. Figure 4.11 shows the interface for creating sentence graph connections. Similar to the token graph task, an annotator can visualize the sentence graph as well.

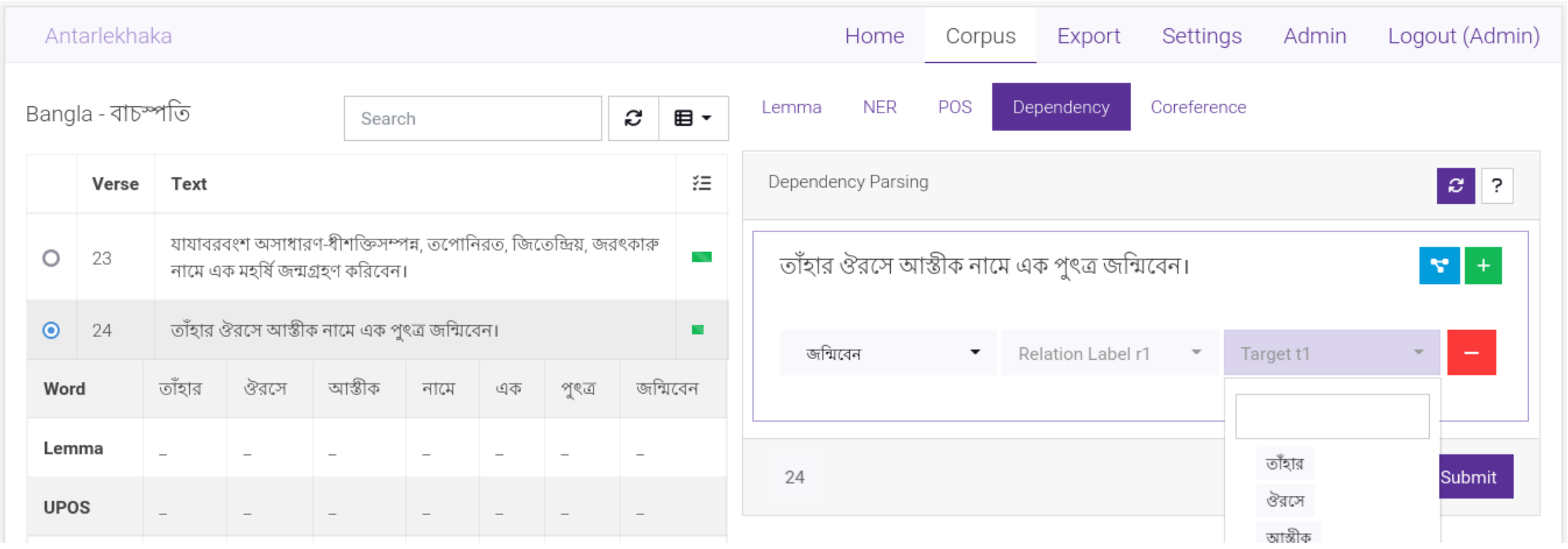

**Figure 4.12:** Annotation Interface: Corpus area shows text split into small units, and annotation area highlights various annotation task tabs

#### 4.1.1.4 Language Independence

Unicode is a widely used computer industry standard for encoding, representing, and handling text in a uniform and consistent way across various computing platforms and applications. The standard assigns unique numerical codes to each character in a large number of scripts, including those used for writing many of the world's languages. With full support for Unicode, it means that text data from any language that is encoded using Unicode can be uploaded and processed by the system. This allows users to work with text data in their preferred languages.

We have seen illustrations of individual interfaces in Section 4.1.1.3 for a Sanskrit corpus. Figure 4.12 displays the overall annotation interface for a corpus of Bengali language. This also highlights the language-agnostic nature of the tool.

#### 4.1.1.5 Schema

*Antarlekhaka* utilizes a relational database to store information such as, corpus data, user data, task data and annotations. A relational database allows for efficient storage and retrieval of functional data, as well as the ability to establish relationships between different pieces of data. For example, annotations of specific verses by spe-cific users can be linked allowing the system to quickly locate and display relevant annotations when needed.

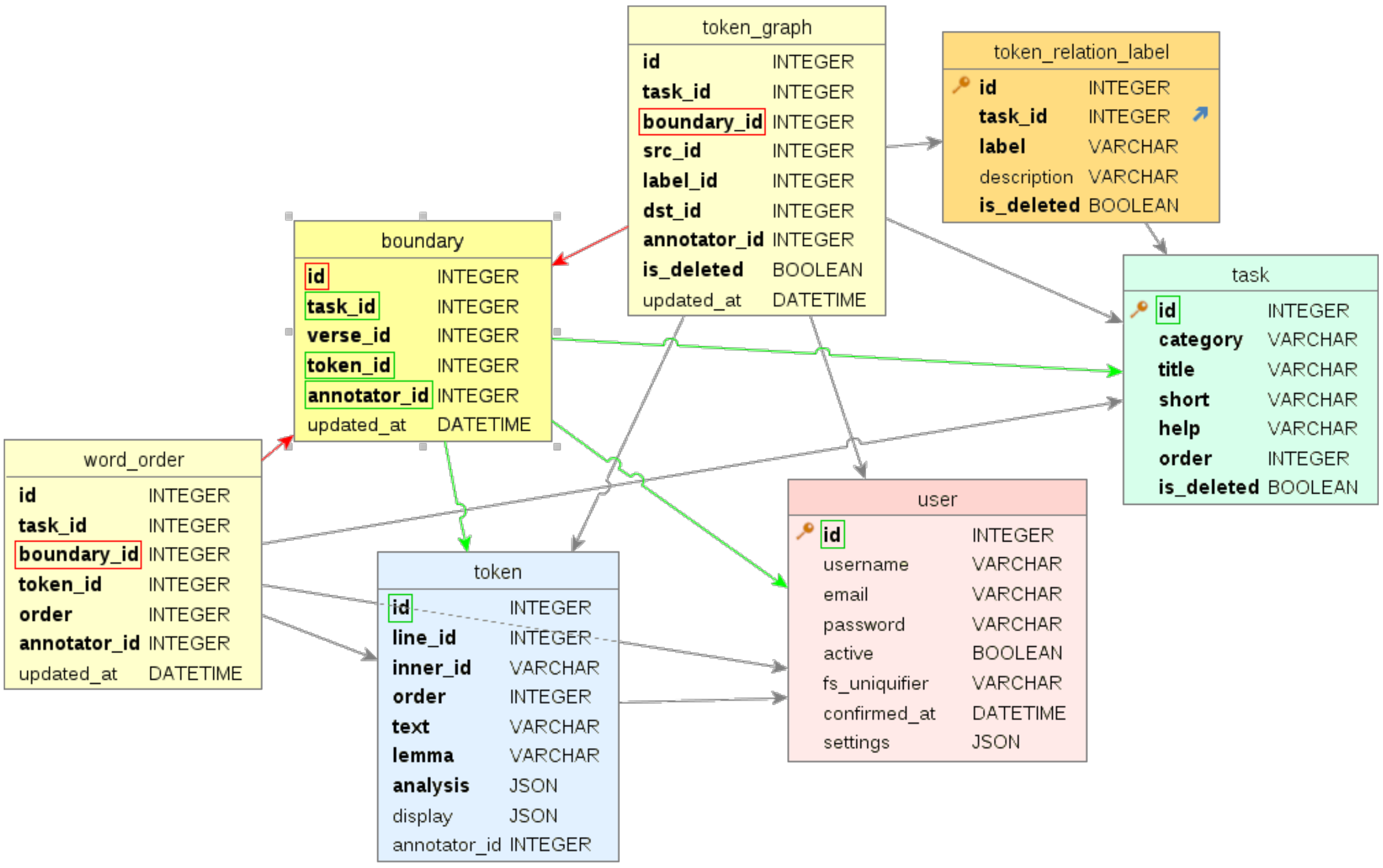

**Figure 4.13:** Entity Relationship Diagram showing selective tables: Task, User, Token, Sentence Boundary Annotation, Word Order Annotation, Token Graph Annotation, Token Graph Relation Ontology. Tables are color coded. Yellow: Annotation Tables, Orange: Ontology Tables, Blue: Corpus Tables, Pink: User Tables, Green: Task Information Table. The annotation table for 'Sentence Boundary' task is highlighted, showing the references incoming (red) and outgoing (green) references to other tables.

**4.1.1.5.1 Tasks** The information regarding tasks is stored in a single table within the relational database. This table serves as a centralized repository for information related to each task, including its title, category, and instructions for annotators. Each task is assigned a unique identifier known as a 'task id', which serves as a means of easily referring or linking to a specific task.

**4.1.1.5.2 Ontology** Ontology encompasses a collection of tags or categories that annotators can select and assign to tokens or sentences. It typically holds relevance for any type of classification tasks. Ontology is required for four task categories: token classification, token graph, sentence classification, and sentence graph. The ontology information is stored as a flat list of labels in four separate tables, each specific to a particular task category. There may be multiple tasks corresponding to each of these categories. Therefore, every ontology table also has a 'task id' column which associates the ontology entries with the corresponding tasks. This setup allows for clear organization and linking of the ontology information with the relevant tasks.

**4.1.1.5.3 Annotations** There are eight annotation tables, each corresponding to a different category of annotation tasks. Annotations of all tasks belonging to each category are stored in the corresponding table. The annotations are linked to the semantic units of text, specifically, the sentences marked in the sentence boundary task[7]. The other seven annotation tables include a reference to the 'boundary id'. In cases where the sentence boundary task is not necessary, the boundaries of the annotation text units (e.g., verse) are considered as sentence boundaries and annotated automatically in the background using a special annotation user. Additionally, to facilitate multiple instances of tasks from each task category, every annotation table contains a reference to the 'task id'. Finally, each annotation table sports a tailored schema to support the recording of task specific annotations. An 'annotator id' associated with every task annotation table, allows for proper organization and tracking of the annotations.

Figure 4.13 shows the Entity Relationship (ER) diagram on a subset of tables from the relational database of *Antarlekhaka*.

#### 4.1.1.6 Pluggable Heuristics

The tool supports the use of heuristics as 'pre-annotations' to assist annotators. Heuristics are custom functions that generate suggestions for the annotators to use or ignore. These heuristics are often specific to the language and corpus, and thus, must be implemented by the administrator when setting up the tool. The tool outlines the format and type specifications of the heuristics, making them a pluggable component.

[7]If an annotator initially marks an incorrect sentence boundary but later recognizes the mistake and corrects it, the earlier annotations associated with that particular 'boundary id' are subsequently removed from all related tasks. Annotations related to sentences that are not connected to that specific boundary remain unaltered.

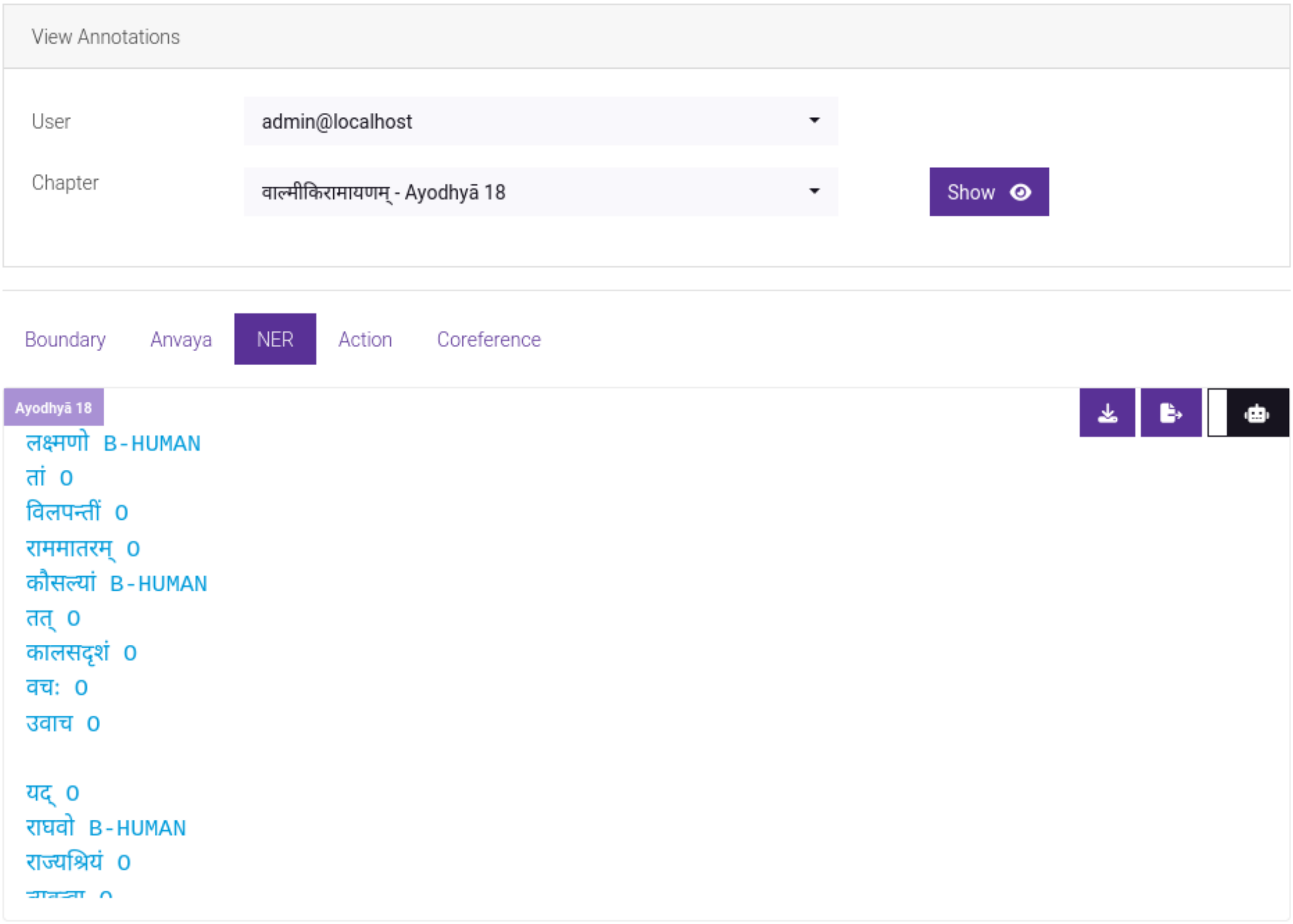

**Figure 4.14:** Export Interface: NER data in the standard BIO format

### 4.1.1.7 Export

The export interface enables the access, retrieval and visualization of the annotated data for each task in a clear and straightforward manner. Annotator can easily view and export the data in two formats (1) a human-readable format for easy inspection and (2) a machine-readable format compatible with the standard NLP tools. The specifics of the standard format depend on the task. For example, a standard format for NER datasets is the BIO format [Tjong Kim Sang and De Meulder, 2003], which stands for *begin*, *inside*, and *outside*. The B-tag marks the beginning of a named entity, while the I-tag indicates the continuation of a named entity. The O-tag signifies that a word is not part of a named entity. Figure 4.14 illustrates the export interface showcasing the capability to export NER data in the standard BIO format. The interface facilitates the visualization and export of graph representations for tasks related to graphs.

The export interface is accessible not only to annotators but also to curators, allowing them to review annotations made by other users. This feature serves as a mechanism for quality control.

### 4.1.2 Evaluation

The tool is being used for the voluminous task of annotation of a large corpus in Sanskrit, Vālmīki Rāmāyaṇa. The details of this project are described in Section 4.3. Additionally, the tool is also being used for the annotation of plain text corpus in Bengali language.

The time taken for annotation may not be a reliable measure of evaluation for annotation because annotators often spend more time processing the text to identify the relevant information than physically annotating. Further, the annotations may be spread out over multiple sessions of varying lengths over a prolonged period.

We have evaluated our tool using the two-fold evaluation method of subjective and objective evaluation, which was also by [Terdalkar and Bhattacharya, 2021a]. The subjective evaluation consisted of an online survey, wherein, $16$ annotators participated and rated the tool, out of 5, on various categories such as ease of use, annotation interface, and overall performance. The tool received mostly positive ratings, obtaining ratings of $4.3$ for ease of use, $4.4$ for annotation interface and an overall score of $4.1$.

The objective evaluation was done using the scoring mechanism used in previous works [Neves and Ševa, 2021, Terdalkar and Bhattacharya, 2021a]. The extra categories introduced by [Terdalkar and Bhattacharya, 2021a] were retained and additional relevant categories for the NLP tasks were added. As a result, a total of $29$ categories were used for evaluation. *Antarlekhaka* scored $0.79$, performing better than other tools such as *Sangrahaka* ($0.74$), *FLAT* ($0.71$) and *WebAnno* ($0.67$). Table 4.2 enlists $29$ categories used for scoring the annotation tools objectively.

**Table 4.2:** Evaluation of *Antarlekhaka* in comparison with other annotation tools using objective evaluation criteria. Each feature is evaluated on a ternary scale of 0, 0.5 and 1, where 0 indicates absence of the feature, 0.5 indicates partial support and 1 indicates full support for the feature.

| Criteria | | | Tools | | | | | |
|---|---|---|---|---|---|---|---|---|
| **ID** | **Description** | **Weight** | **WebAnno** | **doccano** | **FLAT** | **BRAT** | **Sangrahaka** | **Antarlekhaka** |
| P1 | Year of the last publication | 0 | 1 | 0 | 0 | 1 | 1 | 1 |
| P2 | Citations on Google Scholar | 0 | 1 | 0 | 0 | 1 | 0 | 0 |
| P3 | Citations for Corpus Development | 0 | 1 | 0 | 0 | 1 | 0 | 0 |
| T1 | Date of the last version | 1 | 1 | 1 | 1 | 0.5 | 1 | 1 |
| T2 | Availability of the source code | 1 | 1 | 1 | 1 | 1 | 1 | 1 |
| T3 | Online availability for use | 1 | 0 | 0 | 1 | 0 | 0 | 0 |
| T4 | Easiness of Installation | 1 | 0 | 1 | 1 | 0.5 | 1 | 1 |
| T5 | Quality of the documentation | 1 | 1 | 1 | 1 | 1 | 0.5 | 0.5 |
| T6 | Type of license | 1 | 1 | 1 | 1 | 1 | 1 | 1 |
| T7 | Free of charge | 1 | 1 | 1 | 1 | 1 | 1 | 1 |
| D1 | Format of the schema | 1 | 1 | 1 | 1 | 0.5 | 1 | 1 |
| D2 | Input format for documents | 1 | 1 | 0.5 | 1 | 1 | 1 | 1 |
| D3 | Output format for annotations | 1 | 1 | 1 | 1 | 0.5 | 0 | 0 |
| F1 | Allowance of multi-label annotations | 1 | 1 | 0 | 1 | 1 | 1 | 1 |
| F2 | Allowance of document level annotations | 0 | 0 | 0 | 0 | 0 | 0 | 0 |
| F3 | Support for annotation of relationships | 1 | 1 | 0 | 0 | 1 | 1 | 1 |
| F4 | Support for ontologies and terminologies | 1 | 1 | 0 | 1 | 1 | 1 | 1 |
| F5 | Support for pre-annotations | 1 | 0.5 | 0 | 0.5 | 0.5 | 0 | 0 |
| F6 | Integration with PubMed | 0 | 0 | 0 | 0 | 0 | 0 | 0 |
| F7 | Suitability for full texts | 1 | 0.5 | 0.5 | 1 | 1 | 1 | 1 |
| F8 | Allowance for saving documents partially | 1 | 1 | 1 | 1 | 1 | 1 | 1 |
| F9 | Ability to highlight parts of the text | 1 | 1 | 1 | 1 | 1 | 1 | 1 |
| F10 | Support for users and teams | 1 | 0.5 | 0.5 | 1 | 0.5 | 0.5 | 0.5 |
| F11 | Support for inter-annotator agreement | 1 | 1 | 0.5 | 0 | 0.5 | 0.5 | 0.5 |
| F12 | Data privacy | 1 | 1 | 1 | 1 | 1 | 1 | 1 |
| F13 | Support for various languages | 1 | 1 | 1 | 1 | 1 | 1 | 1 |
| A1 | Support for querying | 1 | 0 | 0 | 0 | 0 | 1 | 0 |
| A2 | Crash tolerance | 1 | 0 | 0 | 0 | 0 | 1 | 0.5 |
| A3 | Web-based / Distributed annotation | 1 | 1 | 1 | 1 | 1 | 1 | 1 |
| A4 | Sequential Annotation | 1 | 0 | 0 | 0 | 0 | 1 | 1 |
| A5 | Support for Sentence Boundary Annotation | 1 | 0 | 0 | 0 | 0 | 0 | 1 |
| A6 | Support for Word Order Annotation | 1 | 0 | 0 | 0 | 0 | 0 | 1 |
| A7 | Support for Token Classification Tasks | 1 | 1 | 1 | 1 | 1 | 1 | 1 |
| A8 | Support for Sentence Classification Tasks | 1 | 0 | 0 | 0 | 0 | 0 | 1 |
| | **Total** | 29 | 19.5 | 16.0 | 20.5 | 18.5 | 21.5 | **23.0** |
| | **Score** | | 0.67 | 0.55 | 0.71 | 0.64 | 0.74 | **0.79** |

## 4.2 Potential for NLP Research

A general natural language annotation tool has a potential to enable a variety of research in the field of NLP. The primary benefit of such a tool is its ability to aid the creation of datasets for the training and testing machine learning models. This can be directly relevant for various common NLP tasks, such as lemmatization, named entity recognition (NER), part-of-speech (POS) tagging, co-reference resolution, text classification, sentence classification, and relation extraction, among others.

Creating NLP datasets involves annotating text with specific information or labels, which machine learning models can then learn from. The annotation require-ments are often specific and can be tedious. A tool with annotator-friendly and intuitive interfaces can simplify this process to a great extent. High-quality, manually annotated training datasets contribute directly towards improving the accuracy of NLP models.

There are also higher-level NLP tasks such as: Question Answering (QA), where questions in natural language are automatically answered; Grammatical Error Correction (GEC), where errors in written text are automatically detected and corrected; Machine Translation (MT), which involves translating text from one language to another; and Text Summarization, which involves reducing a large piece of text into a shorter, more concise summary. The effectiveness of these higher-level tasks often relies on the success of several low-level tasks.

For example, domain-specific question answering often involves building knowledge graphs [Voorhees, 1999, Hirschman and Gaizauskas, 2001, Kiyota et al., 2002, Yih et al., 2015]. This requires identifying named entities, linking co-references, tagging parts of speech, and identifying dependency relations. These tasks must be performed on the same corpus and the results from each task must be used in conjunction with one another. A multiple task annotation tool can support the annotation of all of these tasks and interlink the data and annotations, leading to a more accurate knowledge graph. The tool's ability to handle large amounts of data and multiple users simultaneously can also contribute to faster completion of these tasks.

## 4.3 Case Study: Annotation of Vālmīki Rāmāyaṇa

Vālmīki Rāmāyaṇa is one of the Itihāsa literature in Sanskrit and holds a wealth of diverse and intricate content. It encompasses a wide range of characters, events, emotions, and settings, providing a comprehensive canvas for annotation. The text of Vālmīki Rāmāyaṇa exhibits linguistic richness, including poetic verses, metaphors, similes, and descriptive passages. It offers a challenging yet rewarding opportunity for linguistic analysis and annotation. Further, it holds immense cultural and historical significance. It serves as a foundation for moral and ethical values, re-ligious beliefs, and social norms in many communities. Annotating this text can help preserve and promote cultural heritage. Over the years, it has inspired various adaptations and translations in different languages and cultures. Annotation can facilitate comparative studies and analysis across different versions, shedding light on its cross-cultural impact. Additionally, Vālmīki Rāmāyaṇa offers ample scope for research in areas such as character analysis, narrative structure, mythological symbolism, ethical dilemmas, and philosophical teachings. Annotation can uncover new dimensions and contribute to scholarly discourse. Therefore, we have undertaken the voluminous task of annotating Vālmīki Rāmāyaṇa.

We use *Antarlekhaka* for this large-scale annotation task. The text is obtained from Digital Corpus of Sanskrit [Hellwig, 2021] and contains 18754 verses (38029 lines) across 606 chapters.

The task will be completed in multiple phases. The annotators participating in the task are graduate or post-graduate level students of Sanskrit with sufficient familiarity with the corpus. In the first phase of the annotation, annotation of 4 tasks per verse is being performed with the help of several annotators. At its peak, there were 51 annotators online and annotating at the same time. So far, 883 verses have been annotated with the help of 26 annotators, amounting to the completion of total 3532 tasks. These annotations have resulted in five task-specific datasets. The annotation of Vālmīki Rāmāyaṇa is an ongoing task, and we are in the continuous process of enriching these datasets. The ontology used for NER and action graph annotation is available at `https://sanskrit.iitk.ac.in/valmikiramayana/ontology/`.

### 4.3.1 Sentence Boundary Dataset

Sentences boundaries in Sanskrit need not coincide with the verse boundaries. A sentence may span across multiple verses, or a single verse may contain multiple sentences. Therefore, the task of marking sentence boundaries is very relevant in the context of Sanskrit NLP. Our sentence boundary dataset contains 1928 sentence markers across 1394 verses.

### 4.3.2 Canonical Word Ordering Dataset

Anvaya, the task of arranging words of a sentence in a manner that is most natural to grasp its meaning, is another important tasks for Sanskrit corpora. We have collected word ordering dataset consisting of 1847 sentences.

### 4.3.3 Named Entity Recognition Dataset

We have also developed a rich ontology consisting of 89 categories for NER. The ontology covers various plant, animal and humanoid species, places, vehicles, weapons, ornaments, instruments, clans, time, and a wide variety of concepts relevant to Vālmīki Rāmāyaṇa. This ontology has been used for the identification and classification of 1644 named entities from 886 verses.

### 4.3.4 Co-reference Resolution Dataset

The text of Vālmīki Rāmāyaṇa is rich with several characters and their interaction with each other. It serves as a good corpus for the task of co-reference resolution. We have created a dataset for co-reference resolution which consists of 2226 co-reference connections.

### 4.3.5 Action Graph Dataset

An action graph is a sentence-level graph that captures 'action words', i.e., verbs, participles, and any other words that denote actions, and their relations with other words in the sentence. We first create a list of 44 relations through which an action word may connect to other words. This compilation is an outcome of minor adaptations made to the dependency tagset for Sanskrit developed by [Kulkarni, 2020]. Using this set of relations, We have collected 29 action graphs consisting 250 relations from 25 verses.

## 4.4 Summary

We have developed a web-based multi-task annotation tool called *Antarlekhaka* for sequential annotation of various NLP tasks. The tool is language-agnostic and has a full Unicode support. The tool sports eight categories of annotation tasks and an annotator-friendly interface for each category of task. Multiple annotation tasks from each category are supported. The tool enables creation of datasets for computational linguistics tasks without expecting any programming knowledge from the annotators or administrators. The tool has a potential to propel several research opportunities in the field of natural language processing.

The tool is actively being used for a large-scale annotation project, involving a large Sanskrit corpus and a significant number of annotators, as well as another annotation task in Bengali language. The tool is also actively maintained. In the future, we plan to integrate various state-of-the-art NLP tools to add out-of-the-box support for several languages and aid annotators by providing suggestions.

# Chapter 5

# Chandojñānam: Sanskrit Meter Identification and Utilization

Majority of the Sanskrit literature is in the form of poetry that adheres to the rules of Sanskrit prosody or Chandaḥśāstra, which is the study of Sanskrit *meters*, known as chandas. The purpose of chanda is primarily to add rhythm to the text so that it is easier to memorize. Additionally, it also helps in preserving the correctness to some extent.

We now present Chandojñānam, a web-based Sanskrit meter (chanda) identification and utilization system. In addition to the core functionality of identifying meters, it sports a friendly user interface to display the scansion, which is a graphical representation of the metrical pattern. The system supports identification of meters from uploaded images by using optical character recognition (OCR) engines in the backend. It is also able to process entire text files at a time. The text can be processed in two modes, either by treating it as a list of individual lines, or as a collection of verses. When a line or a verse does not correspond exactly to a known meter, Chandojñānam is capable of finding fuzzy (i.e., approximate and close) matches based on sequence matching. This opens up the scope of a meter based correction of erroneous digital corpora. The system is available for use at `https://sanskrit.iitk.ac.in/jnanasangraha/chanda/`, and the source code in the form of a Python li-brary is made available at `https://github.com/hrishikeshrt/chanda/`.

## 5.1 Introduction

The digitization of Sanskrit text is achieved primarily through two methods: (1) manual data entry and (2) scanning of documents followed by optical character recognition (OCR). The former suffers from human error while the latter is prone to inaccuracies due to automated processing. Further, with the rise of social media, blog sites and Unicode, there is a large replication of Sanskrit text on the Internet with little quality control, thereby further increasing errors in the text.

We argue that a non-trivial portion of the errors introduced in texts through various sources can be detected by the process of meter identification. The Chandojñānam system exhibits tolerance towards erroneous texts and is able to locate the errors as well as make suggestions for fixing them.

### 5.1.1 Motivation

The motivation behind Chandojñānam can be better understood with the following scenarios.

- A Sanskrit enthusiast wants to identify the meter of a verse from a PDF file with Sanskrit text. She however, lacks the capability to effectively type the Sanskrit text, and prefers to upload a screenshot of the specified verse to identify the meter. With Chandojñānam, she is able to perform meter identification directly from the image.

- A teacher of Chandaḥśāstra wants to explain the rules of Sanskrit prosody to her students using several examples. She does not want to spend precious time on writing down all the intermediate steps in the identification of a meter. Instead, she wants a system that will do this job for her. Chandojñānam fits the bill nicely.

- A budding poet is trying to compose Sanskrit poetry adhering to a specific meter. However, she is not an expert, and may have made some errors. She wants to locate these errors so that she can correct it. Chandojñānam lets her locate these errors, and also provides suggestions for correction.

- A Sanskrit researcher wants to analyse a large Sanskrit text file and obtain metrical statistics of the entire corpus. Chandojñānam allows her to upload the text file and quickly obtain the required statistics to aid her in her research.

These examples highlight the utility of Chandojñānam, in addition to just satisfying one's curiosity about Sanskrit meters.

### 5.1.2 Background

The classification of syllables into laghu (*short*) and guru (*long*) forms the core concept of Chandaḥśāstra [Deo, 2007]. The classification is related to the amount of time it takes to pronounce a specific syllable; in particular, the short syllables are termed laghu and the long syllables guru. Specific sequences or combinations of laghu and guru letters result in a particular rhythm or chanda.

A verse (śloka) is composed of four parts, each known as a pāda. Every meter has a constraint on the sequence of syllables that should be followed in a pāda. For example, the meter Pañcacāmara is defined by each pāda having 16 syllables with the following laghu-guru sequence of syllables: `LGLGLGLGLGLGLGLG`[1]. The perfect alternation of laghu and guru syllables is a reason for its energetic tempo, as can be experienced in the hymns such as Śivatāṇḍavastotram[2] or Narmadāṣṭakam[3]. In a similar manner, numerous Sanskrit meters have been defined in the texts on Sanskrit prosody such as Vṛttaratnākara.

The unique sequence of laghu-guru markers used to identify a meter is hereon referred to as *lg-signature* of that meter. The term, in reference to an arbitrary Sanskrit line, is used to refer to its decomposition in the laghu-guru sequence. Chandaḥśāstra, for brevity and ease of remembering, identifies all laghu-guru sequences of length $3$ with a unique letter. This unique 3-length sequence is known as a Gaṇa. The possible number of gaṇas is $2^3 = 8$. More details about the terminology and rules of Chandaḥśāstra can be found at `https://sanskrit.iitk.ac.in/jnanasangraha/chanda/help`. Several previous works [Deo, 2007, Mishra, 2007, Melnad et al., 2013] also discuss the theory of Chandaḥśāstra in detail.

[1] We use letters `L` and `G` to denote laghu and guru syllables respectively.
[2] `https://shlokam.org/shivatandavastotram/`
[3] `https://shlokam.org/narmadashtakam/`

### 5.1.3 Related Work

There have been several efforts in the area of automatically identifying meter from Sanskrit text. Some of these tools [Mishra, 2007, Melnad et al., 2013] were only presented as web interfaces, which are no longer functional.

More recent works [Rajagopalan, 2020, Neill, 2023] provide both a web interface and a software library. However, the web interfaces provided by [Rajagopalan, 2020] and [Neill, 2023] both assume that a single verse will be provided as an input, limiting its usability to quickly check meters for a set of verses. Thus, even for a text consisting of a small number of verses, the input has to be provided as many times as there are number of verses. [Neill, 2023] attempts to address this issue by allowing upload of text files. However, it still lacks a display in case one wants to visualize meters for multiple verses. More significantly, identifying the meter of a single pāda or a partial verse is not possible at all, as any text entered is assumed to consist of exactly four pādas.

Error tolerance, i.e., the capacity to identify the meter of a verse in the presence of errors, is an important requirement for a meter identification system. To be useful for error correction, a system should be further able to detect exact locations of error and suggest corrections. [Neill, 2023] uses a scoring system resembling majority rule on the pādas of a verse to identify its meter. However, it assumes pāda matching to be exact. In other words, even if there is a single error in a pāda, the system reports that pāda as a mismatch. As a result, when there are errors in three pādas, due to the majority of pādas being mismatched, the overall meter for the verse is reported as '*unknown*'. Further, it lacks any suggestion mechanism. Barring the admirable work by [Rajagopalan, 2020], none of the other tools attempts to perform *fuzzy matching* (i.e., *approximate matching*) to handle erroneous text. The chanda suggested by [Rajagopalan, 2020] is, however, the one that is perceived as the best by its fuzzy matching algorithm, which may not always be the actual chanda of the verse. Therefore, providing a top-$k$ ranked list of matches is more useful.

**Table 5.1:** Feature comparison of extant meter identification systems

| **Features** | | [Mishra, 2007] | [Melnad et al., 2013] | [Rajagopalan, 2020] | [Neill, 2023] | Chandojñānam |
|---|---|---|---|---|---|---|
| Availability | Web Interface | ✓[4] | ✓[5] | ✓ | ✓ | ✓ |
| | Software Library | | | ✓ | ✓ | ✓ |
| Input | Text | ✓ | ✓ | ✓ | ✓ | ✓ |
| | Arbitrary Lines | | | | | ✓ |
| | Multiple Verses | | | | | ✓ |
| | Textfile Upload | | | | ✓ | ✓ |
| | Image Upload | | | | | ✓ |
| Functionality | Meter Identification | ✓ | ✓ | ✓ | ✓ | ✓ |
| | Error Tolerance | | | ✓ | ✓ | ✓ |
| | Fuzzy Matching | | | ✓ | | ✓ |

Further, none of the existing tools has a functionality to upload images for meter identification.

Chandojñānam attempts to overcome the shortcomings of the previous tools by providing a comprehensive set of user-friendly features. There is a special focus on *fuzzy matching* (explained in Section 5.2.4.2) and the utility of Chandaḥśāstra for correction of digital corpora.

Table 5.1 presents a feature matrix comparing the other works with Chandojñānam.

### 5.1.4 Contributions

We present, Chandojñānam, a Sanskrit meter identification and utilization system, which in addition to identifying a meter, also aims to catch errors in the text and suggest corrections. The aim of the tool is to make this process easy for a non-programmer. The salient features of Chandojñānam can be summarized as follows:

[4] `http://sanskrit.sai.uni-heidelberg.de/Chanda/HTML/` is no longer functional.
[5] `https://sanskritlibrary.org:8080/MeterIdentification/` is no longer functional.

- The tool is available as a web-application that can be used from any standard browser and requires no other installation from the user. The library is also made available for the benefit of programmers.
- There are three prominent input options: (1) plain text, (2) images (screenshots) and (3) text files.
- The input can be in any of the standard transliteration scheme that can be detected by the *indic-transliteration* [Sanskrit programmers, 2021] library. This applies to all three input methods.
- The image files can be processed using one of the two OCR engines, namely, Google OCR and Tesseract OCR, and the detected text can be further edited.
- The lines from input are detected based on several standard line-end markers, such as '\n', '।', '॥' and '.'
- Meter identification can be performed on the line (pāda) level. The input is treated as a set of lines. Therefore, the input can be any arbitrary set of pādas.
- Meter identification can also be performed on a verse (śloka) level. In this case, the system treats the lines as being parts of a verse. Cumulative cost of each line is minimized to identify the meter of the verse. Multiple verses can be provided.
- For erroneous inputs, the tool provides a robust fuzzy matching support using edit distance comparison, which helps in identifying and highlighting the places where an error might be present. Additionally, there is a suggestion module to help the user understand what changes can be made to the input.
- Informative display shows the steps involved in the meter identification, which is aimed to help learners of Chandaḥśāstra.
- Results can be downloaded in the JSON format.

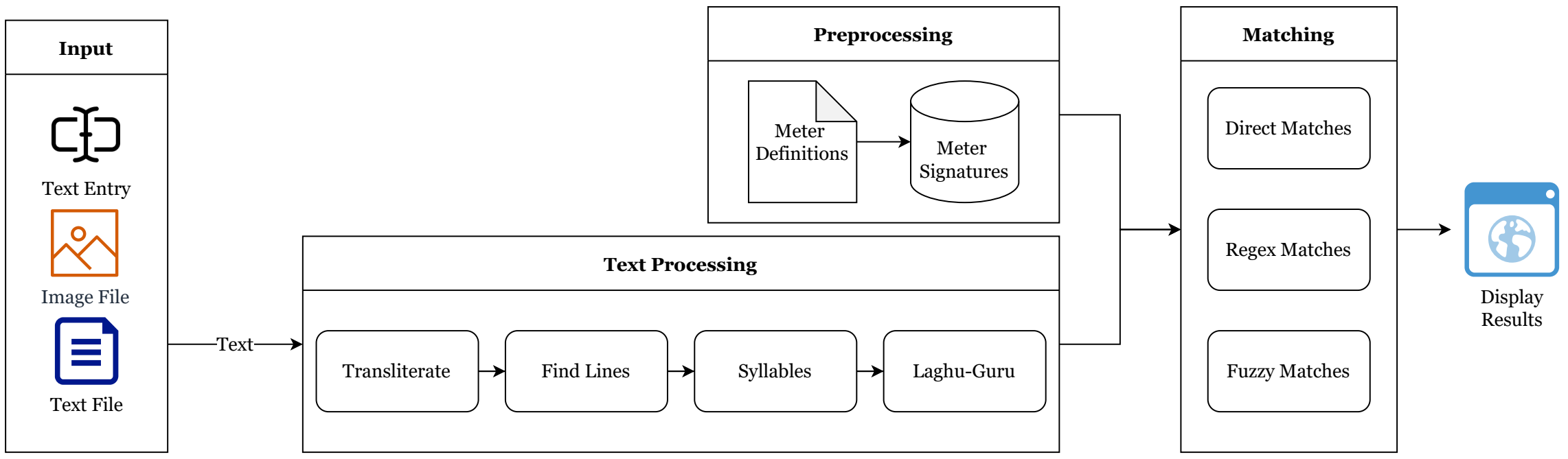

**Figure 5.1:** Workflow of the Chandojñānam system

The Chandojñānam system can be accessed online at `https://sanskrit.iitk.ac.in/jnanasangraha/chanda/`. The source code is a available at `https://github.com/hrishikeshrt/chanda/`.

## 5.2 The Chandojñānam System

In this section, we discuss the inner workings of the Chandojñānam system. Figure 5.1 illustrates the overall workflow of the system. Initially, definitions of Sanskrit meters are read into the system and stored in the form of a dictionary, referred to hereon as the *metrical database*. A user may specify the input in any of the three specified formats, namely, plain text, image containing text, and text file. For image, the text is extracted using OCR systems. After the text is extracted, the transliteration scheme is detected, and converted to the internal transliteration scheme, which is 'Devanagari'. From the text, lines are detected which, in turn, are split into syllables to obtain the *lg-signature* of the lines. The metrical database is then queried using the meter detection algorithm and the matches are presented to the user along with useful information. In case of erroneous inputs, $k = 10$ closest matches are also shown, along with the suggestions for corrections.

### 5.2.1 Chanda Definitions

There are two main types of meters, Varṇavṛtta and Mātrāvṛtta [Melnad et al., 2013]. Currently, the system only deals with Varṇavṛtta. In future, we will enable the sys-tem to handle Mātrāvṛtta as well. Varṇavṛtta can be further divided into three categories, namely, Samavṛtta, Ardhasamavṛtta and Viṣamavṛtta. This categorization is performed based on the symmetry or lack thereof of the metrical pattern exhibited by the four pādas of a śloka adhering to the meter. All four pādas following the same pattern is termed as Samavṛtta, odd and even pādas following a different pattern is termed as Ardhasamavṛtta, and all four pādas following a different pattern corresponds to Viṣamavṛtta.

| वृत्त | पाद | गण | लक्षण | अक्षरसङ्ख्या | मात्रा | यति |
|---|---|---|---|---|---|---|
| शार्दूलविक्रीडित | | मसजसततग | गगगललगलगललगगगलगगलग | 19 | 30 | 12,7 |
| शालिनी | | मततगग | गगगगगलगगलगग | 11 | 20 | 4,7 |
| अपरवक्त्र | 1 | ननरलग | लललललललगलगलग | 11 | 14 | |
| अपरवक्त्र | 2 | नजजर | ललललगललगलगलग | 12 | 16 | |
| सौरभ | 1 | सजसल | ललगलगललललगल | 10 | 13 | |
| सौरभ | 2 | नसजग | ललललललगलगलग | 10 | 13 | |
| सौरभ | 3 | रनभग | गलगललललगललग | 10 | 14 | |
| सौरभ | 4 | सजसजग | ललगलगललललगलगलग | 13 | 18 | |
| अनुष्टुभ् | 1 | ----लग-- | ----लग-- | 8 | | |
| अनुष्टुभ् | 2 | ----लगल- | ----लगल- | 8 | | |

**Figure 5.2:** Generic chanda definition format

**Table 5.2:** Chanda Definitions specification format

| Column | Requirement | Description |
|---|---|---|
| Vṛtta | required | Name of the meter described in the row |
| Pāda | required | Index of pāda in the corresponding vṛtta; the possible values are <blank>, 1, 2, 3 and 4. |
| Lakṣaṇa | required | *lg-signature* of the meter |
| Gaṇa | optional | Signature of the meter in the compressed (trika) notation |
| Akṣarasaṃkhyā | optional | Number of letters in the pāda |
| Mātrā | optional | Number of mātrās in the pāda |
| Yati | optional | Indices corresponding to yati |

Chanda definitions are specified in the tabular format as illustrated in Figure 5.2. The legend is described in Table 5.2. It can be noted that the value corresponding to the column pāda denotes the index of pāda in the meter described in that row. This way of specifying definitions results in a uniform treatment of samavṛtta, ardhasamavṛtta and viṣamavṛtta. Additionally, a regex pattern (*regular expression*) defi-nition can also be specified, where the metrical restriction only applies to a part of the pāda. The most commonly used meter in Sanskrit, namely, anuṣṭubh, is a prime example of such regex patterns. Specifically, the requirements for anuṣṭubh chanda are:

- Every pāda must contain exactly 8 syllables.
- The 5th syllable of every pāda must be a laghu syllable.
- The 6th syllable of every pāda must be a guru syllable.
- The 7th syllable of the even pādas must be a laghu syllable.
- There is no other restriction on the other syllables, i.e., they can be either laghu or guru.

Therefore, the two regex patterns `[LG][LG][LG][LG]LG[LG][LG]` and `[LG][LG][LG][LG]LGL[LG]` correspond to the *lg-signatures* for the odd numbered and even numbered pādas of Anuṣṭubh respectively.

Internally, the meter database is stored in the form of dictionaries in two ways: signature of individual pādas is stored as `CHANDA_SINGLE`, while signature of consecutive pādas is stored as `CHANDA_MULTIPLE`. It may often happen that two pādas of the input verse are given as a single line without any kind of line marker. The second dictionary acts as a fail-safe when the lines couldn't be split in a proper manner due to lack of punctuation or line-breaks.

For example, for the meter Bhujaṅgaprayāta, which has a signature ययय[6], corresponding to the *lg-signature* of `LGGLGGLGGLGG`, two independent entries are maintained.

```
CHANDA_SINGLE = {
    'LGGLGGLGGLGG': ['Bhujaṅgaprayāta'],
    '[LG][LG][LG][LG]LG[LG][LG]': ['Anuṣṭubh (Pāda 1)'],
    '[LG][LG][LG][LG]LGL[LG]': ['Anuṣṭubh (Pāda 2)']
```

[6]Gaṇa य corresponds to the *lg-sequence* LGG.

```
}

CHANDA_MULTIPLE = {
    'LGGLGGLGGLGGLGGLGGLGGLGG': ['Bhujaṅgaprayāta (Pāda 1-2)'],
    '[LG][LG][LG][LG]LG[LG][LG][LG][LG][LG][LG]LGL[LG]':
                                  ['Anuṣṭubh (Pāda 1-2)']
}
```

The system currently contains a database of over 200 meters, which is smaller in comparison to the database utilized [Rajagopalan, 2020]. However, it is important to note that the quantity of data does not always guarantee improved performance, as discussed in detail by [Rajagopalan, 2020]. The primary reason for not importing a large number of meters is that the lower number results in less false positives when dealing with erroneous input. More meters may be included in the system in future based on user feedback.

### 5.2.2 Input

The input for meter identification is Sanskrit text, and it can be provided to our system in three ways. The simplest form of input is a direct text entry, wherein, a user may type or copy-paste Sanskrit text in any valid transliteration scheme into a text input box.

Perhaps a more useful option, which is a novelty of our tool, is the ability to process images containing text. Two well-known OCR engines, Google Drive OCR[7] and Tesseract OCR[8] [Kay, 2007] are supported. Google Drive OCR functions by making use of Google's Drive API v3[9] to upload files to Google Drive, while Tesseract OCR v5 is a neural network (LSTM) based OCR engine. Python libraries `google-drive-ocr`[10] [Terdalkar, 2022] and `pytesseract`[11] [Lee, 2022] are utilized respectively to access the OCR systems. Google Drive OCR is generally more accurate than Tesseract OCR but slower due to the network latency imposed by upload of files. The output of either of the OCR systems, i.e., the recognized text, is treated as if directly entered by the user. The output is not always accurate, especially if the image contains noisy text. Therefore, the system also lets user edit the optically recognized text and re-submit it to the system. Figure 5.3 shows the interface to upload images for meter identification.

[7] https://support.google.com/drive/answer/176692?hl=en
[8] https://github.com/tesseract-ocr/
[9] https://developers.google.com/drive/api/v3/reference
[10] https://github.com/hrishikeshrt/google_drive_ocr
[11] https://github.com/madmaze/pytesseract

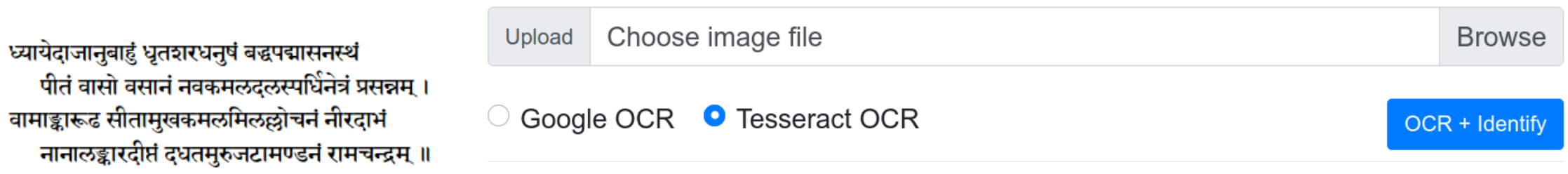

**Figure 5.3:** Upload a screenshot of a verse to Chandojñānam for meter identification

For bulk processing, the option to upload a text file is also available. After reading the uploaded file, the text is again treated as if directly entered by the user.

The identical treatment of all input allows the same meter identification pipeline to follow. For example, as the transliteration module is triggered after the processing of input text, all three input methods have full support for input in any valid transliteration scheme.

### 5.2.3 Text Processing

Once the text input is obtained and cleaned, it passes through four primary processing steps.

1. **Transliteration:** The task of transliterating Devanagari text can be considered as a solved problem thanks to the presence of robust transliteration tools such as *indic-transliteration*[12] [Sanskrit programmers, 2021] and *Aksharamukha*[13] [Rajan, 2018]. We rely on *indic-transliteration* for detection of input scheme; thus, any transliteration scheme detectable by *indic-transliteration* is also supported by our system. The internal transliteration scheme is set to 'Devanagari', which is a convenience choice.

[12] https://pypi.org/project/indic-transliteration/
[13] https://aksharamukha.appspot.com/

2. **Pāda Split:** The contiguous text is now split into lines (pādas) based on several standard line-end markers, such as '\n', '।', '॥' and '.'. Chandojñānam uses pāda as a unit for meter identification.

3. **Syllabification**: The process of splitting text into syllables is fairly straightforward for majority of Indian languages due to phonetic consistency of the alphabet. The Devanagari Unicode has special vowel markers for all vowels except 'अ' (a). The presence of vowels with consonants, therefore, can be easily identified with these markers, and syllables can be separated. The absence of vowel marker for the vowel 'अ' (a) can be treated by noting that for a joint consonant, Unicode always uses the halanta marker '्'. So, two consecutive consonant characters indicate the presence of a vowel 'अ' with the first consonant. For example, the Unicode string "भारत" consists of 4 Unicode characters: 'भ', 'ा', 'र' and 'त'. However, it corresponds to 6 Devanagari letters (varṇa), namely, 'भ्', 'आ', 'र्', 'अ', 'त्' and 'अ'. Every vowel signifies an end of a syllable, resulting in three syllables, namely, 'भा', 'र' and 'त'. Thus, in a single scan of the string, syllabification can be performed. We use the `sanskrit-text`[14] Python library to perform this task.

4. **Laghu-Guru Marks**: The syllabification process is a prerequisite for obtaining the *lg-signature* of a Sanskrit line. Standard rules of Sanskrit prosody described in numerous articles [Deo, 2007, Melnad et al., 2013, Rajagopalan, 2020] are followed to mark each syllable as laghu or guru. A general rule of Piṅgala's Chandaḥśāstra states that the last syllable of a pāda should be treated as a guru. However, there exist meters in the Chandaḥśāstra whose signatures contain the last syllable as laghu (Pādānta Laghu). Therefore, we compute the *lg-signature* without forcing the last letter to be guru.

This completes the text processing which makes the text ready for the next stage, which is meter identification.

[14] `https://pypi.org/project/sanskrit-text/`

**Algorithm 1:** Meter Identification

**Data:** Metrical Database ($MD$)
**Input:** *lg-signatures* corresponding to each 'line' in the input ($T = \{lg_1, lg_2, \ldots, lg_n\}$)
**Output:** Result set containing exact or fuzzy matches

**forall** $lg \in T$ **do**
    $SM_1$ = `FindDirectMatch(`$lg$, *'CHANDA_SINGLE'*`)`
    $SM_2$ = `FindDirectMatch(`$lg$, *'CHANDA_MULTIPLE'*`)`
    $RM$ = `FindRegexMatch(`$lg$, *'CHANDA_SINGLE'* + *'CHANDA_MULTIPLE'*`)`
    $DM = SM_1 + SM_2 + RM$
    $FM = \phi$
    **if** $DM = \phi$ **then**
        $FM$ = `FindFuzzyMatch(`$lg$`)`
    **end**
    **return** $DM + FM$
**end**

**Algorithm 2:** Direct Matching

**Input:** *lg-signature*
**Output:** Result set containing exact matches

**Function** `FindDirectMatch(`*lg*, *'MD'*`)`
    $M_1$ = Query($lg$, 'MD')    `// dictionary lookup`
    $M_2 = \phi$
    **if** $M_1 = \phi$ **then**    `// if no match found`
        **if** *the last letter of* $lg$ *is* laghu **then**
            $lg_1$ = *replace last letter of* $lg$ *with* guru
            $M_2$ = Query($lg_1$, 'MD')
        **end**
    **end**
    **return** $M_1 + M_2$

### 5.2.4 Meter Identification Algorithm

The *meter identification* problem, in its simplest form, is a dictionary lookup on *lg-signatures* of meters. In practice, however, there can be several additional considerations. The algorithm used by our system is described in Algorithm 1. The algorithm consists of a direct dictionary lookup, regular-expression based lookup and fuzzy matching. These are explained in the subsequent sections.

#### 5.2.4.1 Direct Matching

Finding a direct match involves trying to find the match in the both the dictionaries in the metrical database that store the meter signatures. Further, for meters with regex specification, regular-expression-based matching is performed.

As mentioned in Section 5.2.3, we compute the *lg-signature* without enforcing the last syllable to be guru. If the last syllable was laghu and no direct match was found, we attempt to find a direct match by treating the last syllable as guru. Algorithm 2 describes this process. The function `Query` corresponds to a simple dictionary lookup and returns a set of matches.

In theory, this should be sufficient. However, in practice, one often encounters erroneous text. This is expected since a large amount of Sanskrit text available digitally is a result of either manual entry or post-scanning OCR followed by manual correction. As a result, due to human errors or OCR inaccuracies, errors such as the following may be found in the texts available online [Sankaran et al., 2013, Kumar and Lehal, 2016].

- Characters may be misspelt, e.g., रु (ru) as रू (rū)
- Characters may be missing, e.g., वर्गै (vargai) as वगै (vagai)
- Characters may be misidentified, e.g., ऋ (ṛ) as क्र (kra)
- Characters may get split, e.g., ख (kha) as रव (rava)

Most of these errors may also affect the metrical pattern of a line and, therefore, meter identification is a useful tool to identify them. It should be mentioned here that not all errors lead to metrical mismatch. For example, an error such as the confusion between letter pairs such as व, ब or म, स would not affect the meter pattern and, thus, cannot be captured by only meter identification.

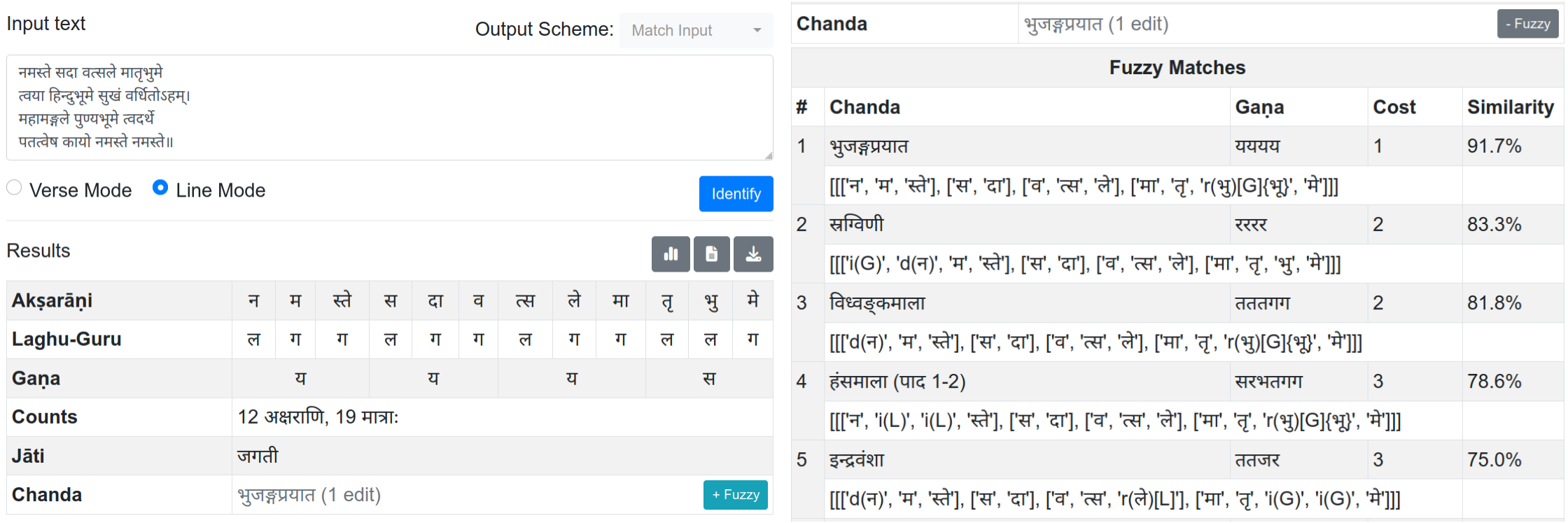

**Figure 5.4:** Meter identification with fuzzy matching and suggestions

**Algorithm 3:** Fuzzy Matching

**Input:** *lg-signature* and $k$

**Output:** Result set containing top-$k$ fuzzy matches

```
Function FindFuzzyMatch(lg, 'MD', k)
    results = ϕ
    forall chanda in 'MD' do
        cost, suggestion = Transform(lg, lg-signature of chanda)
        if suggestion then
            results += (chanda, cost, suggestion)
        end
    end
    return top-k results with lowest cost
```

#### 5.2.4.2 Fuzzy Matching

For non-exact matches, we model the problem as that of finding the *nearest matching string* for the *lg-signature* of the text. In particular, we compute the Levenshtein[15] edit distance of the observed pattern with all the known patterns. We then find the similarity by first normalizing the edit distance using the length of the target match, and then subtracting it from 1.

$$\text{Similarity} = 1 - \frac{\text{Levenshtein distance}}{\text{length of target match}}$$

We present the topmost $k$ matches as the possible fuzzy matches (currently, $k = 10$). Algorithm 3 and Algorithm 4 describes the core functions pertaining to fuzzy matching. The *Similarity* column in the interface (Figure 5.4) shows the similarity

[15] https://en.wikipedia.org/wiki/Levenshtein_distance

**Algorithm 4:** Transform String Sequences

**Input:** String sequences $seq_1$, $seq_2$

**Output:** Suggested edit operations required to transform $seq_1$ into $seq_2$ and the cost of transformation

1 **Function** `Transform(`$seq_1$, $seq_2$`)`

2 edit_ops = GetLevenshteinEditOps($seq_1$, $seq_2$) `// using python-Levenshtein`

3 weights = {‘replace’: 1, ‘delete’: 1, ‘insert’: 1}

4 cost = $\sum_{op \in \text{edit_ops}}$ weights[op]

5 suggestion = AddEditOpsMarkers($seq_1$, edit_ops) `// add suggestions`

6 **return** cost, suggestion

instead of the edit distance.

For finding the edit-distance and edit-operations, we use the *python-Levenshtein*[16] library [Haapala, 2014]. The `AddEditOpsMarkers` function (Line 5) formats the original list of syllables by adding suggestions based on the edit operations. The suggestions shown contain the characters from the original string, along with the suggested places where a change might be required. Suggested changes for each meter correspond to the changes needed to transform the given line into that meter.

The suggestions are identified by three letters `i`, `d` and `r`, indicating *insert*, *delete* and *replace* operations respectively. They are used in the following manner.

- `i(L/G)`:

  This notation indicates the insertion of a new syllable. `i(L)` (respectively, `i(G)`) indicates that a laghu (respectively, guru) syllable needs to be inserted.

- `r(letter)[L/G]{suggestion}`:

  This notation indicates that the syllable needs to be replaced by a laghu or a guru syllable (depending upon the marker). In some of the cases, a suggested change of syllable may be included. A simple heuristic that is followed replaces the laghu syllables corresponding to the laghu vowels ‘इ’, ‘उ’ and ‘ऋ’ by their guru vowel counterparts, namely, ‘ई’, ‘ऊ’ and ‘ॠ’, respectively, and vice versa.

[16] `https://pypi.org/project/python-Levenshtein/`

- `d(letter)`:

  This notation indicates that the syllable needs to be deleted.

The process of fuzzy matching can be better understood with the help of an example. Consider the first line of the example depicted in Figure 5.4.

- Line: नमस्ते सदा वत्सले मातृभुमे
- LG-Signature: `LGGLGGLGGLLG`
- Nearest Match: `LGGLGGLGGLGG` (cost 1) (Bhujaṅgaprayāta)
- Suggestion: [['न', 'म', 'स्ते'], ['स', 'दा'], ['व', 'त्स', 'ले'], ['मा', 'तृ', 'r(भु)[G]{भू}', 'मे']]

The exact match for the line is not found in the system and, therefore, we find the edit-distance of this *lg-signature* with every *lg-signature* from the database. It is found that the *lg-signature* `LGGLGGLGGLGG` of Bhujaṅgaprayāta meter is the closest and is only 1 edit-distance away. The required change is to replace the 11th letter from `L` (laghu) to `G` (guru). It is also suggested[17] to replace the laghu letter भु into a guru letter भू.

#### 5.2.4.3 Verse Processing

Meter identification can be performed by treating the input either as a set of arbitrary lines (*Line mode*) or as a collection of verses (*Verse mode*). The *Line mode* is useful for checking meter of a single line or a set of lines. The treatment of the text as a verse differs insofar as it attempts to minimize the cumulative cost of the meter matches over each line.

Consider a sample verse adhering to the meter Śālinī, albeit with a deliberate small error in the first pāda. Figure 5.5 highlights the difference between the two modes of meter identification. The correct word मत्पिता has been replaced by two words मम पिता. As a result, the first pāda does not find an exact match. Further, the closest match based on edit distance is found to be the meter Vātormī with an edit distance of 1. Therefore, when in the *Line mode*, the suggestion is made as Vātormī. However, rest of the three pādas are exact matches for Śālinī. Therefore, the cumulative cost of Śālinī over the entire śloka is the lowest (here, it is 2). Thus, in the *Verse mode*, the identified meter for the verse is Śālinī. This highlights the utility of the *Verse mode*.

[17] The suggestion module is currently limited to the suggestions of the nature of toggling between a laghu and a guru vowel marker.

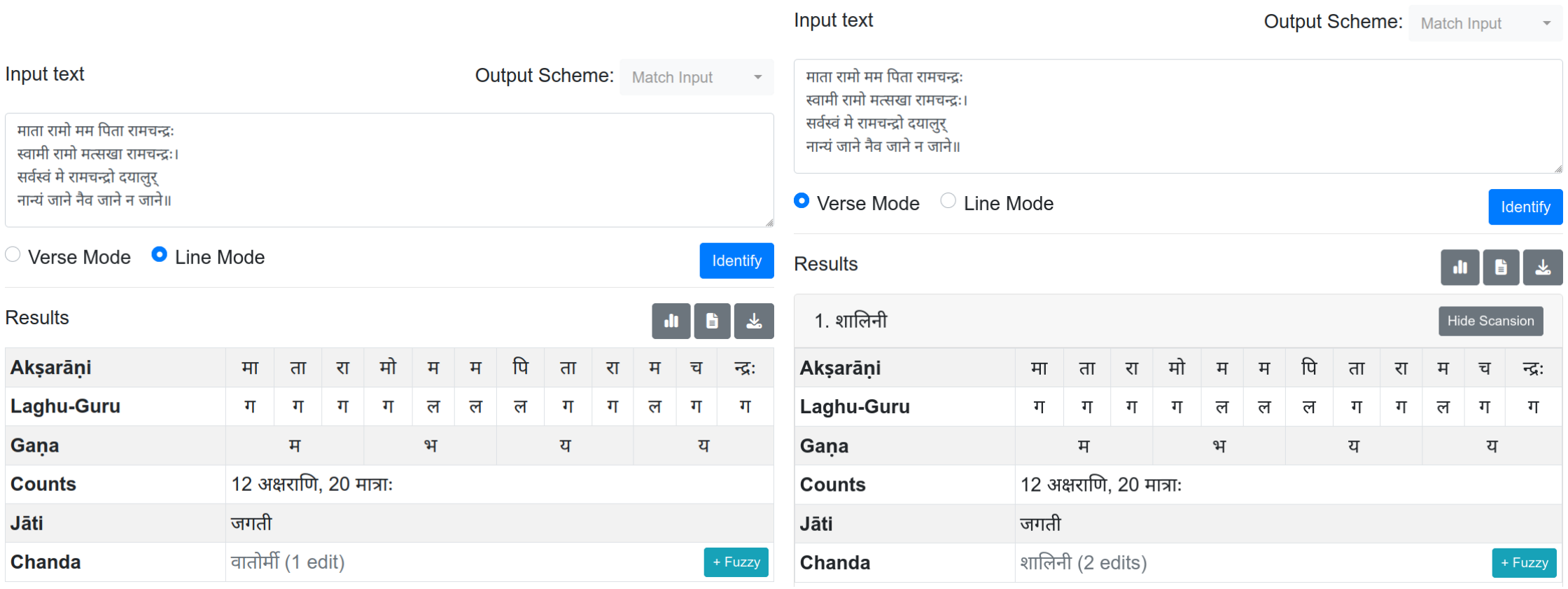

**Figure 5.5:** Meter identification from a verse in (a) *Line mode* and (b) *Verse mode*

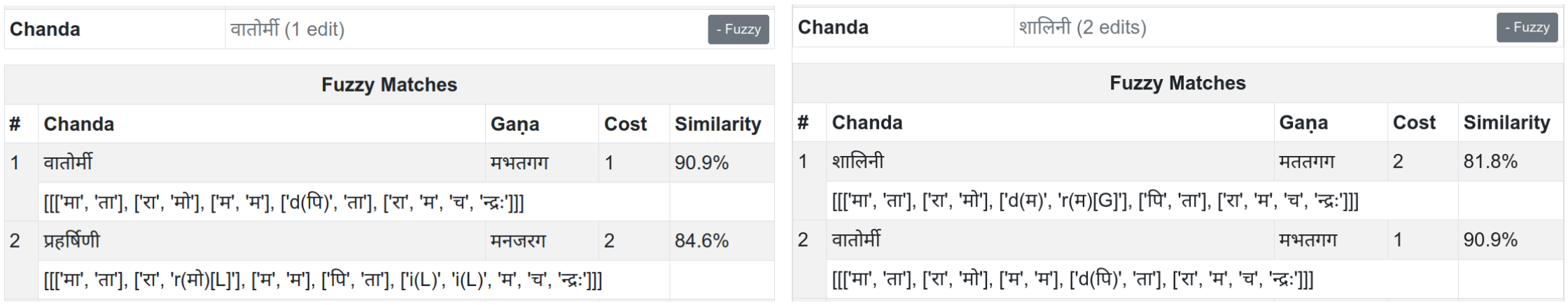

**Figure 5.6:** Fuzzy matches in (a) *Line mode* and (b) *Verse mode*

In the list of fuzzy matches shown, the meter identified for the verse is shown first. Figure 5.6 shows that the meter Śālinī is shown for the first pāda despite having a higher edit cost.

### 5.2.5 Output

A blackbox that only identifies meter would not be as interesting as a system that also explains ('teaches') the steps. Therefore, in addition to the meter name, scansion details such as *lg-signature*, gaṇa-signature, count of letters and count of mātrās are also displayed. Additional information displayed in the case of fuzzy matches is already discussed in Section 5.2.4.2. Entire result is displayed in a neat tabular format (as displayed in Figure 5.4) which makes the system useful for learners of Chandaḥśāstra.

Cumulative statistics such as number of lines identified, number of lines not identified, frequency distribution of exact and fuzzy matches etc. can also be viewed. Additionally, the data can also be downloaded. The data download supports two formats, a compact format for quick visual inspection, and a detailed JSON format for further computational processing.

### 5.2.6 Utility

We can identify several use-cases for Chandojñānam, some of which have already been discussed in Section 5.1.1:

- The primary use-case, naturally, is the ability to identify meters from Sanskrit text.
- Ability to upload of images and identification of meters from it is another important scenario.
- Identification of errors from the Sanskrit text which has been created through the process of OCR, but not manually corrected yet, serves as another prominent use-case.
- Additionally, the system can be used to obtain metrical statistics from large texts automatically. These statistics may also help a user identify errors, if any, at a quick glance. For example, in the case of texts such as Meghadūta, which follows a single meter, namely, Mandākrāntā, any anomaly will be quickly spotted by appearance of a different meter in the statistics.
- A user may be interested in creating poetry in Sanskrit, and may require assistance in quickly checking the metrical consistency of her creation.
- Learners of Chandaḥśāstra may want to try out several examples to understand the process of meter identification in-depth.

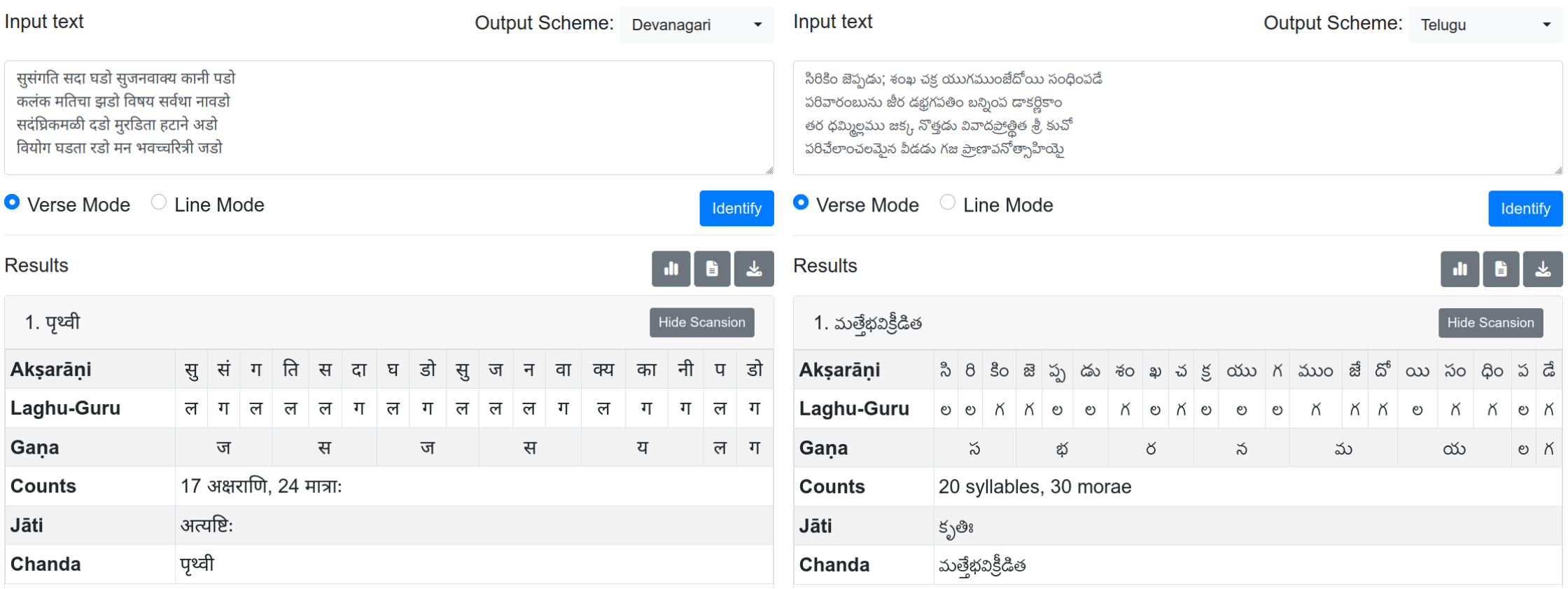

**Figure 5.7:** Meter identification from other Indian languages, e.g., (a) Marathi (b) Telugu

- Several Indian languages, e.g., Marathi, Telugu, etc., exhibit rules of prosody similar to Sanskrit. Due to comprehensive transliteration support provided by the *indic-transliteration* library, text written in scripts other than Devanagari can be also used in the same manner. Figure 5.7 illustrates the usage of Chandojñānam for meter identification from Marathi and Telugu. Here, we are assuming that the same metrical database is used. However, there may be language-specific differences in the rules of prosody as well as variety of meters. Hence, the multilingual support is still primitive.

## 5.3 Evaluation for Error Correction

One of the primary goals of Chandojñānam is to facilitate error detection from the Sanskrit text obtained from various sources. Consequently, we evaluate the ability of the tool to correctly identify the meters from erroneous text.

### 5.3.1 Corpus

Evaluation of a meter identification system requires tagged metrical data. As mentioned by [Rajagopalan, 2020], Meghadūta composed by Kālidāsa [Kale, 2011], being entirely in Mandākrāntā meter, is an ideal corpus for evaluation. The same text, available from various sources, can also differ greatly in terms of encoding, character-by-character comparison and errors present. We use three online sources, namely,

Wikisource[18], sanskritdocuments.org[19] and GRETIL[20] to obtain three different versions of Meghadūta.

In addition to Meghadūta, we also use texts with more metrical variety. We choose three texts from Wikisource, namely, Śāntavilāsa[21], Śrīrāmarakṣāstotra[22] and Rājendrakarṇapūra[23]. We manually tag meters for each verse from these texts. Together, the 4 datasets contain 1038 verses, exhibiting 17 distinct meters.

Further, to evaluate the proposed use-case of post-OCR correction, we simulate the digitization pipeline. First, we synthetically create a PDF from each corpus in the following manner:

- We open the text file in a text editor. (We use `gedit` with `Sanskrit2003` font.)
- We next trigger the operating system's (Ubuntu 18.04) print dialogue (`Ctrl+P`)
- We use the '*Print to File*' option and save the file as PDF.

A PDF file created in this manner is the best-case scenario for OCR engines, as it contains no noise. Now, we run both the OCR systems and obtain the OCR-ed versions of the text. As a result of this simulation, we can now realistically evaluate the detection and correction of errors introduced in the process of OCR.

### 5.3.2 Results

We use the following abbreviations for different versions of the corpora:

- **WS**: Wikisource
- **SD**: sanskritdocuments.org
- **GR**: GRETIL

[18] https://sa.wikisource.org/s/1c5
[19] https://sanskritdocuments.org/doc_z_misc_major_works/meghanew.html
[20] http://gretil.sub.uni-goettingen.de/gretil/1_sanskr/5_poetry/2_kavya/kmeghdpu.htm
[21] https://sa.wikisource.org/s/7xr
[22] https://sa.wikisource.org/s/7up
[23] https://sa.wikisource.org/s/7xq

- **GO**: Google Drive OCR on a synthetically generated PDF
- **TO**: Tesseract OCR on a synthetically generated PDF

We evaluate our Chandojñānam system as well as those by [Rajagopalan, 2020] and [Neill, 2023] on each version of the corpora. We evaluate the ability of these systems to identify the chanda of the verse in the presence of errors. Table 5.3 illustrates the result of this evaluation. Chandojñānam was able to identify the correct meter from the erroneous text in $98.2$% of the cases, performing better than the systems by [Rajagopalan, 2020] ($91.9$%) and [Neill, 2023] ($80.3$%). Corpora, code and results of the experiments are made available with the source code.

### 5.3.3 Error Analysis

We now analyse the errors in a more detailed manner.

Despite the Wikisource version being prepared through manual correction of OCR, Chandojñānam was still able to detect $2$ errors from Meghadūta. The errors are described below:

- Line: कालक्षेपं ककुभसुरभौ पर्वते पर्वेते ते (Pāda 3, Śloka 1.23)
- Suggestion: [[['का', 'ल', 'क्षे', 'पं'], ['क', 'कु', 'भ', 'सु', 'र', 'भौ'], ['प', 'र्व', 'ते'], ['प', 'r(र्वे)[L]', 'ते'], ['ते']]]
- Description: The error is due to the incorrect word पर्वेते. It is likely that this was due to an oversight by the curator. It can be seen that the system correctly points to the location where a change is required.
- Line: साभिज्ञानप्रहितकुशलैस्ततद्वचोभिर्ममापि (Pāda 3, Śloka 2.53)
- Suggestion: [[['सा', 'भि', 'ज्ञा', 'न', 'प्र', 'हि', 'त', 'कु', 'श', 'लै', 'd(स्त)', 'त', 'द्व', 'चो', 'भि', 'र्म', 'मा', 'पि']]]
- Description: The error is due to an extra letter present in this line, where an extra त appears in the sandhi of words कुशलैः and तद्वचोभिः, resulting in कुशलैस्ततद्वचोभिः instead of कुशलैस्तद्वचोभिः. The system is able to identify the error, and point out that a syllable needs to be deleted. However, we can see that the system points to an incorrect syllable स्त to be deleted. This can be explained from the fact that both स्त and त are laghu letters, and deletion of either letter results in the correct metrical signature. So, it is impossible for a meter identification based system to correctly say which character is to be deleted without any notion of semantic consideration. Such type of semantic error correction is out-of-scope as of yet.

**Table 5.3:** Error tolerance of meter identification systems. (Versions are WS: Wikisource, GO: Google OCR, TO: Tesseract OCR, SD: sanskritdocuments.org, GR: GRETIL.) Chandojñānam is able to detect correct chanda from erroneous verses 98.2% of the times.

| | | Meghadūta | | | | | Śāntavilāsa | | | Rāmarakṣā | | | Rājendrakarṇapūra | | | Total |
|---|---|---|---|---|---|---|---|---|---|---|---|---|---|---|---|---|
| | | SD | GR | WS | GO | TO | WS | GO | TO | WS | GO | TO | WS | GO | TO | |
| Number of Verses | | 117 | 111 | 123 | 123 | 123 | 36 | 36 | 36 | 39 | 39 | 39 | 72 | 72 | 72 | 1038 |
| Unique Chanda | | 1 | 1 | 1 | 1 | 1 | 12 | 12 | 12 | 9 | 9 | 9 | 4 | 4 | 4 | 17 |
| Erroneous Verses | | 20 | 79 | 2 | 31 | 77 | 13 | 16 | 31 | 1 | 4 | 13 | 12 | 26 | 71 | 396 |
| Correct | [Neill, 2023] | 20 | 79 | 2 | 30 | 66 | 11 | 13 | 14 | 0 | 2 | 9 | 12 | 24 | 36 | 318 (80.3%) |
| Meters | [Rajagopalan, 2020] | 19 | 79 | 2 | 30 | 75 | 12 | 15 | 24 | 1 | 2 | 9 | 12 | 26 | 58 | 364 (91.9%) |
| Identified | Chandojñānam | 20 | 79 | 2 | 31 | 77 | 13 | 16 | 29 | 1 | 3 | 9 | 12 | 26 | 71 | 389 (98.2%) |

Majority of the 'errors' in GRETIL stem from the presence of a whitespace in the text at the joining point of sandhi. For example, the text मेघमाश्लिष्टसानुं is written as two words (मेघम् आश्लिष्टसानुं), which if considered as is, would make the second syllable of the phrase घम् (guru) instead of घ (laghu). Despite not being linguistic errors, from the point of view of Chandaḥśāstra, they may change the number of syllables in a pāda and subsequently result in the change or loss of the meter. Such errors can be fixed by ignoring the spaces while computing the metrical signature.

It is important to remember that the errors detected are only the errors reported by the Chandojñānam system. Errors that do not result in the breaking of metrical pattern are impossible to be corrected by metrical analysis. It is interesting to note that the fuzzy matching performs better for Google OCR than Tesseract OCR. It can be explained as follows. As the Google OCR system performs better and makes fewer errors per line, the possible deviations are fewer. This results in less false positives. On the contrary, if the OCR system makes numerous errors, the *lg-signature* is farther away from the actual *lg-signature* and, therefore, 'the closest match' of the erroneous *lg-signature* might also be erroneous.

Finally, it can be seen that although the meter identification can be a useful tool for detection of errors in the Sanskrit text, there are several other factors that affect the error rate of such a system.

## 5.4 Summary

In this chapter, we described a Sanskrit meter identification tool that adds many user-friendly features and focuses on tolerance towards erroneous text as well as correction of such text. The features such as meter identification from images are useful for Sanskrit enthusiasts without much programming background. Bulk analysis of text files is a useful aid for digitization of Sanskrit texts using the methodology of post-OCR manual correction.

The Chandojñānam system is accessible online at `https://sanskrit.iitk.ac.in/jnanasangraha/chanda/` and the source code is available at `https://github.com/hrishikeshrt/chanda/`.

# Chapter 6

# Miscellaneous Computational Tools for Sanskrit

In this chapter, we delve into a diverse collection of computational tools, web interfaces, and Python libraries designed to enhance the processing and analysis of Sanskrit. These tools serve a dual purpose, catering to both the general public with limited Sanskrit or programming knowledge and researchers in the field of Natural Language Processing (NLP). The web interfaces provide user-friendly access to linguistic resources and functionalities, enabling a wider audience to explore and interact with Sanskrit texts. On the other hand, the Python libraries offer a seamless integration of state-of-the-art tools into NLP projects, accelerating research and development in the field. Through an exploration of these tools, we aim to showcase their capabilities, usability, and their valuable contributions to advancing Sanskrit language processing and NLP research.

## 6.1 Jñānasaṅgrahaḥ: Computational Interfaces

Jñānasaṅgrahaḥ is a collection of several web-based computational applications related to the Sanskrit language. The aim is to highlight the features of Sanskrit language in a way that is approachable for an enthusiastic user, even if she has a limited Sanskrit background. Jñānasaṅgrahaḥ is available at `https://sanskrit.iitk.ac.`

in/jnanasangraha/. The applications part of Jñānasaṅgrahaḥ are described in the following sections.

### 6.1.1 Saṅkhyāpaddhatiḥ

In the ancient India, it was a common practice to represent numeric values using letters, syllables or words from a natural language. The primary reason to use such systems is, ease of remembrance of numbers. We present a user-friendly web-based interface, Saṅkhyāpaddhatiḥ, which implements three such ancient Indian numeral systems, Kaṭapayādi Saṅkhyā, Āryabhaṭīya Saṅkhyā and Bhūtasaṅkhyā. The former two are alpha-syllabic numeral systems, while the latter is a number notation that uses ordinary words having implication of numeral values.

The central idea of an alpha-syllabic systems is that numeric values of the syllables are defined based on the constituent consonants and vowels. Usually, more than one syllable is assigned the same numerical value, however, every syllable has a unique numerical value, i.e. a many-to-one mapping of syllables to numbers. As a result, there is a unique value associated with a valid word or a phrase in a system, but there might be many valid representations of a number in the language.

The core interface for each of the system consists of an encoding interface to encode numeric values into a valid text representation a decoding interface to decode any valid text representation into the corresponding numeric value.

The Saṅkhyāpaddhatiḥ system is available at https://sanskrit.iitk.ac.in/jnanasangraha/sankhya/.

#### 6.1.1.1 Kaṭapayādi Saṅkhyā

The Kaṭapayādi system of encoding numbers as words by substituting each digit by a character was developed in ancient India. Each of the letters, क (k), ट (ṭ), प (p) and य (y), is assigned the number 1. The subsequent characters are assigned the numbers 2, and so on, thus giving rise to the term Kaṭapayādi, which signifies that 'k', 'ṭ', 'p', and 'y' are the first characters of this sequence. Multiple characters can be mapped to the same number, however there is only one number for each character. Thus, the system allows a number to be represented in multiple ways. A famous example of the use of this system is 'भद्राम्बुद्धिसिद्धजन्मगणितश्रद्धा स्म यद् भूपगी:' in Sadratnamālā of Śaṅkaravarmā. In the Kaṭapayādi system, this evaluates to 314159265358979324, denoting the value of $\pi$ up to 17 decimals.

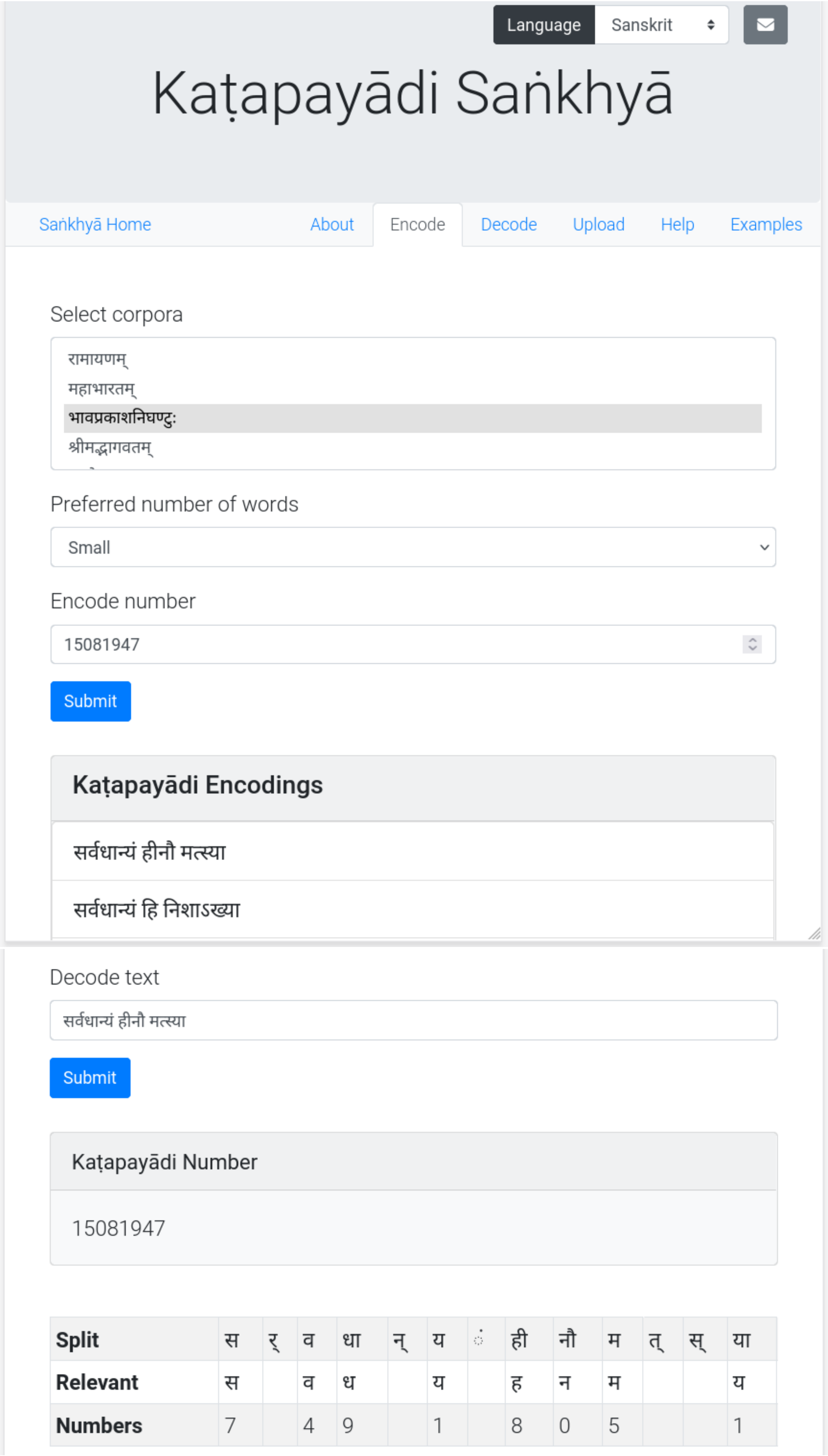

**Figure 6.1:** Saṅkhyāpaddhatiḥ: Kaṭapayādi Encoding and Decoding

While decoding a text representation to the corresponding numerical value is straightforward, as each character represents a single digit, encoding poses challenges due to the possibility of multiple combinations. To address this, we employ a data-driven approach that leverages multiple Sanskrit corpora. We decode all the words in these corpora and store them in an inverted index. This index includes not only single words but also the corresponding decoded numbers for bi-grams and tri-grams.

When we need to encode a number, we first search the inverted index for a direct match. If no match is found, we divide the number into smaller parts and search for matches for each constituent part. The encoded words we retrieve are guaranteed to be grammatically correct since they are sourced from actual Sanskrit texts. However, it is important to note that the resulting combination of words may not always carry meaningful semantic significance since they are formed by arbitrary word combinations.

Figure 6.1 shows the encoding and decoding capabilities of Kaṭapayādi system.

#### 6.1.1.2 Āryabhaṭīya Saṅkhyā

The Āryabhaṭīya numerical system was developed by the Indian mathematician and astronomer Āryabhaṭa. This system is described in the first chapter called Gītika Padam of his work Āryabhaṭīya. In this system, each syllable formed by a combination of consonant and vowel in Sanskrit phonology is assigned a numerical value in a systematic manner. Vowels are assigned values of even powers of ten, for example the vowel अ (a) is given the value $10^0 = 1$, the vowel इ (i) is assigned the value $10^2 = 100$ and so on, with औ (au) holding the largest value of $10^{18}$. The system does not differentiate between short and long vowels, so the pairs of vowels अ (a) and आ (ā), इ (i) and ई (ī), and so on have same values. On the other hand, the consonants क (k) through म (m) are assigned values from 1 to 25 with the remaining consonants य (y) to ह (h) are given values 30, 40, . . ., 100.

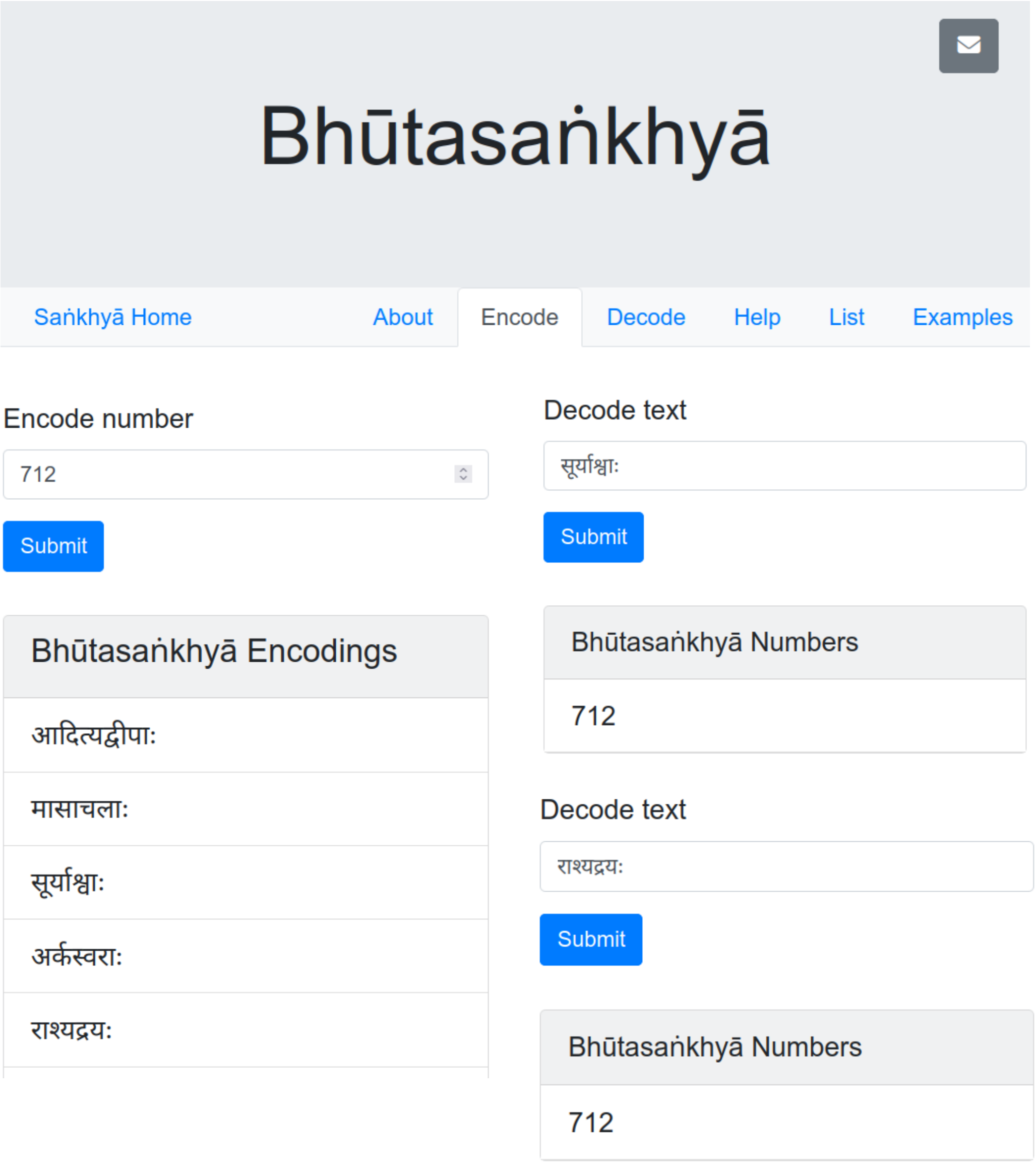

**Figure 6.2:** Saṅkhyāpaddhatiḥ: Bhūtasaṅkhyā Encoding and Decoding

The encoding and decoding processes in the Āryabhaṭīya Saṅkhyā system are relatively straightforward, owing to its strong connection with the decimal system. To assist with the manipulation of Sanskrit alphabets (varṇa), we utilize the Python library *sanskrit-text*, which is further described in Section 6.3.3.

#### 6.1.1.3 Bhūtasaṅkhyā

The Bhūtasaṅkhyā system is a method of expressing numbers through the use of common words that inherently carry numerical connotations. For instance, the word Veda symbolizes the number 4 as there are four primary Veda. Similarly, words like ṛtu, ripu, rasa, darśana, and others can represent the numerical value 6, as these words are associated with the number 6. It is important to note that sometimes a single word may have multiple numbers naturally linked to it. For instance, rasa signifies the number 6 (ṣaḍrasāḥ) when considered in the context of taste in Āyurveda, whereas in the domain of Nāṭyaśāstra, it signifies the number 9 (navarasaḥ). Unlike the previous two systems, the Bhūtasaṅkhyā system operates in a many-to-many fashion, where multiple numbers can be associated with a single word. However, it is important to note that such exceptions are relatively infrequent.

To facilitate the encoding and decoding operations in the Bhūtasaṅkhyā system, we maintain a list of words along with their corresponding numerical associations. For decoding text, we break down the input text into constituent words with the help of the Sanskrit Sandhi and Compound Splitter [Hellwig and Nehrdich, 2018]. These words are then searched in the maintained index for their numerical representations.

When encoding numbers, we begin by dividing the input number into valid[1] groups of digits, giving precedence to larger digit groups. Next, we search for words associated with each digit group and construct a samāsa (compound word). To facilitate the process of combining the constituent words of the samāsa, we utilize the Python package *sandhi*[2] to perform sandhi (combination) between the individual words.

Figure 6.2 shows the encoding and decoding interface of Bhūtasaṅkhyā system.

[1]Number is considered valid if it has a word associated with it.
[2]`https://pypi.org/project/sandhi/`

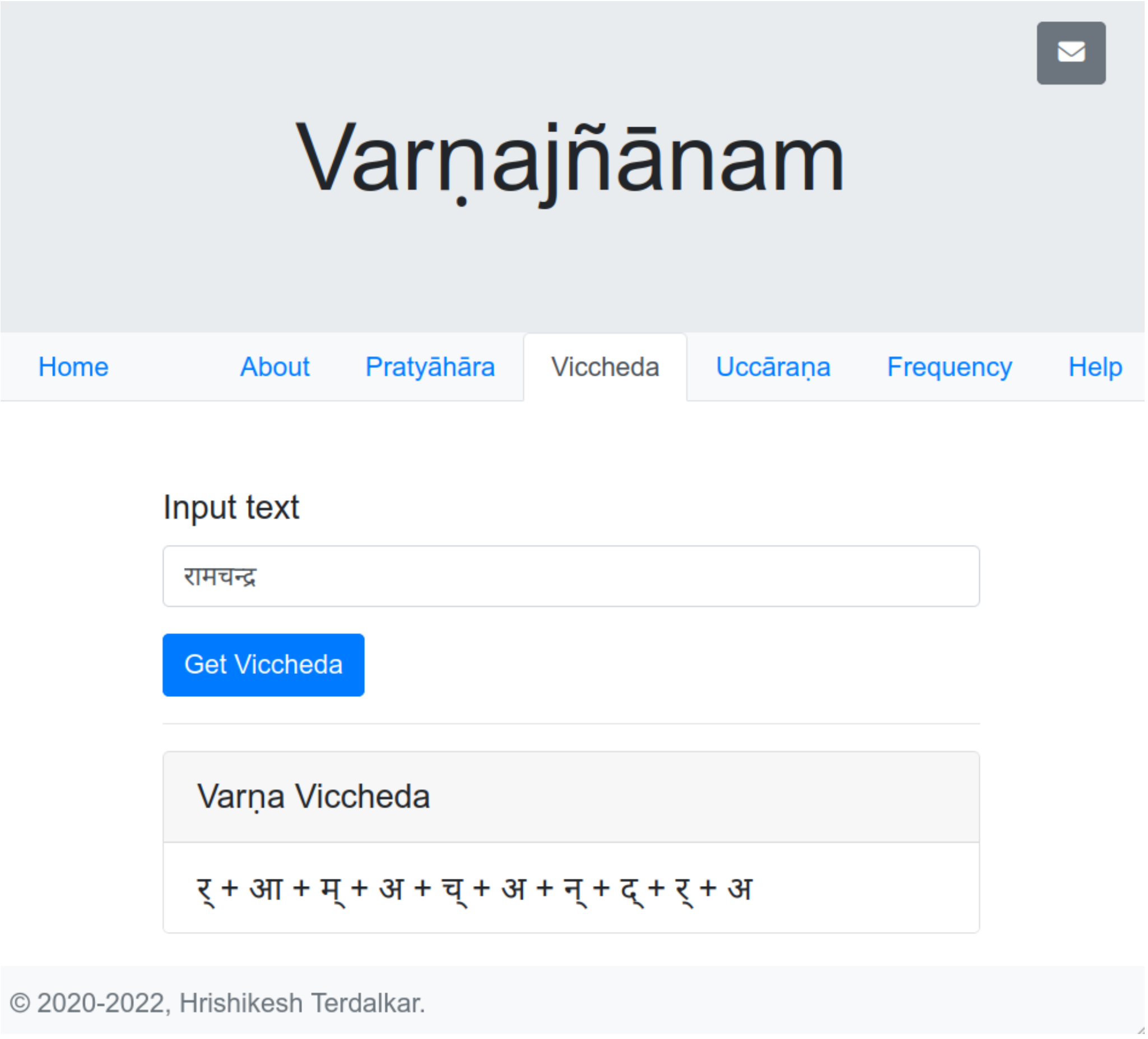

**Figure 6.3:** Varṇajñānam: Splitting varṇas (varṇavicchedaḥ)

### 6.1.2 Varṇajñānam

Varṇa (वर्ण) is a phonetic unit of Sanskrit language. The Varṇajñānam system consists of utility functions related to varṇa information and manipulation. These include the following utilities,

1. Pratyāhāra Manipulation: Formation and resolution of Pratyāhāra, which correspond to grouping of several letters in small groups in Sanskrit grammar.

2. Varṇa Viccheda: Splitting a Sanskrit word into its component alphabets (varṇa).

3. Uccāraṇa Information: Information about the pronunciation aspects of varṇa, including uccāraṇa sthāna (place of articulation) and prayatna (effort or intensity of pronunciation).

4. Frequency Calculator: Calculate the frequency of the varṇa as well their pro-

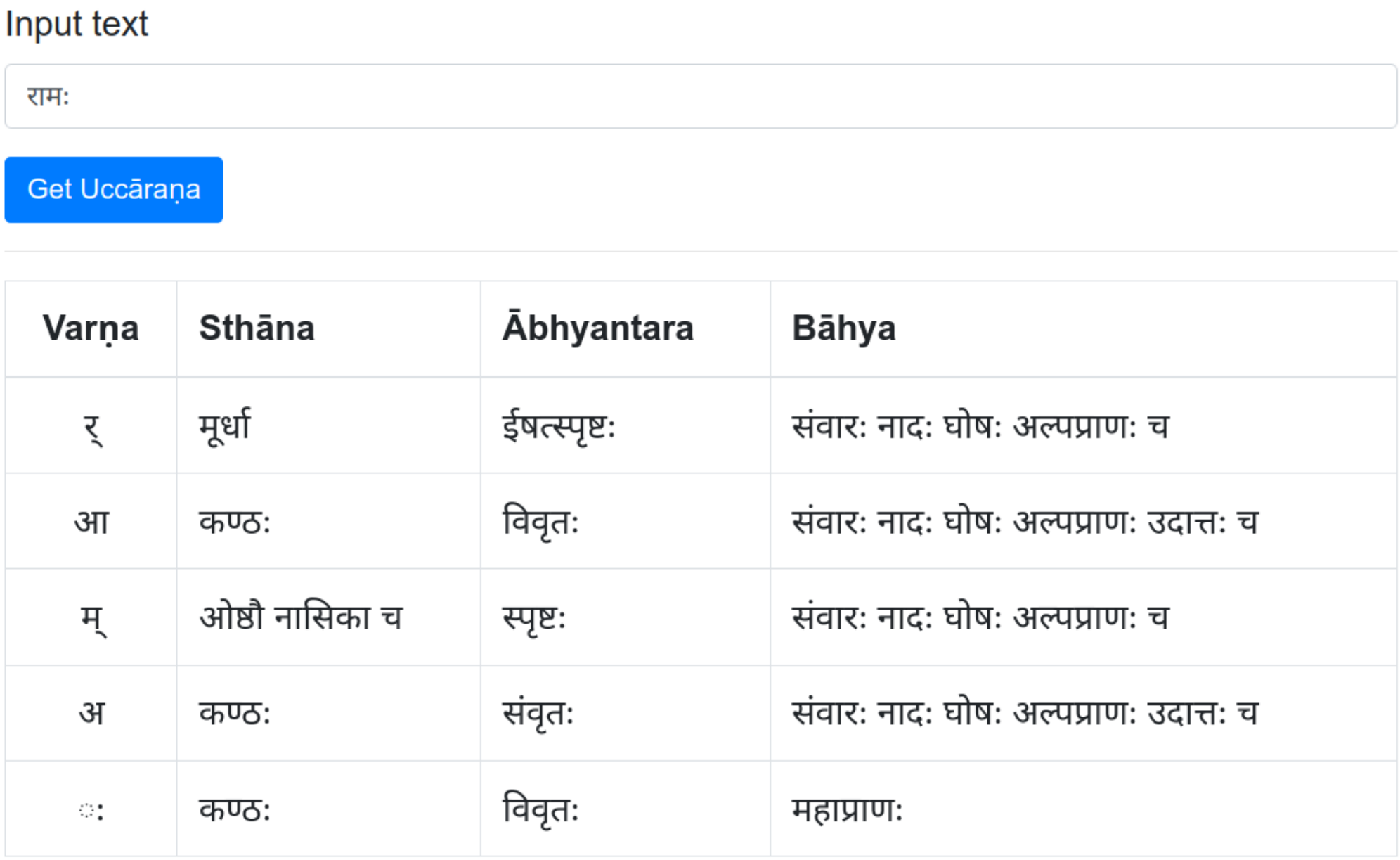

| Varṇa | Sthāna | Ābhyantara | Bāhya |
|---|---|---|---|
| र् | मूर्धा | ईषत्स्पृष्टः | संवारः नादः घोषः अल्पप्राणः च |
| आ | कण्ठः | विवृतः | संवारः नादः घोषः अल्पप्राणः उदात्तः च |
| म् | ओष्ठौ नासिका च | स्पृष्टः | संवारः नादः घोषः अल्पप्राणः च |
| अ | कण्ठः | संवृतः | संवारः नादः घोषः अल्पप्राणः उदात्तः च |
| ः | कण्ठः | विवृतः | महाप्राणः |

**Figure 6.4:** Varṇajñānam: Pronunciation information Uccaraṇasthānayatna

nunciation classes from Sanskrit text.

Figures 6.3 and 6.4 showcase the interface for splitting varṇa and the interface for displaying pronunciation information. The system is available at `https://sanskrit.iitk.ac.in/jnanasangraha/varna/`.

## 6.2 Vaiyyākaraṇaḥ: A Sanskrit Grammar Bot for Telegram

Rise of social media opens novel opportunities of learning. Telegram[3] is an instant messaging service available as a cross-platform, freemium software. Telegram bots are instances of Telegram clients that are capable of performing actions in response to various user actions. Vaiyyākaraṇaḥ is a telegram bot aimed towards helping the learners of Sanskrit grammar (vyākaraṇa).

Vaiyyākaraṇaḥ is made using *Telethon*[4], a Telegram client library in Python 3. It makes use of data (Dhātupāṭhaḥ) from `https://ashtadhyayi.com` for conjugations. The bot also uses some of the state-of-the-art Sanskrit computational linguistic tools. The Heritage Platform [Goyal et al., 2012] is used for tasks related to declensions. Sanskrit Sandhi and Compound Splitter [Hellwig and Nehrdich, 2018] is used for the word segmentation task. The user of the bot may type a set of provided keywords as commands followed by the appropriate input text in their Telegram client to obtain the output.

[3] `https://telegram.org/`
[4] `https://docs.telethon.dev/en/latest/`

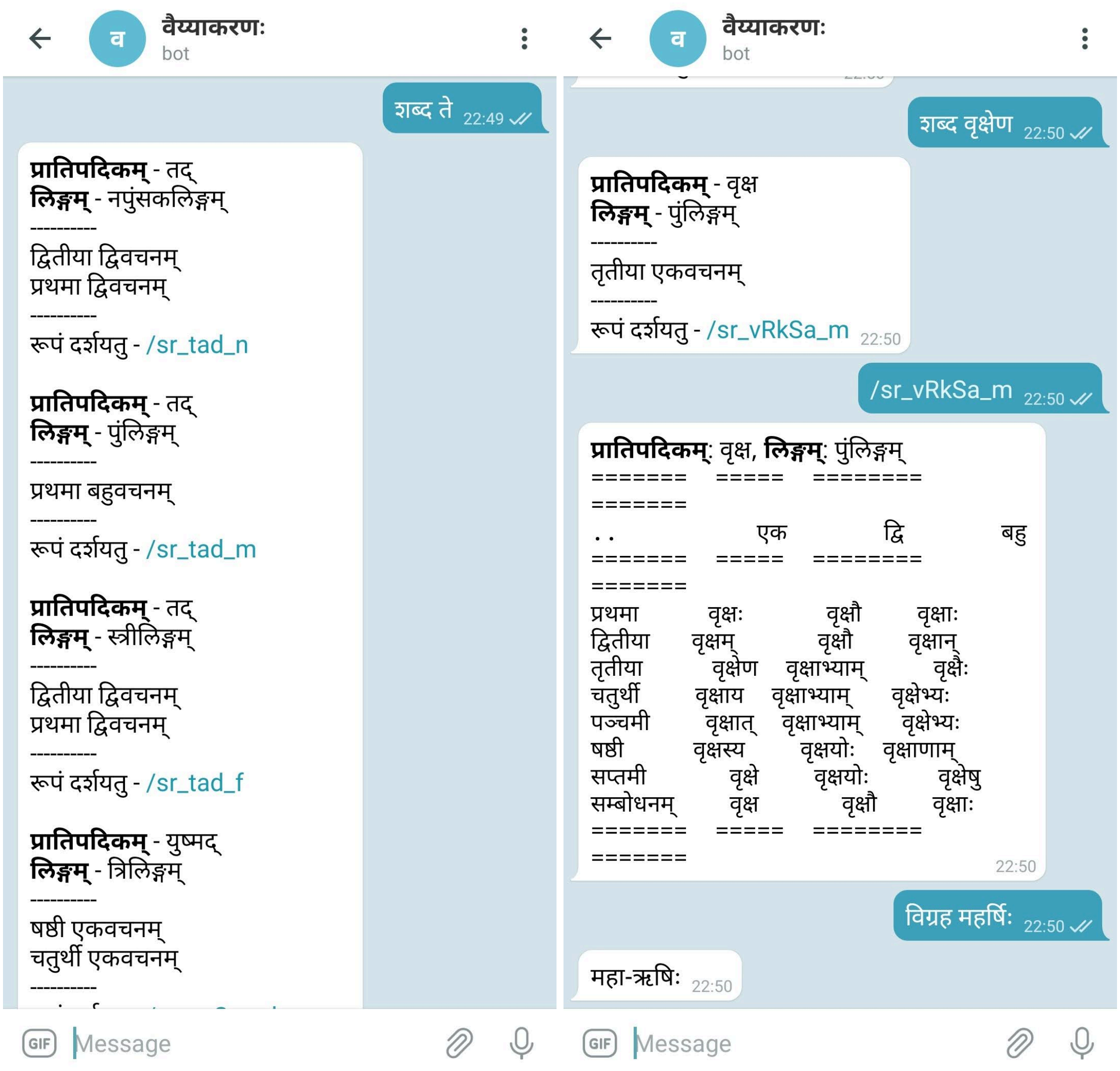

**Figure 6.5:** Stem Finder and Declension Generator

The salient features of the bot are:

- Stem finder (Prātipadikam)

  The keyword '\shabda' can be used, followed by a word form (subanta) to obtain the stem (prātipadikam) and morphological information.

- Declension generation (Subantāḥ)

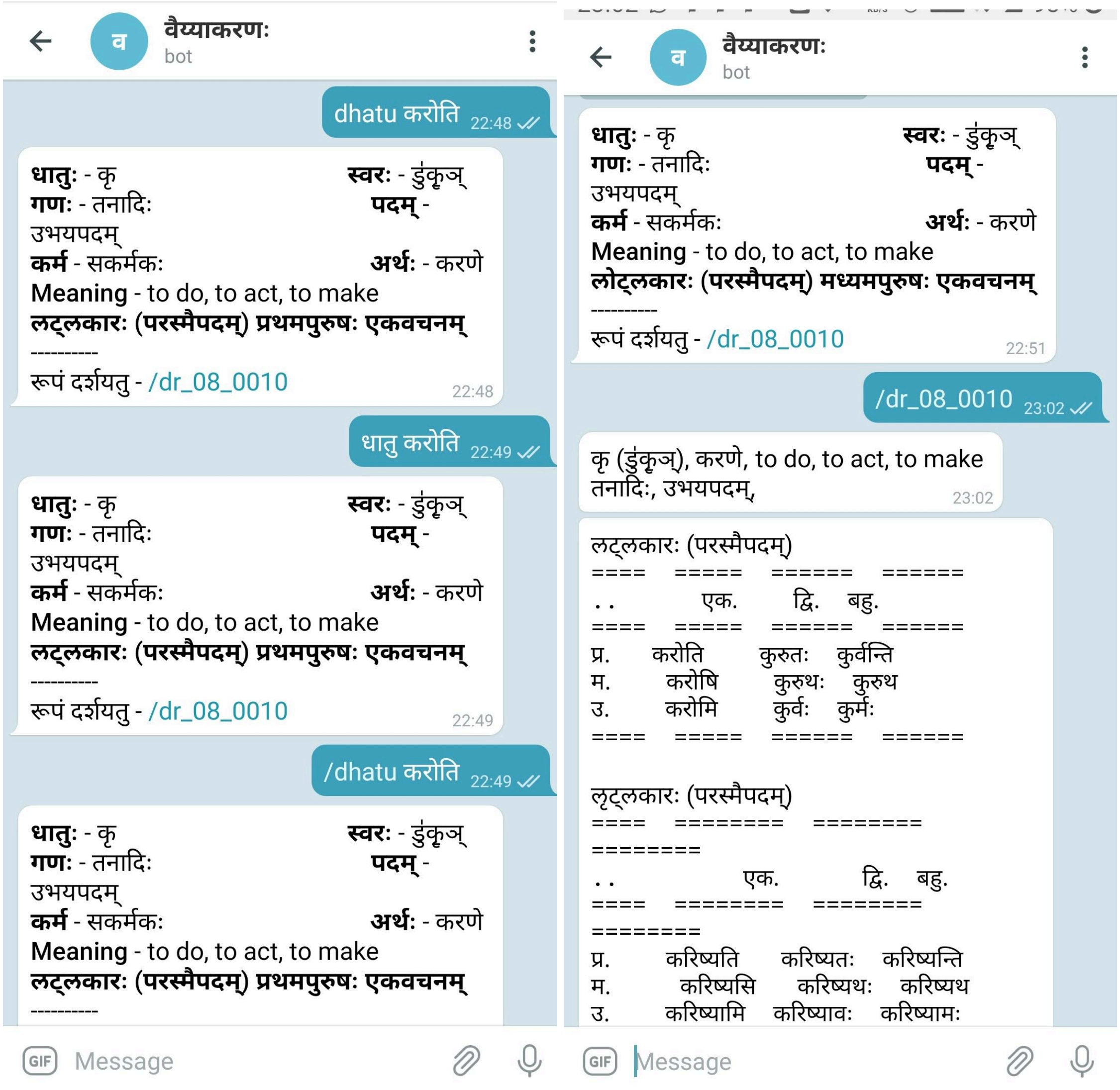

**Figure 6.6:** Root Finder and Conjugation Generator

Upon searching a word form, the bot also provides an option to show all morphological forms (declensions) of the provided word.

- Root finder (Dhātuḥ)

  The keyword '`\dhatu`' can be used, followed by a verb form (tiṅanta) to obtain the root (dhātu) and morphological information.

- Conjugation generation (Tiṅantāḥ)

  Upon searching a verb form, the bot also provides and option to display all morphological forms (conjugations) of the provided verb.

- Word segmentation(Sandhisamāsau)

  The keyword '`\vigraha`' can be used, followed by Sanskrit text can be pro-vided to obtain the word segmentation (splitting both sandhi and samāsa).

Figure 6.5 and Figure 6.6 showcase the diverse capabilities of Vaiyyākaraṇaḥ. A video demo is available at `https://sanskrit.iitk.ac.in/vaiyyaakarana/`. The bot can be accessed at `https://t.me/vyakarana_bot`. The source code can be found at `https://github.com/hrishikeshrt/vaiyyakarana/`[5].

## 6.3 Python Libraries

We have developed a set of Python packages available on the Python Package Index (PyPI[6]) to facilitate NLP researchers in working with Sanskrit text and corpora. These packages can be conveniently installed using the '`pip install`' command. In the following sections, we will provide a description of each package.

### 6.3.1 *PyCDSL*: A Programmatic Interface to Cologne Digital Sanskrit Dictionaries

*PyCDSL* is a Python library that provides programmer friendly interface to Cologne Digital Sanskrit Dictionaries (CDSD) [cds, 2022]. The library serves as a corpus management tool to download, update and access dictionaries from CDSD. The tool provides a command line interface for ease of search and a programmable interface for using CDSD in computational linguistic projects written in Python 3.

The command line interface is provided in two modes (1) a console command (`cdsl`) and (2) an interactive REPL[7] interface. Both modes come with a rich search functionality. Users may search by key (a dictionary entry) or value (meaning provided in the dictionary) or both. The search can be performed on multiple dictionaries at the same time. The interactive REPL mode can be triggered by passing option '`-i`' to the console command '`cdsl`'. Once in this mode, a user may simply input a term to perform search. Search related settings can also be changed in this

[5] Please refer to `INSTALL.md` for installation instructions.
[6] `https://pypi.org/`
[7] `https://en.wikipedia.org/wiki/Read-eval-print_loop`

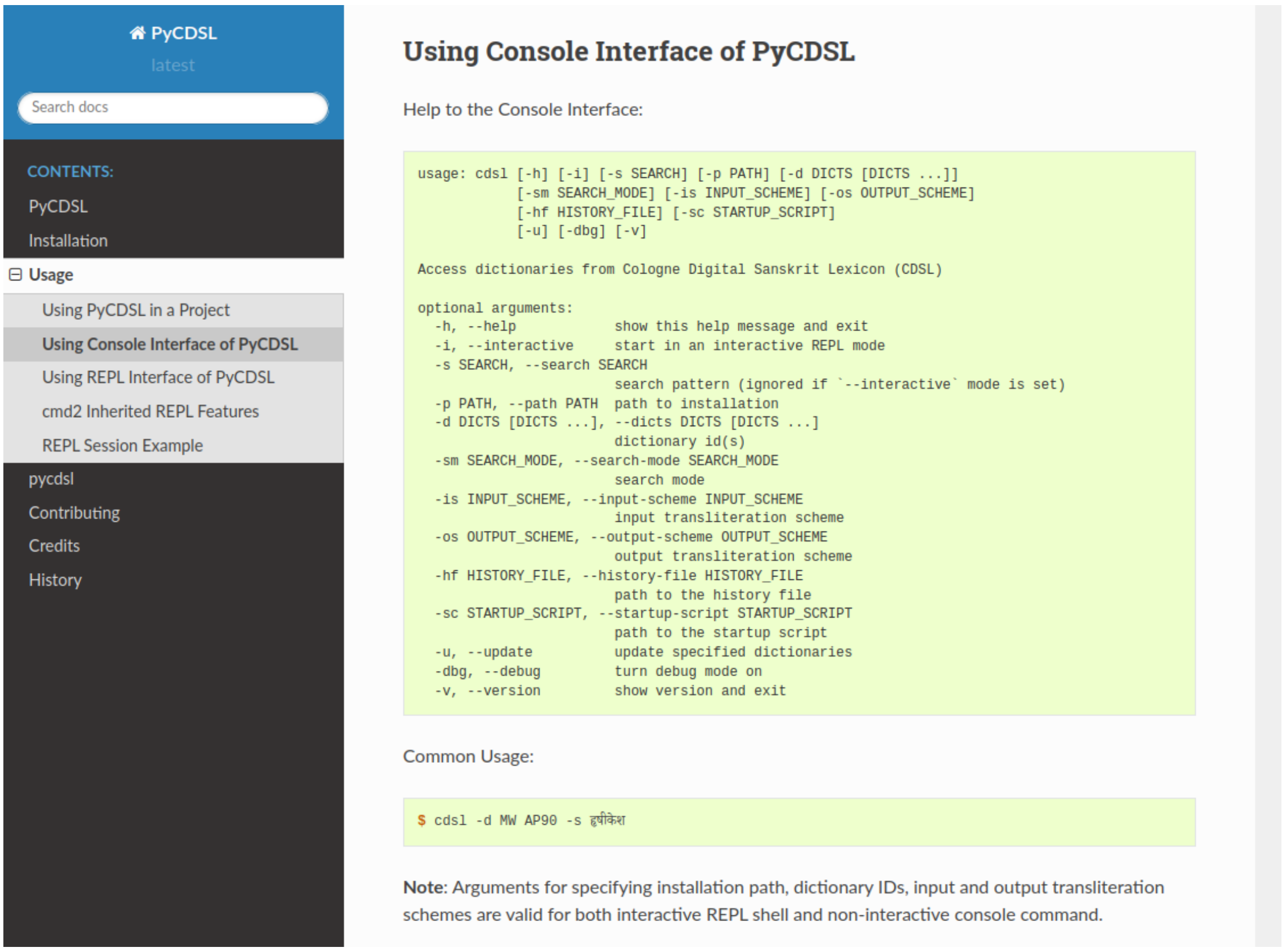

**Figure 6.7:** Usage Instructions for CLI

mode. The library also comes with extensive transliteration support powered by Python library *indic-transliteration*[8]. The input can be provided in any of the supported schemes and the results can be exported in a scheme of choice as well. The results can be copied to clipboard (powered by *Pyperclip*[9])

Every dictionary uses specific conventions in the meaning (value) of the dictionary entries (key). The library can be extended with custom parsers to parse the entries in the dictionaries in a custom manner.

Figure 6.7 shows a part of the documentation detailing usage instructions for the console command. The detailed documentation can be found at `https://pycdsl.readthedocs.io/en/latest/`. The library is available at `https://pypi.org/project/PyCDSL/` and can be installed using the command, `pip install PyCDSL`. The source code can be found at `https://github.com/hrishikeshrt/PyCDSL/`.

[8] `https://github.com/indic-transliteration/`
[9] `https://pypi.org/project/pyperclip/`

### 6.3.2 *Heritage.py*: Python Interface to The Sanskrit Heritage Site

*Heritage.py* is a Python package that serves as an interface to The Sanskrit Heritage Site [Goyal et al., 2012]. It makes a number of features offered by the Heritage Platform available to use in Python projects for working with Sanskrit. The features include morphological analysis, sandhi formation, declensions, and conjugations.

*Heritage.py* offers two distinct modes of operation:

- **Web Mirror Mode:** In this mode, the package utilizes a compatible web mirror of The Heritage Platform (i.e., `https://sanskrit.inria.fr/index.en.html`). This mode does not necessitate any installation steps but relies on HTTP requests for each task, which may introduce a slight delay in obtaining results.

- **Local Installation Mode:** To leverage the full potential of *Heritage.py* with enhanced performance, a local installation[10] of The Heritage Platform is required. This mode significantly accelerates result acquisition by eliminating the need for frequent HTTP requests.

The detailed documentation is available at `https://heritage-py.readthedocs.io/en/latest/`. The Python package is available on PyPI at `https://pypi.org/project/heritage/` and can be installed using '`pip install heritage`' on the terminal. The source code is available at `https://github.com/hrishikeshrt/heritage`.

### 6.3.3 *sanskrit-text*: Sanskrit Text Utility Functions

*sanskrit-text* is a Python package designed to provide various utility functions for working with Sanskrit text in the Devanagari script. It offers a range of functionalities that assist developers in processing Sanskrit text and corpora.

The library's capabilities include:

- Syllabification: The library provides functions to syllabify Sanskrit words, splitting them into their constituent syllables.

[10]The Heritage Platform Installation Instructions: https://sanskrit.inria.fr/manual.html#installation

- Varṇa Viccheda: This feature allows one to split Sanskrit words into individual phonetic units (varṇa)
- Pratyāhāra Encoding-Decoding: Pratyāhāra, in the context of Sanskrit grammar, are shortened representations of a larger set of characters. The library supports encoding and decoding of pratyāhāras.
- Uccāraṇa Sthāna Yatna Utility: This utility assists in identifying the information about pronunciation of sounds in a Sanskrit text.
- The library offers additional functions for tasks such as cleaning text, identifying lines from a Sanskrit text, manipulating mātrās (diacritical marks), and more.

The library serves as a backend for Varṇajñānam interface described in Section 6.1.2. It provides a comprehensive set of functions to aid in the character or syllable level processing of Sanskrit text.

Its documentation, available at `https://sanskrit-text.readthedocs.io/en/latest/`, offers detailed information on each function and its usage. The package itself is available on PyPI at `https://pypi.org/project/sanskrit-text/` and can be installed using the command, `pip install sanskrit-text`, in the console. The source code for *sanskrit-text* library is available at `https://github.com/hrishikeshrt/sanskrit-text`.

## 6.4 Summary

We have showcased a wide range of computational tools, web interfaces, and Python libraries that support various tasks related to Sanskrit language understanding and exploration. These tools serve multiple purposes, catering to both general users with limited Sanskrit or programming knowledge, as well as Sanskrit NLP researchers and developers.

# Chapter 7

# Conclusions

In this thesis, we set out to construct knowledge-based systems in Sanskrit for the advancement of research in Sanskrit NLP as well as for promoting the dissemination of knowledge for the benefit of researchers, scholars, and the wider community. We explored the development of knowledge-based systems for Sanskrit, focusing on the construction of a question answering (QA) system and the utilization of a knowledge graph. We successfully developed a QA system that leverages the knowledge graph constructed automatically from Sanskrit texts, in order to answer natural language queries posed by the users. Additionally, we designed and implemented task-specific annotation tools to facilitate the construction and refinement of the knowledge graph. These annotation tools proved to be essential in annotating Sanskrit text corpora, enabling the semantic extraction of knowledge and the construction of a comprehensive knowledge graph. In this process we also created several useful web-interfaces, tools and software libraries to cater to the needs of Sanskrit researchers as well as enthusiasts.

We will now summarize the key contributions and discuss the scope for future work as well as the research directions enabled by this thesis.

## 7.1 Key Contributions

The significant contributions made by this thesis to the field of knowledge-based systems for Sanskrit are the following.

- We developed a framework to automatically construct a knowledge graph through processing of Sanskrit texts. We employed this knowledge graph in a domain-specific question answering system to answer questions based on human kinship relationships.

- We discovered several limitations in the automatic text processing of Sanskrit corpora and identified the need for manual annotation for higher level semantic tasks.

- We designed and implemented two intuitive and user-friendly annotation tools, *Sangrahaka* and *Antarlekhaka*, to enable distributed manual annotation of Sanskrit texts. Both the tools have received mostly positive reviews in a subjective evaluation and outperform other annotation tools on an objective scale.

- *Sangrahaka* is a task-specific annotation tool that streamlines the process of constructing and refining the Sanskrit knowledge graph. It has been used in the semantic annotation of Bhāvaprakāśanighaṇṭu, an Āyurveda text and a knowledge graph has been constructed on three chapters: Dhānyavarga, Śākavarga and Māṃsavarga from Bhāvaprakāśanighaṇṭu. The knowledge graph contains more than 1600 nodes and more than 1700 relationships. The system, titled Āyurjñānam, is available at `https://sanskrit.iitk.ac.in/ayurveda/`.

- *Antarlekhaka* is general purpose annotation tool for the annotation towards a comprehensive set of NLP tasks. It introduces the concept of sequential annotation wherein an annotator completes a stipulated set of tasks for a small unit of text (such as a verse from a poetry corpus) before moving on to the next unit. The tool supports eight generic categories of annotation tasks which cover a much larger set of NLP tasks. These categories are, sentence boundary detection, canonical word ordering, token text annotation, token classification, token graph, token connections, sentence classification and sentence graph. The two tasks, sentence boundary detection and canonical word ordering, are not supported by any of the other extant annotation tools. Each category has a special annotation interface that makes it easy for annotators to capture the relevant information for each task. The tool is being used for a large-scale task of the annotation of a voluminous Sanskrit text, Vālmīki Rāmāyaṇa.

- We created, Chandojñānam, a system for identification and utilization of Sanskrit meters that can identify meters from Sanskrit text as well as images through use of state-of-the-art OCR engines. Through its ability to provide approximate and close matches in the absence of direct matches we demonstrated the utility of Sanskrit prosody in the digitization of Sanskrit corpora through locating the errors in Sanskrit text based on deviations from known metrical patterns.

- We also presented a collection of web-interfaces, tools, and software libraries related to computational aspects of Sanskrit. These include, Jñānasaṅgrahaḥ, a collection of web-interfaces showcasing interesting applications of Sanskrit, Vaiyyākaraṇaḥ, a Telegram bot to aid the learners of Sanskrit grammar and several Python libraries such as *PyCDSL*, *Heritage.py* and *sanskrit-text* which enable Sanskrit programmers to deal with Sanskrit texts in a more effective manner.

## 7.2 Future Work

While this thesis has made significant contributions through the development of knowledge-based systems for Sanskrit, there are several areas that offer avenues for future research and development.

- **Completion of Annotation Tasks:** The completion of ongoing annotation tasks

described in this thesis can lead to construction of more competent QA systems.

- **Expansion of the Knowledge Graph:** The knowledge graphs constructed in this thesis serve as a foundation, but the scope of the KGs can be further expanded and enriched with additional Sanskrit texts and resources.

- **Improvement of Annotation Tools:** The task-specific annotation tools developed in this thesis can be further refined and enhanced to improve the efficiency and accuracy of the annotation process. Integration with advanced natural language processing techniques, state-of-the-art tools, and machine learning algorithms can automate and streamline certain annotation tasks.

- **Semantic Search and Reasoning:** While the developed QA system provides accurate answers to user queries, incorporating semantic search and reasoning capabilities can enhance the system's ability to understand complex queries and provide more sophisticated answers.

- **Correction of Digital Corpora:** We explored the scope of Sanskrit prosody in the correction of digital Sanskrit corpora. The system is capable of identifying locations where errors may be present. Use of semantic resources, dictionaries, language models in combination with the metrical analysis may result in a more robust solution for providing suggestions in case of errors.

# Publications